\documentclass[11pt,letterpaper]{article}

\usepackage[title]{appendix}

\usepackage{caption}
\usepackage[dvipsnames,table]{xcolor}
\usepackage[margin = 1 in]{geometry}
\usepackage{amsmath}
\usepackage{arydshln} 
\usepackage{booktabs, multirow, makecell, tabularx, array, xcolor, colortbl, hhline}
\usepackage{subcaption}
\usepackage{longtable}
\usepackage{supertabular}
\usepackage{float}
\usepackage{graphicx} 
\usepackage{subcaption} 
\usepackage{float}
\usepackage{url} 
\usepackage{wrapfig}
\usepackage{siunitx} 
\usepackage{caption} 
\usepackage{stfloats}
\usepackage{placeins}

\usepackage{hyperref}
\usepackage[capitalise,noabbrev]{cleveref}
\usepackage[table]{xcolor}
\AtBeginEnvironment{appendices}{\crefalias{section}{appendix}}

\newcommand{\corrsym}{\protect\footnotemark[1]}
    \makeatletter
\def\@fnsymbol#1{\ensuremath{\ifcase#1\or \dagger\or *\or \ddagger\or
   \mathsection\or \mathparagraph\or \|\or **\or \dagger\dagger
   \or \ddagger\ddagger \else\@ctrerr\fi}}
    \makeatother

\usepackage{titling}
\thanksheadextra{}{}
\thanksheadextra{}{} 
\usepackage{lipsum} 

\title{On the Brittleness of Maximum Likelihood Estimation for Gaussian Process Hyperparameter Optimization}
\date{\vspace{-5ex}}
\usepackage{authblk}
\author[1]{Tyler R. Johnson}
\author[1]{Kian Ben-Jacob}
\author[1]{Christopher P. Muller}
\author[1,2]{Ramin Bostanabad\corrsym}  
\affil[1]{Department of Mechanical and Aerospace Engineering, University of California, Irvine}
\affil[2]{Department of Civil and Environmental Engineering, University of California, Irvine}

\begin{document}

    \pagenumbering{arabic}
    \sloppy
    \maketitle
    
    \begingroup
    \renewcommand{\thefootnote}{}
    \footnotetext{\textsuperscript{†}Corresponding Author: Raminb@uci.edu}
    \endgroup

    \section*{Abstract}

Machine learning (ML) has become an indispensable part of modern engineering design workflows. A crucial step in training an ML model is the selection of the loss function which can be systematically formulated via various techniques such as maximum likelihood estimation (MLE) and cross-validation . While MLE is one of the most popular, effective, and intuitive mechanisms for training ML models, it is brittle: if the assumptions underpinning it are not met, the trained ML model may generalize poorly. This brittleness affects even Gaussian processes (GPs) which are widely used in engineering design and are often (incorrectly) presumed to be very robust to overfitting. In this paper, we fundamentally evaluate the brittleness of MLE in the context of training GPs for probabilistic regression or classification tasks. We compare theoretically grounded metrics against MLE and propose practical solutions. Our extensive studies demonstrate the effectiveness of our solutions in downstream design tasks such as Bayesian optimization and provide a blueprint for practitioners to build accurate and robust GPs that can even outperform tabular foundation models in terms of prediction accuracy, uncertainty quantification, and inference cost. Our contributions are publicly available via GitHub at https://github.com/Bostanabad-Research-Group/GP-vs-TabPFN-vs-GPyTorch.

\noindent \textbf{Keywords:} Gradient-Based Optimization; Bayesian Optimization; Classification; Gaussian Processes; Tabular Foundational Models; Uncertainty Quantification.
    \section{Introduction} \label{sec introduction}
Machine learning (ML) has become an integral component of engineering design. In many design workflows, ML models are used to surrogate expensive simulations or experiments \cite{RN545,RN704,RN1024,RN791,RN1242,RN2082,RN2132}, model customer preferences \cite{10.1115/1.4044258, 10.1115/1.4044522, 10.1115/1.4044198}, fuse multi-source data \cite{RN1275,RN1577,RN1838,RN1845,RN2255,RN2254}, conduct sensitivity analyses \cite{RN452,RN535,RN459,RN463}, or calibrate computer models \cite{RN277,RN272,RN271,RN705,RN295,RN270,RN645}. As engineering systems grow in complexity, the reliance on data-driven surrogates continues to increase.

A central challenge in building an ML model lies in the specification of a loss function that determines how its parameters are learned from data. This can be done by several systematic strategies including maximum likelihood estimation (MLE) \cite{self1987asymptotic,andersen1970asymptotic}, cross-validation (CV) \cite{RN1694,RN146}, proper scoring rules \cite{Gneiting01032007}, convex surrogate modeling \cite{bartlett2006convexity}, and Kullback-Leibler Divergence \cite{kullback1951information}. Among these, MLE is perhaps the most adopted approach due to its strong statistical foundations, computational tractability, and interpretability. Many popular ML models such as neural networks (NNs) and Gaussian processes (GPs) are often trained by maximizing likelihood-based objectives.

Despite its popularity, likelihood-based training can become ineffective. The theoretical guarantees underlying MLE typically rely on assumptions such as correct model specification, independent observations, and sufficiently large data sets but many engineering design problems do not satisfy these conditions. When these assumptions are violated, models optimized through likelihood-based objectives can fit the training data well while generalizing poorly \cite{RN332}. This mismatch between apparent training performance and actual predictive capability is particularly problematic when designers need uncertainty quantification (UQ) too, e.g., in workflows such as Bayesian optimization (BO) where emulators (i.e., probabilistic surrogates) are needed.


The preceding discussions highlight the importance of careful selection of the ML model class and loss function, particularly in the presence of uncertainties such as lack of data and noise. 
These efforts contrast with the emerging paradigm built on foundation models (FMs) which are designed to generalize across many tasks with minimal task-specific training. FMs are increasingly based on the transformer architectures \cite{RN904} that employ attention mechanisms to capture complex dependencies within data. Beyond their success in domains such as natural language, FMs are used to approximate Bayesian posterior predictive quantities by learning from synthetic datasets \cite{RN2145,RN2147}. This capability enables them to produce probabilistic predictions for new tasks with no retraining. One prominent example in the tabular-data setting is the TabPFN family of models \cite{RN2143,RN2146}, which effectively replaces the traditional per-problem training phase with a pre-trained inference mechanism. Due to this design, TabPFN and similar models have emerged as compelling alternatives to classical probabilistic methods such as GPs \cite{Yu2025GITBO}.

These developments raise two important questions that have motivated this work: (1) how can we improve MLE-based training so that emulators such as GPs provide more accurate mean predictions \textit{and} UQ, and (2) how do modern FMs compare to GPs trained via MLE? 

We systematically investigate the brittleness of MLE in the context of training GP models for engineering design problems involving regression or classification. We propose practical remedies for mitigating this brittleness and discuss alternative loss functions that are often compared to MLE.
Our evaluation includes a comprehensive study involving $3,170$ independent simulations that explore a range of problem characteristics including varying noise levels, dataset sizes, and input dimensionalities. In addition to assessing predictive accuracy, we examine the quality of uncertainty estimates and analyze how training choices influence downstream performance in design workflows such as BO.

The rest of the paper is organized as follows. Relevant technical backgrounds are reviewed in \cref{sec technical background} and the proposed approach is described in \cref{sec method}. The results of our extensive comparative studies are provided in \cref{sec results} and the paper is concluded in \cref{sec conclusion}.

    \section{Technical Background} \label{sec technical background}

\subsection{Gaussian Processes} \label{subsec gp}
GPs provide priors over functions and are completely characterized by their mean function $m(\boldsymbol{x}; \boldsymbol{\beta})$ and kernel or covariance function $c(\boldsymbol{x}, \boldsymbol{x'}; \boldsymbol{\theta})$ where $\boldsymbol{\beta}$ and $\boldsymbol{\theta}$ are hyperparameters. Given a training dataset, GP emulation is typically\footnote{Bayesian approaches find a posterior distribution (as opposed to point estimates) for the hyperparameters.} cast as optimizing the hyperparameters of its mean and covariance functions. After optimization, the conditional formulas of multivariate normal distributions are used to make probabilistic predictions at any $\boldsymbol{x}^*$.

The majority of works on GPs adopt simple mean and covariance functions where the former is set to zero or a constant (i.e., $m(\boldsymbol{x}; \beta) = \beta$) while the latter is formulated via stationary kernels such as the Gaussian \cite{RN2089}:
\begin{equation}
    c(\boldsymbol{x}, \boldsymbol{x}'; \boldsymbol{\theta}) = \sigma^2 \exp\left(\frac{-1}{2}\sum_{i=1}^{D_x} \left(\frac{x_i - x_i'}{l_i} \right)^2 \right),
    \label{eq gaussian typical}
\end{equation}
where $\boldsymbol{\theta} = \{ \sigma^2, \boldsymbol{l} \}$ and $D_x$ denotes the dimensionality of $\boldsymbol{x}$ which may include categorical variables. With Gaussian noise, $\mathrm{cov}(y, y')$ is then $c(\boldsymbol{x}, \boldsymbol{x}'; \boldsymbol{\theta}) + {\sigma_\epsilon}^2\mathbf{1}(\boldsymbol{x} = \boldsymbol{x}')$
 where $\sigma^2_\varepsilon$ is the nugget corresponding to the noise variance.

Once the mean and covariance function are formulated, their hyperparameters are typically optimized via MLE which involves minimizing the negative log-marginal likelihood $\mathcal{L}(\cdot)$ defined as (constants are dropped):
\begin{equation}
\begin{aligned}
    \widehat{\boldsymbol{\beta}}, \widehat{\boldsymbol{\theta}}, \widehat{\sigma}^2_{\varepsilon} &=
    \underset{\boldsymbol{\beta}, \boldsymbol{\theta}, \sigma^2_{\varepsilon}}{\operatorname{argmin}}\, \hspace{2mm} \mathcal{L}(\boldsymbol{\beta},  \boldsymbol{\theta}, \sigma^2_{\varepsilon}) \\
    &= \underset{\boldsymbol{\beta}, \boldsymbol{\theta}, \sigma^2_{\varepsilon}}{\operatorname{argmin}}\, 
    \frac{1}{2}\ln |\boldsymbol{C}_{\varepsilon}| + \frac{1}{2}(\boldsymbol{y}-\boldsymbol{m})^{T} \boldsymbol{C}_{\varepsilon}^{-1}(\boldsymbol{y}-\boldsymbol{m})
\end{aligned}
\label{eq mle}
\end{equation}
where $\boldsymbol{y}$ is the observation vector, $\boldsymbol{m} = m(\boldsymbol{x}; \boldsymbol{\beta})$, and $\boldsymbol{C}_{\varepsilon} = c(\boldsymbol{x}, \boldsymbol{x}'; \boldsymbol{\theta}) + {\sigma_\epsilon}^2\boldsymbol{I}$.

\subsection{TabPFN} \label{subsec tabpfn}
TabPFN is built on the Prior-Data Fitted Network (PFN) framework which aims to approximate the Bayesian posterior predictive distribution induced by a user-defined prior over supervised learning tasks. This setup requires one to specify a prior over datasets and training involves repeatedly drawing datasets from this prior and optimizing the model to predict held-out labels conditioned on the context. Hence, upon offline training, the model can output posterior predictive distributions for new datasets through conditioning on the provided context.

TabPFN's predictive and UQ capabilities are tightly linked to its training data since the model is trained to return predictive distributions consistent with the posterior predictive implied by the synthetic data-generating process. Hence, to expose the model to diverse input-output relationships, TabPFN is trained on synthetic data generated by a mixture of generative mechanisms such as causal-style generators and Bayesian NNs.

Architecturally, TabPFN treats a dataset as the input object and uses attention to condition each query on the entire data. 
Such conditioning indicates that traditional ML models like GPs will be faster once fitted because TabPFN effectively performs dataset conditioning and prediction jointly. Additionally, the transformer attention has compute that scales quadratically with the sequence length (number of tokens) \cite{lin2021surveytransformers}. For TabPFN-style in-context inference, the computational cost scales as:
\begin{equation}
    \mathrm{Cost} = \mathcal{O}\!\left(N_c(N_c+N_q)\right),
\end{equation}
where $N_c$ is the number of labeled context (training) rows and $N_q$ is the number of query rows \cite{lin2021surveytransformers, RN2143}.
In practice, the effective token length also depends on the feature representation, but the dominant scaling in the number of rows remains quadratic \cite{PriorLabs2025TabPFN25}.
Advancements have been made to accelerate computations \cite{grinsztajn2026tabpfn3technicalreport}.


\subsection{Bayesian Optimization} \label{subsec bo_background} 
BO is a sample-efficient iterative strategy designed for the global optimization of expensive black-box objective functions \cite{RN2501}. The core philosophy of BO is to sequentially gather data and update a probabilistic belief about the objective's landscape to make increasingly improved guesses. This procedure relies on two main components: an emulator and an acquisition function (AF) that uses the emulator’s probabilistic predictions to determine the next optimal point for evaluation.




GPs have long been the predominant surrogates in BO since they effectively learn from small data and render many AFs analytic which, in turn, allows to optimize them via gradient-based solvers. However, this conventional approach requires retraining (typically via MLE) at every BO step. This repeated retraining incurs computational overhead and leaves the BO framework susceptible to the previously discussed brittleness of MLE. 

FMs such as TabPFN offer a fundamentally different operational paradigm for BO \cite{RN2144}. On the one hand, their in-context learning ability removes the need to train an emulator in each BO iteration; reducing computational overheads associated with selecting and training an emulator. On the other hand, their inference cost is high and the quantified prediction uncertainties do not result in analytic AFs; making it infeasible to optimize AFs with gradient-based solvers. To address this issue, BO with PFNs often relies on evaluating the AF across a massive pool of candidate points and then selecting the one with the highest AF.

\subsection{Classification via Gaussian Processes} \label{subsec classification gp}
In contrast to GP regression, discrete labels in classification necessitate a non-Gaussian likelihood (e.g., categorical or Bernoulli), for which exact posterior inference becomes intractable. Consequently, standard GP classifiers rely on computationally intensive approximate inference methods such as Laplace’s method, variational inference, or expectation propagation \cite{RN332}.
In this work, we leverage the Dirichlet target construction of Milios et al. \cite{milios2018dirichlet} who cast classification as a regression problem. As detailed in \cite{milios2018dirichlet}, in this approach discrete class labels are mapped to continuous regression targets through Dirichlet parameters, and GP regression is performed on these targets using a Gaussian likelihood, allowing for exact GP regression inference. The resulting predictions can then be transformed back to class probabilities, with the Dirichlet formulation offering a natural representation of uncertainty via its concentration parameters. These concentration parameters are determined by a single hyperparameter, $\alpha_\epsilon$, which dictates the degree of certainty assigned to each training label. We visualize the effect of $\alpha_\epsilon$ with a 1D example in \Cref{fig:1d_classification}.

\section{Mitigating the Brittleness of MLE} \label{sec method}
Multiple arguments indicate that MLE can produce sub-optimal ML models even in the case of GPs which are commonly (but incorrectly) assumed to be very robust to overfitting. The first argument relies on the asymptotic properties of MLE which hold in the case of large data. However, number of samples in many applications is finite and data density rapidly decreases as input dimensionality or $D_x$ increases. That is, the attractive asymptotic properties of MLE do not hold in many real world applications. 

Second, the optimization problem in \cref{eq mle} is highly non-convex and since state-of-the-art optimizers such as Adam and L-BFGS are gradient-based, it is very common to converge to local optima. While good local optima can improve generalization, optimizers may converge to bad local optima. In the case of GPs, such a behavior can especially affect the estimate of the nugget parameter which heavily controls the behavior of a GP.

The third argument concerns the objective of MLE: the sole purpose of MLE is to ensure that the selected model does its best in explaining the data. Hence, MLE does not directly indicate if the model class is selected optimally and generalizes well to unseen data. In GP modeling, the objective function of MLE does not provide any direct feedback that indicates whether the kernel is formulated optimally. For example, if the Gaussian kernel is selected (which encodes our assumption on the smoothness or infinite differentiability) while the underlying function is non-smooth, MLE aims to optimize the kernel hyperparameters such that the corresponding GP explains the data as good as it can, i.e., MLE will not indicate that the kernel is chosen poorly. Recent works typically address this issue by learning the kernel (e.g., deep kernels), combining them \cite{RN1901,RN2252}, or building GPs with different kernels and then selecting the best one. Our studies are based on the latter approach but we note that all these strategies are directly affected by the choice of the loss function which typically involves MLE.

It must be highlighted that the above three issues are interconnected. For example, the smoothness and non-convexity of $\mathcal{L}(\cdot)$ in \cref{eq mle} is directly related to the dataset size, as supported by the asymptotic properties of MLE and empirical studies \cite{RN940,RN649} where it is shown that the profile of $\mathcal{L}(\cdot)$ becomes increasingly convex in the case of GPs with Gaussian kernels. These empirical findings, however, do not readily transfer to more complex formulations such as deep kernels. 

To address the above issues associated with training GPs via MLE, we first discuss some alternative loss functions and their relations to MLE in \cref{subsec alternative loss}. Then, we discuss critical strategies for successfully implementing MLE in  \cref{subsec opt} to show that while MLE is perhaps not the best loss function, it can produce highly accurate GPs as long as it is implemented with care. 

\subsection{MLE vs Alternative Loss Functions} \label{subsec alternative loss}
To guard against incorrect modeling assumptions (e.g., selected kernel or noise model), it is often argued that variants of CV must be used in training since CV directly approximates the generalization error. 
Loss functions based on leave-one-out (LOO) log-predictive probability (aka log pseudo-likelihood) belong to this category and can be calculated analytically for GPs in one shot (rather than training $N$ independent models) via:
\begin{equation}
\begin{aligned}
    \mathcal{L}_{PL}(\boldsymbol{\beta},  \boldsymbol{\theta}, \sigma^2_{\varepsilon}) &= 
    \sum_{i=1}^{N} \ln p(y_i|\boldsymbol{x}, \boldsymbol{y}_{-i}; \boldsymbol{\beta},  \boldsymbol{\theta}, \sigma^2_{\varepsilon})\\
    &\propto \sum_{i=1}^{N} \ln{ [\boldsymbol{C}_{\varepsilon}^{-1}]_{ii} } - 
    \frac{[\boldsymbol{C}_{\varepsilon}^{-1}(\boldsymbol{y} - \boldsymbol{m})]^2_i}{[\boldsymbol{C}_{\varepsilon}^{-1}]_{ii}}
    \label{eq loocv}    
\end{aligned}
\end{equation}
where $\boldsymbol{y}_{-i}$ indicates all the observations except the $i^{th}$ one. In our experience, GPs trained with \cref{eq mle} tend to incur lower training times and generalization error than those trained via \cref{eq loocv}. We believe this trend is expected since \cref{eq loocv} is based on leaving only one data point out at a time. 

Another class of loss functions involves kernel flows, proposed in \cite{RN2238} based on the idea that the accuracy of a \textit{good} kernel-based interpolant should be minimally affected if half the training data is dropped. This concept can be formalized using the reproducing kernel Hilbert space (RKHS) associated with the kernel. Specifically, kernel flow objective $\rho(\boldsymbol{\beta},  \boldsymbol{\theta}, \sigma^2_{\varepsilon})$ is defined as:
\begin{equation}
    \rho(\boldsymbol{\beta},  \boldsymbol{\theta}, \sigma^2_{\varepsilon}) = 
    1 - \frac{(\boldsymbol{y}_s - \boldsymbol{m}_s)^T\boldsymbol{C}^{-1}_s(\boldsymbol{y}_s - \boldsymbol{m}_s)}
    {(\boldsymbol{y} - \boldsymbol{m})^T\boldsymbol{C}^{-1}(\boldsymbol{y} - \boldsymbol{m})},
    \label{eq rho}
\end{equation}
where variables with subscript $s$ are only calculated for a randomly selected subset of the data (typically half). Intuitively, the numerator quantifies the RKHS-induced error incurred by using the reduced dataset, while the denominator represents the corresponding quantity for the full dataset. A kernel that generalizes well should yield a small value of $\rho$. 

We believe the major advantage of loss functions based on \cref{eq rho} is that they accommodate batch-based training where in each iteration a batch of samples and half of it are used to calculate the denominator and numerator in $\rho$. However, kernel-flow-based loss functions only evaluate interpolation accuracy and lack UQ capabilities. Hence, we do not recommend such loss functions for tasks involving UQ unless they are combined with other losses such as \cref{eq mle}.

An effective strategy for improving MLE involves its penalization using hyperparameter priors---giving rise to maximum a posteriori (MAP) estimates. However, it is not straightforward to specify generalizable priors. Yet another strategy involves regularization based on average negatively oriented interval score (NIS). The interval score (IS) \cite{Gneiting01032007} for a central \((1-\alpha)\) prediction interval \([L_i, U_i]\) is defined as:
\begin{equation}
  \mathrm{IS}_i
  = (U_i - L_i)
  + \frac{2}{\alpha}\, (L_i - y_i)\,\mathbf{1}[y_i < L_i]
  + \frac{2}{\alpha}\, (y_i - U_i)\,\mathbf{1}[y_i > U_i],
  \label{eq interval score}
\end{equation}
where the indicator function $\mathbf{1}[\cdot]$ is used for the two terms that penalize observations that lie outside the width $U_i - L_i$ and $\alpha \in (0,1)$, e.g. $\alpha = 0.05$ yields a 95\% interval. From \cref{eq interval score}, NIS is computed as:
\begin{equation}
    \mathrm{NIS}
    = \frac{1}{c}\, \frac{1}{N} \sum_{i=1}^{N} \mathrm{IS}_i,
    \label{eq nis}
\end{equation}
where $c$ is a constant that can be adjusted so that NIS ranges across different problems are directly comparable (e.g., based on the standard deviation of the observations). 
In our studies, we use \cref{eq nis} to assess test-time UQ capabilities and hence do not use it during training. 

Another widely used proper scoring rule for probabilistic forecasts is the continuous ranked probability score (CRPS), which evaluates the full predictive cumulative distribution function (CDF) rather than a single prediction interval \cite{Gneiting01032007}. For a test observation $y_i$ and predictive CDF $F_i(\cdot)$, the per-point CRPS is defined as:
\begin{equation}
  \mathrm{CRPS}_i
  = \int_{-\infty}^{\infty}
    \bigl[ F_i(z) - \mathbf{1}[z \ge y_i] \bigr]^2 \, dz,
  \label{eq crps}
\end{equation}
where $\mathbf{1}[\cdot]$ is the indicator function. Intuitively, \cref{eq crps} measures the integrated squared distance between the predicted CDF and the empirical CDF of the observation; lower values indicate better calibrated and sharper predictive distributions. Because GPs yield Gaussian predictive distributions $\mathcal{N}(\mu_i, \sigma_i^2)$ at test inputs, \cref{eq crps} admits the closed-form expression:
\begin{equation}
  \mathrm{CRPS}_i
  = \sigma_i \left[
    z_i \bigl( 2\Phi(z_i) - 1 \bigr)
    + 2\phi(z_i)
    - \frac{1}{\sqrt{\pi}}
  \right],
  \quad
  z_i = \frac{y_i - \mu_i}{\sigma_i},
  \label{eq crps gaussian}
\end{equation}
where $\Phi(\cdot)$ and $\phi(\cdot)$ denote the standard normal CDF and PDF, respectively. The average CRPS over $N$ test points is then computed as:
\begin{equation}
    \mathrm{NCRPS}
    = \frac{1}{c}\,\frac{1}{N}\sum_{i=1}^{N}\mathrm{CRPS}_i.
    \label{eq ncrps}
\end{equation}
As with IS, we normalize CRPS by a problem-dependent constant $c$ (e.g., the standard deviation of the test observations) to obtain a comparable metric across benchmarks. In our studies, we use \cref{eq ncrps} to assess test-time UQ capabilities and do not use it during training. For non-Gaussian predictive distributions, such as TabPFN's bar distribution, \cref{eq crps} is evaluated using the predictive CDF or samples from the predictive distribution.

\subsection{Efficient Gradient-based Optimization} \label{subsec opt}
The profile of $\mathcal{L}(\cdot)$ often has many local optimal and flat regions; rendering gradient-based optimizers highly inefficient and sensitive to initialization. To simultaneously address these issues, we strongly recommend the adoption of change of variables in kernel formulations such that $\mathcal{L}(\cdot)$ has large gradients with respect to the hyperparameters; increasing the rate of convergence of the optimizer while helping to identify the global (or good local) optima. Additionally, appropriate reformations can remove constraints on the hyperparameters. For instance, the lengthscales $\boldsymbol{l}$ in \cref{eq gaussian typical} cannot be zero and they also suffer from identifiability issues due to the squared distance term. Hence, we propose to represent the hyperparameters in the log-scale as:
\begin{equation}
    c(\boldsymbol{x}, \boldsymbol{x}'; \boldsymbol{\theta}) = 10^{s} \exp\left(-\sum_{i=1}^{D_x} 10^{\omega_i}(x_i - x_i')^2 \right),
\label{eq gaussian gp+}
\end{equation}
where $\boldsymbol{\theta} = \{ s, \boldsymbol{\omega}\}$. 

Even though $c(\boldsymbol{x}, \boldsymbol{x}'; \boldsymbol{\theta})$ in \cref{eq gaussian gp+} is a theoretically valid kernel for unbounded $s$ and $\boldsymbol{\omega}$, in practice we bound these variables to ensure numerical stability of the kernel evaluations throughout the entire optimization process. To this end, rather than leveraging optimizers such as L-BFGS-B that accept bound constraints, we introduce a \textit{soft} clamp function. Specifically, for the hyperparameter $\theta_i \in [a_i, b_i]$ our soft clamp is formulated as:
\begin{equation}
    \theta_i= 
    \begin{cases}
    a_i + \delta \exp\left(\dfrac{\theta_i - (a_i + \delta)}{\delta}\right) & \text{if } \theta_i < a_i + \delta \\[2em]
    \theta_i & \text{if } a_i + \delta \leq \theta_i \leq b_i - \delta \\[1em]
    b_i - \delta \exp\left(-\dfrac{\theta_i - (b_i - \delta)}{\delta}\right) & \text{if } \theta_i > b_i - \delta
    \end{cases}
\label{eq soft clamp}
\end{equation}
where $\delta$ is the transformation margin (e.g. $\num{1e-2}$). Unlike a regular clamp, \cref{eq soft clamp} does not cause oscillations or non-smooth behavior near the boundaries.
In our GP+ package, the hyperparameters corresponding to the kernel in \cref{eq gaussian gp+} are bounded as $s \in [-5, 4]$ and $\omega_i \in [-6, 3]$. 

The above approach is directly applicable to other widely used kernels. For instance, for the power exponential (PE) kernel $p \in [1, 2]$:
\begin{equation}
    c(\boldsymbol{x}, \boldsymbol{x}'; \boldsymbol{\theta}) = 10^{s} \exp\left(-\sum_{i=1}^{D_x} 10^{\omega_i}|x_i - x_i'|^p \right).
\label{eq pe gp+}
\end{equation}

Finally, to prevent convergence to bad local optima, we recommend initializing the hyperparameters via space-filling designs and selecting the ones with the best loss. In GP+, we use $16$ initializations by default which can be adjusted to balance training costs and chances of finding a good set of hyperparameters. 

    \section{Results and Discussions} \label{sec results}
After detailing our evaluation protocols, we present results for simple 1D examples (\Cref{subsec 1d examples}), followed by regression, BO, and classification tasks (\Cref{subsec results regression,subsec results bo,subsec results classification}). While representative tables and figures are included in the main text, comprehensive results and all materials required for reproducibility are available in a dedicated GitHub repository \cite{GithubRepo}.

\subsection{Evaluation Protocols} \label{subsec evaluation protocols}
We compare GPs against TabPFN v2.0 and v2.5 on regression, BO, and classification tasks in \cref{subsec results regression,subsec results classification,subsec results bo}, respectively. To stress-test GPs, we train them via the default settings of Gpytorch \cite{RN6219} and GP+ \cite{yousefpourGP2024}. All GPs use CPU while TabPFN models require a GPU. L-BFGS and Adam are used for GP hyperparameter optimization in regression and classification, respectively. 

Nine benchmark regression problems are adopted from \cite{simulationlib} (i.e. Ackley, Borehole, Dixon-Price, Griewank, Rosenbrock, Wing Weight, and Zakharov) and we study the effects of categorical inputs in the Buckling function obtained from \cite{RN2505}. 
Across our experiments, $D_x \in \{4, 10, 20, 40, 80\}$ and all datasets have a single response since this is a limitation of TabPFN. We analyze the effect of (1) dataset size by setting $N = 5D_x$ and $N=20D_x$, and (2) noise variance by selecting two noise levels, $\varepsilon \sim \mathcal{N}(0, (0.002\times c)^2)$ and $\varepsilon \sim \mathcal{N}(0, (0.08\times c)^2)$, where $c$ is a constant that depends on the response range in each problem.

We repeat each experiment $10$ times and each time assess the performance using $N_{test}=5000$ test samples. 
Training and prediction times are recorded for comparison. 
We use relative root mean-squared error (RRMSE) to measure prediction accuracy:
\begin{equation}
  \mathrm{RRMSE}
  = \frac{1}{c} \sqrt{\frac{1}{N_{test}} \sum_{i=1}^{N_{test}} (y_i - \hat{y}_i)^2},
  \label{eq rrmse}
\end{equation}
where $y_i$ and $\hat{y}_i$ denote the true and predicted values.
To compare predictive distributions, we use \cref{eq nis}.

For the classification benchmarks, we consider four problems: Electrical Grid Stability Simulated data \cite{electrical_grid_stability_simulated_data__471}, Truss 6D data~\cite{truss_6d}, Stellar Classification (SDSS17)~\cite{stellar_sdss17}, and Steel Plates Faults~\cite{steel_plates_faults_198}. The first is a binary classification problem where the target indicates whether the system is stable, with $D_x=11$ consisting entirely of continuous features. Truss 6D is a binary problem where the target indicates whether a structural truss design is valid, with $D_x=6$ consisting of three continuous and three categorical features. Stellar Classification is a three-class problem distinguishing stars, galaxies, and quasars based on their spectral characterisitics, with $D_x=8$ continuous features. Steel Plates Faults is a seven-class problem identifying surface fault types, with $D_x=26$ consisting of 25 continuous features and one binary feature.The categorical features are one-hot encoded prior to GP+ training. 
We evaluate nested training subsets drawn from a fixed stratified training pool, with a test set containing the remainder of the data. For Electrical Grid Stability and Truss 6D we use $N \in \{50, 100, 200, 300\}$. Steel Plates Faults is limited to $N \in \{50, 100, 200\}$, as its seven classes require fitting a separate latent function per class and training cost grows accordingly. For Stellar Classification we instead use $N \in \{5, 10, 20, 30\}$, as accuracy and ECE have already plateaued for all models at the larger sizes used for the other datasets.

We use accuracy and expected calibration error (ECE) to compare the classifiers. Let $\hat{c}_i = \max_k \hat{p}_{ik}$ denote the predicted confidence for observation $i$. For any bin $B$ of the test set, define:
\begin{equation}
    \mathrm{Acc}(B)
    = \frac{1}{|B|} \sum_{i \in B}
    \mathbf{1}[y_i = \hat{y}_i],
    \label{eq accuracy}
\end{equation}
where $\hat{y}_i$ is the predicted class label for observation $i$. Test accuracy is $\mathrm{Acc}(B_{\mathrm{test}})$ where $B_{\mathrm{test}}$ is the full test set. ECE is computed as:
\begin{equation}
    \mathrm{ECE}
    = \sum_{m=1}^{M} \frac{|B_m|}{N_{\mathrm{test}}}
    \left|
        \mathrm{Acc}(B_m) \;-\; \mathrm{Conf}(B_m)
    \right|,
    \label{eq ece}
\end{equation}

\begin{figure*}[t]
    \centering
    \captionsetup{font=footnotesize}
    \includegraphics[width=0.95\textwidth]{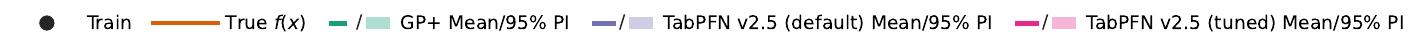}\\[0.35em]
    \begin{minipage}[b]{0.185\textwidth}
        \centering
        \includegraphics[width=\textwidth]{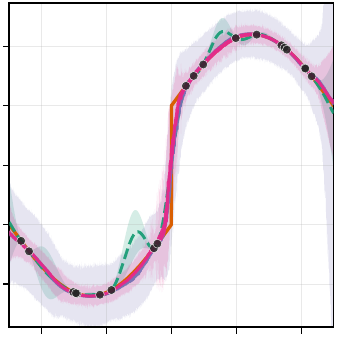}
        \subcaption{Discontinuous Sine}
        \label{fig:1d_examples_discontinuity}
    \end{minipage}
    \hfill
    \begin{minipage}[b]{0.185\textwidth}
        \centering
        \includegraphics[width=\textwidth]{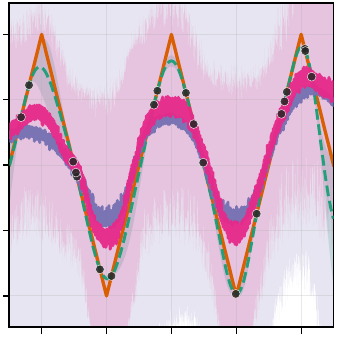}
        \subcaption{Triangle Wave}
        \label{fig:1d_examples_triangle_wave}
    \end{minipage}
    \hfill
    \begin{minipage}[b]{0.185\textwidth}
        \centering
        \includegraphics[width=\textwidth]{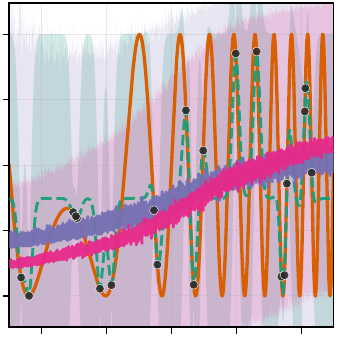}
        \subcaption{Non-Stationary Chirp}
        \label{fig:1d_examples_chirp}
    \end{minipage}
    \hfill
    \begin{minipage}[b]{0.185\textwidth}
        \centering
        \includegraphics[width=\textwidth]{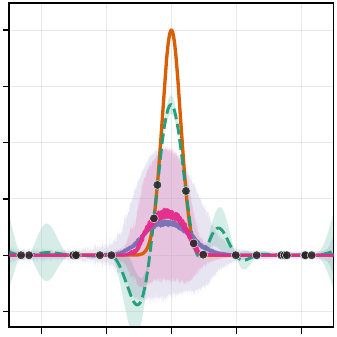}
        \subcaption{Localized Bump}
        \label{fig:1d_examples_localized_bump}
    \end{minipage}
    \hfill
    \begin{minipage}[b]{0.185\textwidth}
        \centering
        \includegraphics[width=\textwidth]{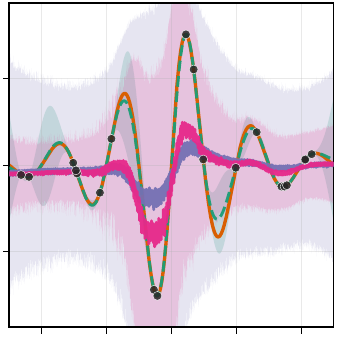}
        \subcaption{Damped Sine}
        \label{fig:1d_examples_damped_sine}
    \end{minipage}
    \caption{1D regression examples with $n{=}20$ training points: Mean predictions with 95\% prediction intervals are shown. Axes span $x\in[-0.5,0.5]$, with tick labels omitted for compactness. The defining equations for each example are given in the project repository~\cite{GithubRepo}.}
    \label{fig:1d_examples}
\end{figure*}

\noindent where $\mathrm{Conf}(B_m) = \frac{1}{|B_m|}\sum_{i \in B_m} \hat{c}_i$ is the mean confidence in bin $m$, and $B_m = \left\{i : \frac{m-1}M < \hat{c}_i \leq \frac{m}{M}\right\}$ defines $M = 10$ equal-width bins grouping the test set. Lower values of $\mathrm{ECE}$ signify better-calibrated predictions, reflecting closer agreement between confidence and accuracy within each bin.

\begin{figure*}[b]
    \centering
    \captionsetup{font=footnotesize}
    \includegraphics[width=0.72\textwidth]{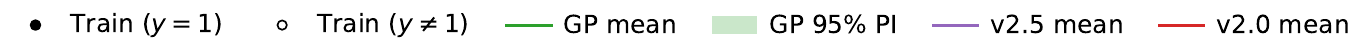}\\[0.35em]
    \begin{minipage}[b]{0.23\textwidth}
        \centering
        \includegraphics[width=\textwidth]{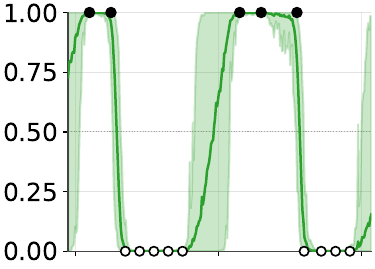}
        \subcaption{$\alpha_\epsilon = 0.001$}
        \label{fig:1d_classification_alpha_0p001}
    \end{minipage}
    \hfill
    \begin{minipage}[b]{0.23\textwidth}
        \centering
        \includegraphics[width=\textwidth]{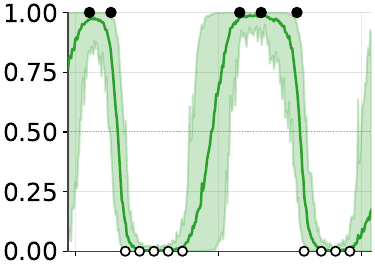}
        \subcaption{$\alpha_\epsilon = 0.01$}
        \label{fig:1d_classification_alpha_0p01}
    \end{minipage}
    \hfill
    \begin{minipage}[b]{0.23\textwidth}
        \centering
        \includegraphics[width=\textwidth]{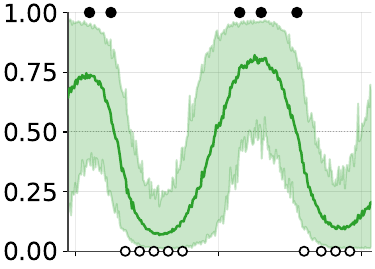}
        \subcaption{$\alpha_\epsilon = 0.1$}
        \label{fig:1d_classification_alpha_0p1}
    \end{minipage}
    \hfill
    \begin{minipage}[b]{0.23\textwidth}
        \centering
        \includegraphics[width=\textwidth]{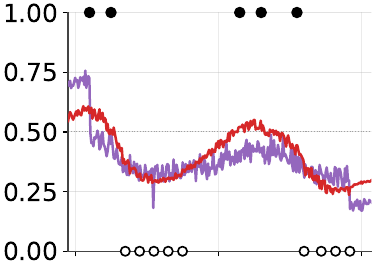}
        \subcaption{TabPFN}
        \label{fig:1d_classification_tabpfn}
    \end{minipage}
\caption{One-dimensional classification examples following Milios et al.~\cite{milios2018dirichlet}, with two classes and gaps in the training data. Curves give the predicted probability $P(y{=}1\mid x)$ and shaded regions the 95\% interval; filled markers denote training points of class~1 and open markers those of class~0. Class~0 probabilities are the complement of those shown and are omitted. Vertical axes show $P(y{=}1\mid x)$ and horizontal axes span $x\in[-1,1]$, with $x$ tick labels omitted for compactness.}
    \label{fig:1d_classification}
\end{figure*}

\subsection{One-Dimensional Examples} \label{subsec 1d examples}
\Cref{fig:1d_examples} shows five noise-free regression problems with $n{=}20$ training points, each stressing a different modeling challenge.
GP+ is used with the library defaults while TabPFN~v2.5 is shown both under its default settings and under a tuned configuration that uses the published ``small-samples'' regressor checkpoint in place of the default, together with an increased ensemble size ($n_{\mathrm{estimators}}{=}16$ vs.\ $8$) and a lower softmax temperature ($T{=}0.45$ vs.\ $0.9$) to sharpen predictive confidence. Because no observation noise is added to the training data, predictive means and intervals that concentrate more tightly around the true $f(x)$ are preferable. 

Across these examples, TabPFN appears more inclined to treat local structure as noise; its mean is often smoother than the underlying function, and its 95\% prediction intervals (PIs) are substantially wider than those of GP+. This conservative uncertainty can potentially be advantageous near discontinuities (\cref{fig:1d_examples_discontinuity}), where TabPFN avoids committing too strongly across the jump, but it also leads TabPFN to underfit oscillatory or sharply localized behavior in \Cref{fig:1d_examples_triangle_wave,fig:1d_examples_chirp,fig:1d_examples_localized_bump,fig:1d_examples_damped_sine} where the predictive mean regresses toward a flatter trend and the interval blankets much of the response range. GP+, by contrast, is more strongly influenced by the sparse observations; it typically tracks the training points closely and recovers sharp local features where data is present. However, these GPs use the smooth and stationary kernel in \Cref{eq gaussian gp+} so they fail to capture discontinuities and non-stationarity. 

In \Cref{fig:1d_classification}, we revisit a 1D classification example from \cite{milios2018dirichlet} where the domain contains 14 points from two classes with deliberate gaps to test the model's UQ in data-free regions. GP+ is used to build two heteroscedastic GPs (one per each class) where both use a Gaussian kernel whose hyperparameters are shared with the exception of the nugget that is class-specific. \Cref{fig:1d_classification_alpha_0p001,fig:1d_classification_alpha_0p01,fig:1d_classification_alpha_0p1}
sweep $\alpha_\epsilon \in \{0.001, 0.01, 0.1\}$, with the shaded region showing a 95\% PI obtained via Monte Carlo integration of the predictive distribution while \Cref{fig:1d_classification_tabpfn} shows TabPFN v2.0 and v2.5 for comparison where only $P(y=1 \mid x)$ is plotted; the class-0 probability is its complement.

Comparing \cref{fig:1d_classification_alpha_0p001,fig:1d_classification_alpha_0p01,fig:1d_classification_alpha_0p1} illustrates a clear trend between the magnitude of $\alpha_\epsilon$ and the posterior distribution: At $\alpha_\epsilon = 0.001$ and $0.01$, the posterior
mean sharply follows class labels, with predictive uncertainty increasing predominantly in the two gaps where data is absent. At $\alpha_\epsilon = 0.1$, however, the posterior mean becomes smoother and probabilities lean more towards an intermediate value. Effectively, $\alpha_\epsilon$ dictates how much class labels should be trusted, assigning greater uncertainty to these labels for higher values of this hyperparameter.

TabPFN in \cref{fig:1d_classification_tabpfn} offers a strong contrast to the behavior of the Dirichlet-based GP in this example. Despite using the same training points, both versions of TabPFN never approach high confidence in either class, even in regions with same-label point clusters present. It is important to note that TabPFN does not yet provide a predictive posterior distribution for classification, and hence its intervals are absent. 

\subsection{Regression Benchmarks} \label{subsec results regression}

\begin{figure*}[hptb]
    \centering
    \captionsetup{font=footnotesize}
    \begin{minipage}[b]{0.48\textwidth}
        \centering
        \includegraphics[width=\textwidth]{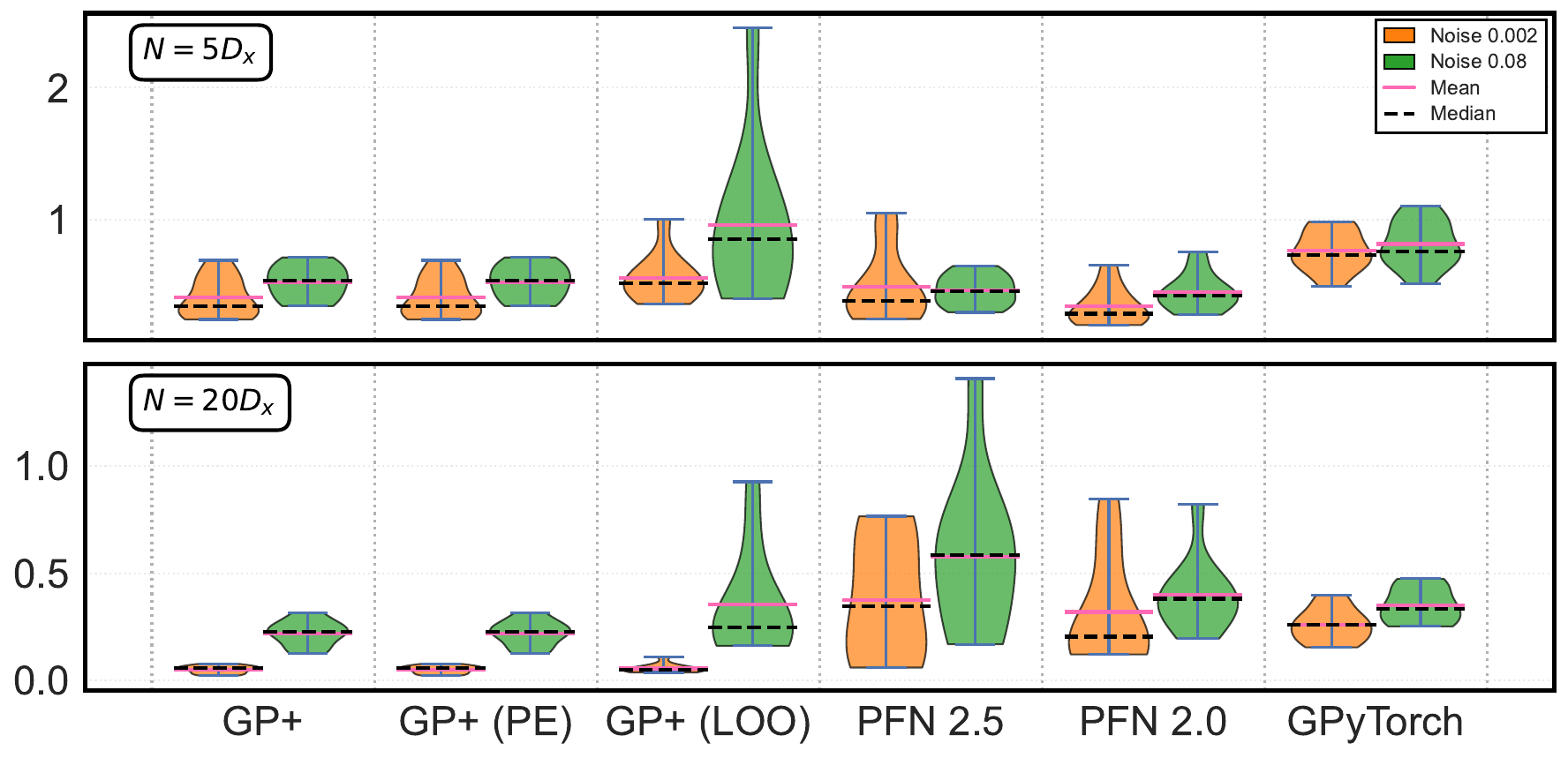}
        \subcaption{Buckling: D$_x$=4}
        \label{fig loo rrrmse buckling}
    \end{minipage}
    \hfill
    \begin{minipage}[b]{0.48\textwidth}
        \centering
        \includegraphics[width=\textwidth]{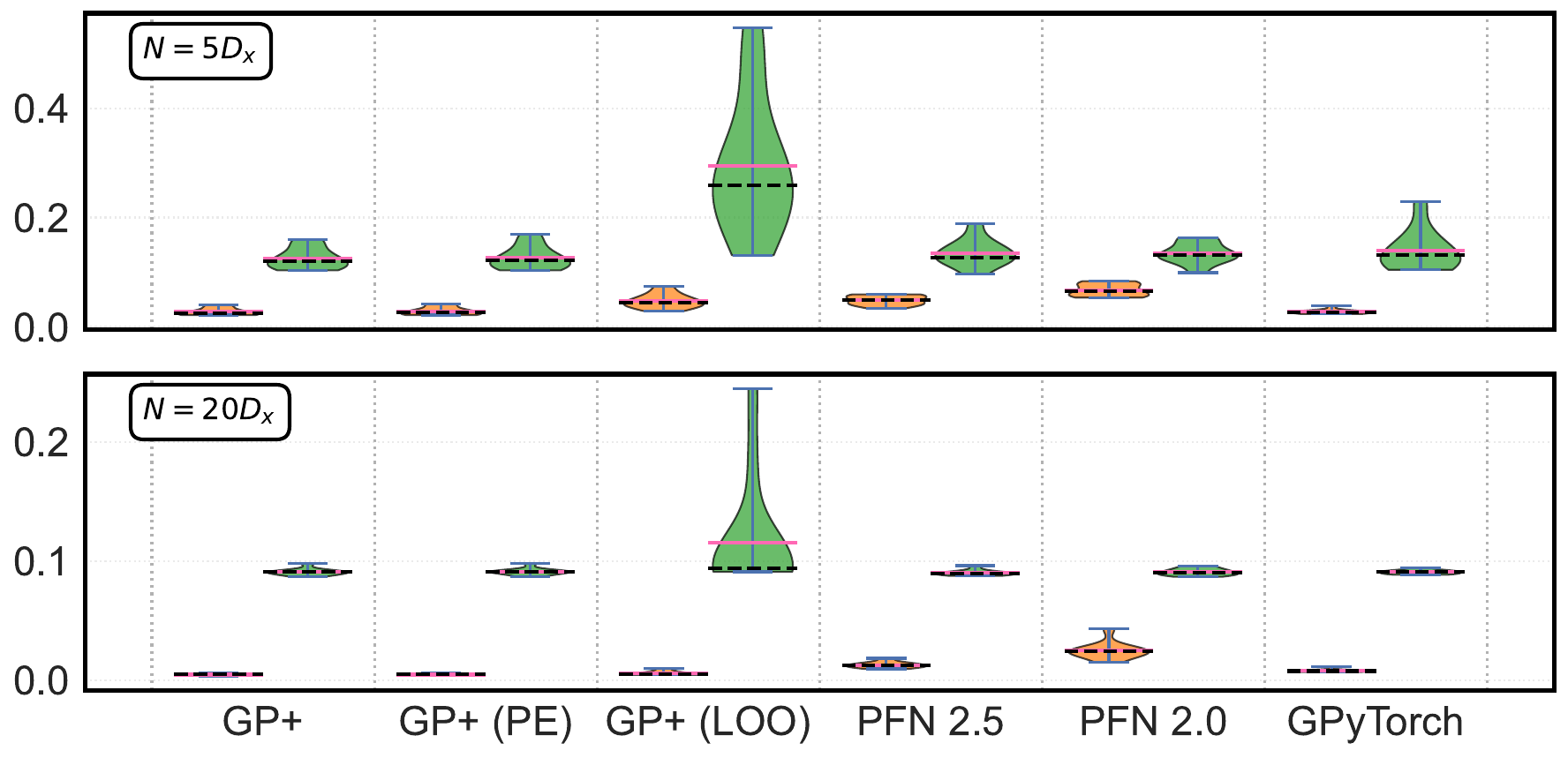}
        \subcaption{Borehole: D$_x$=8}
        \label{fig loo rrrmse borehole}
    \end{minipage}
    
    \vspace{1em}
    
    \begin{minipage}[b]{0.48\textwidth}
        \centering
        \includegraphics[width=\textwidth]{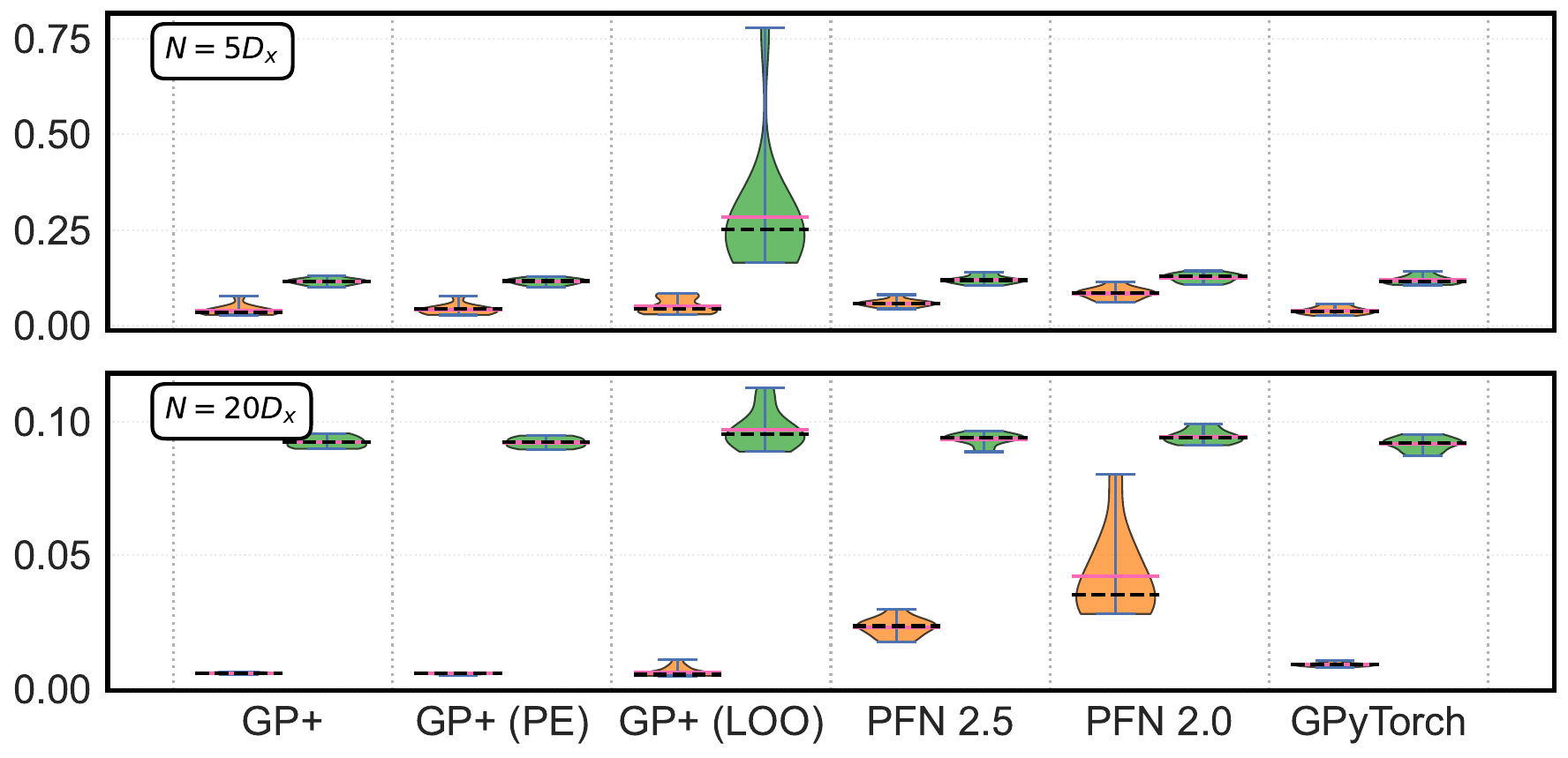}
        \subcaption{Wing Weight: D$_x$=10}
        \label{fig loo rrrmse wing}
    \end{minipage}
    \hfill
    \begin{minipage}[b]{0.48\textwidth}
        \centering
        \includegraphics[width=\textwidth]{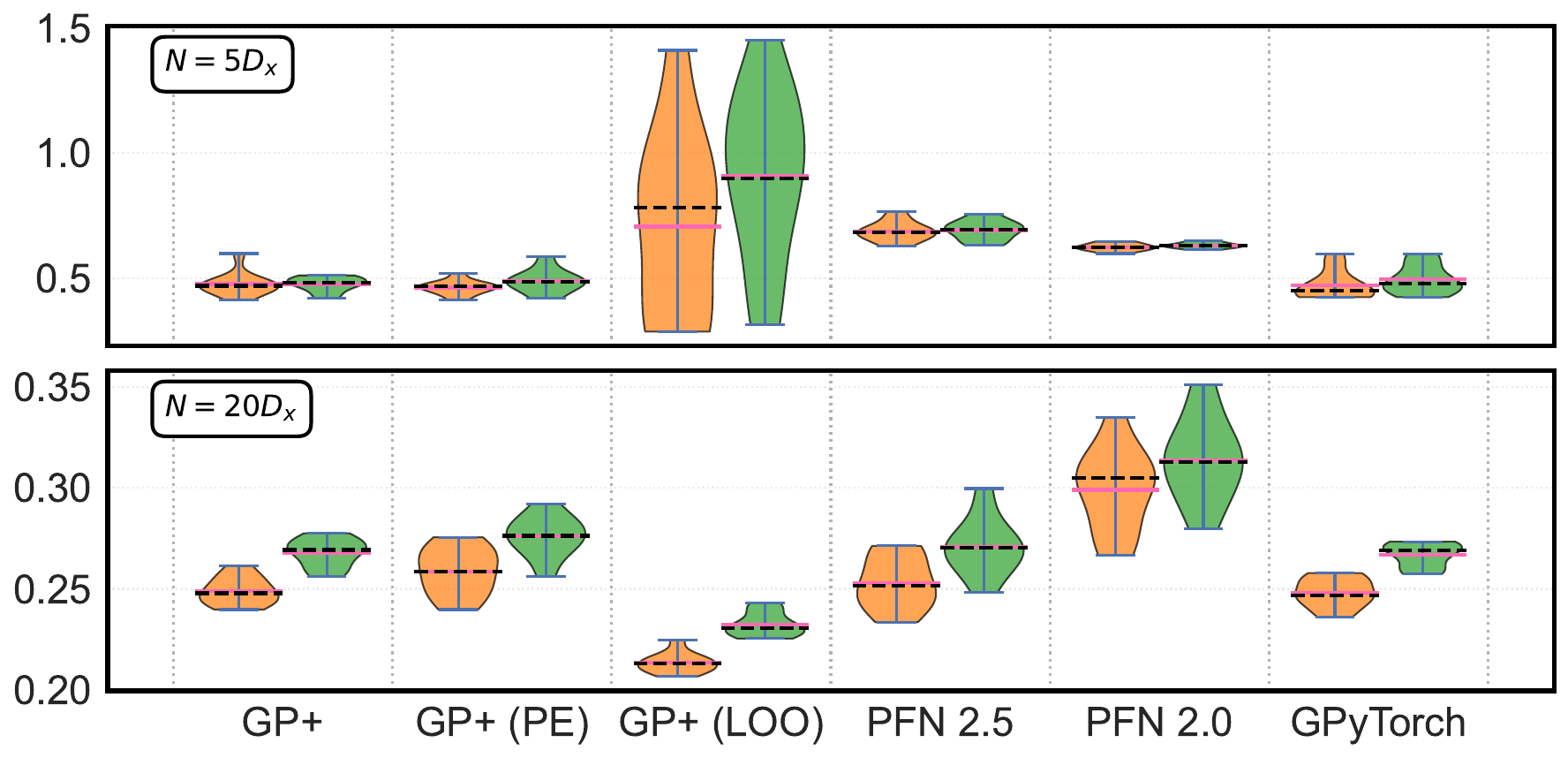}
        \subcaption{Ackley: D$_x$=20}
        \label{fig loo rrrmse ackley20}
    \end{minipage}

    \vspace{1em}
    
    \begin{minipage}[b]{0.48\textwidth}
        \centering
        \includegraphics[width=\textwidth]{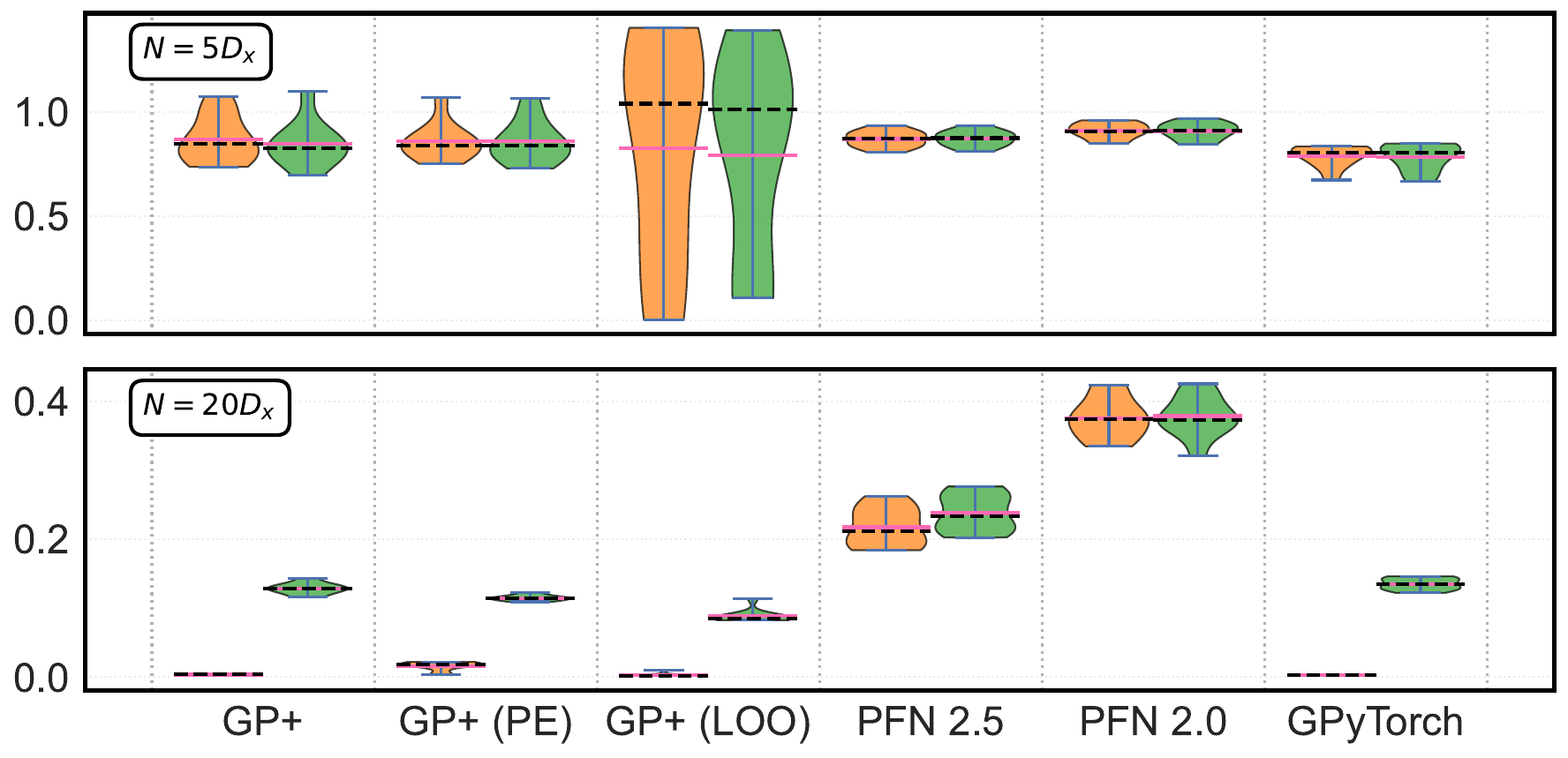}
        \subcaption{Griewank: D$_x$=20}
        \label{fig loo rrrmse griewank}
    \end{minipage}
    \hfill
    \begin{minipage}[b]{0.48\textwidth}
        \centering
        \includegraphics[width=\textwidth]{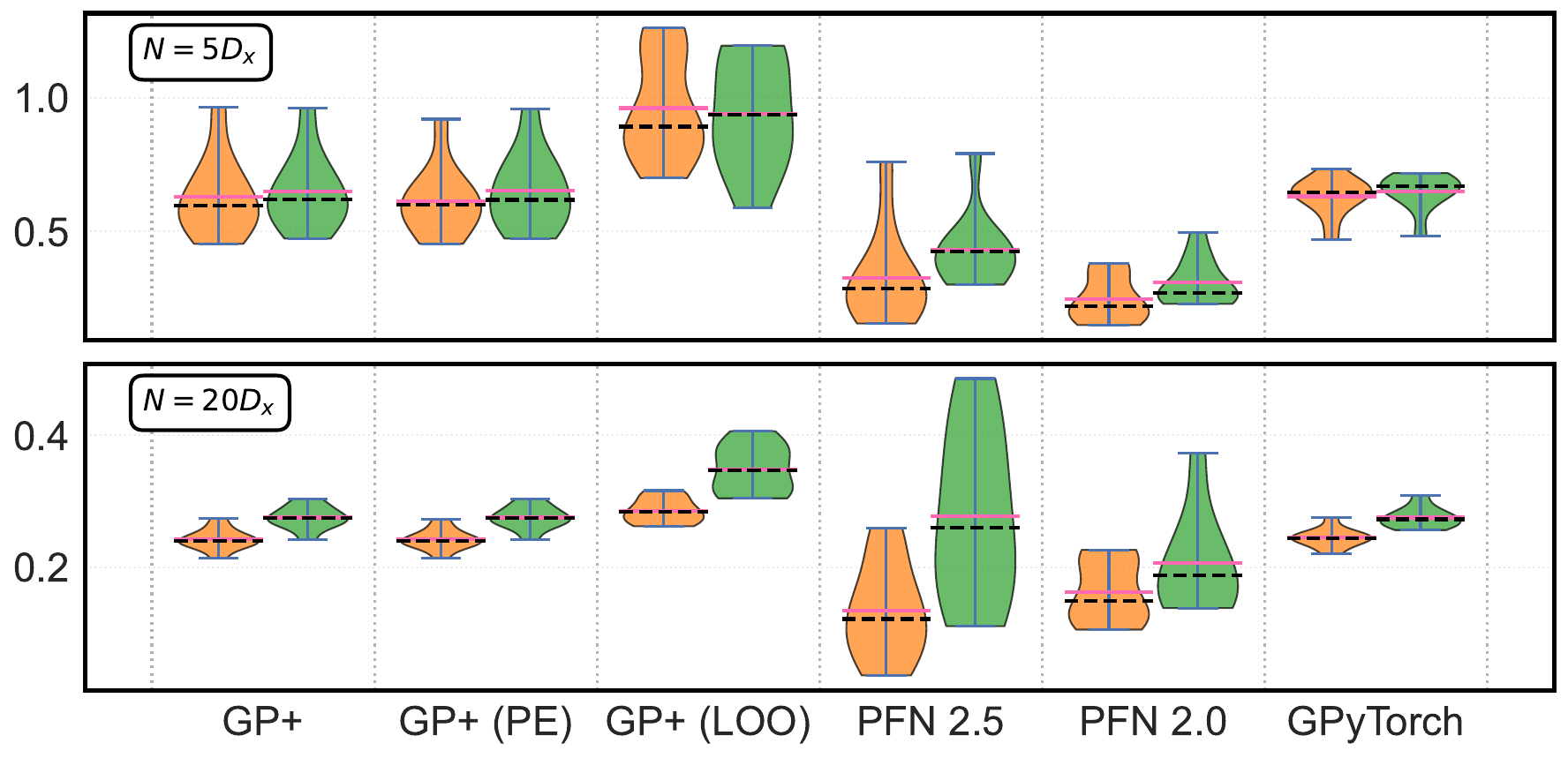}
        \subcaption{Zakharov: D$_x$=20}
        \label{fig loo rrrmse zakharov}
    \end{minipage}

    \vspace{1em}
    
    \begin{minipage}[b]{0.48\textwidth}
        \centering
        \includegraphics[width=\textwidth]{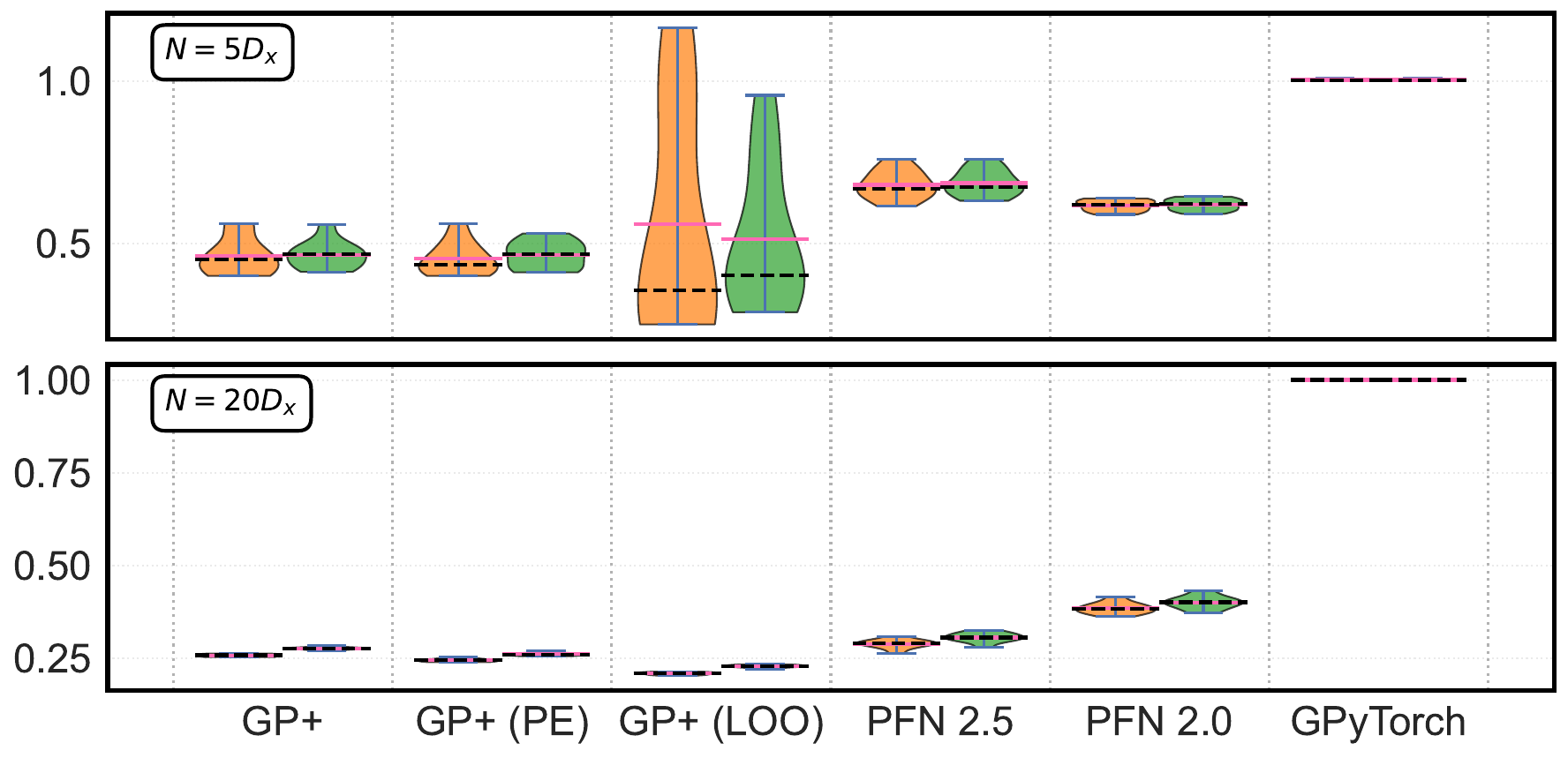}
        \subcaption{Ackley: D$_x$=40}
        \label{fig loo rrrmse ackley40}
    \end{minipage}
    \hfill
    \begin{minipage}[b]{0.48\textwidth}
        \centering
        \includegraphics[width=\textwidth]{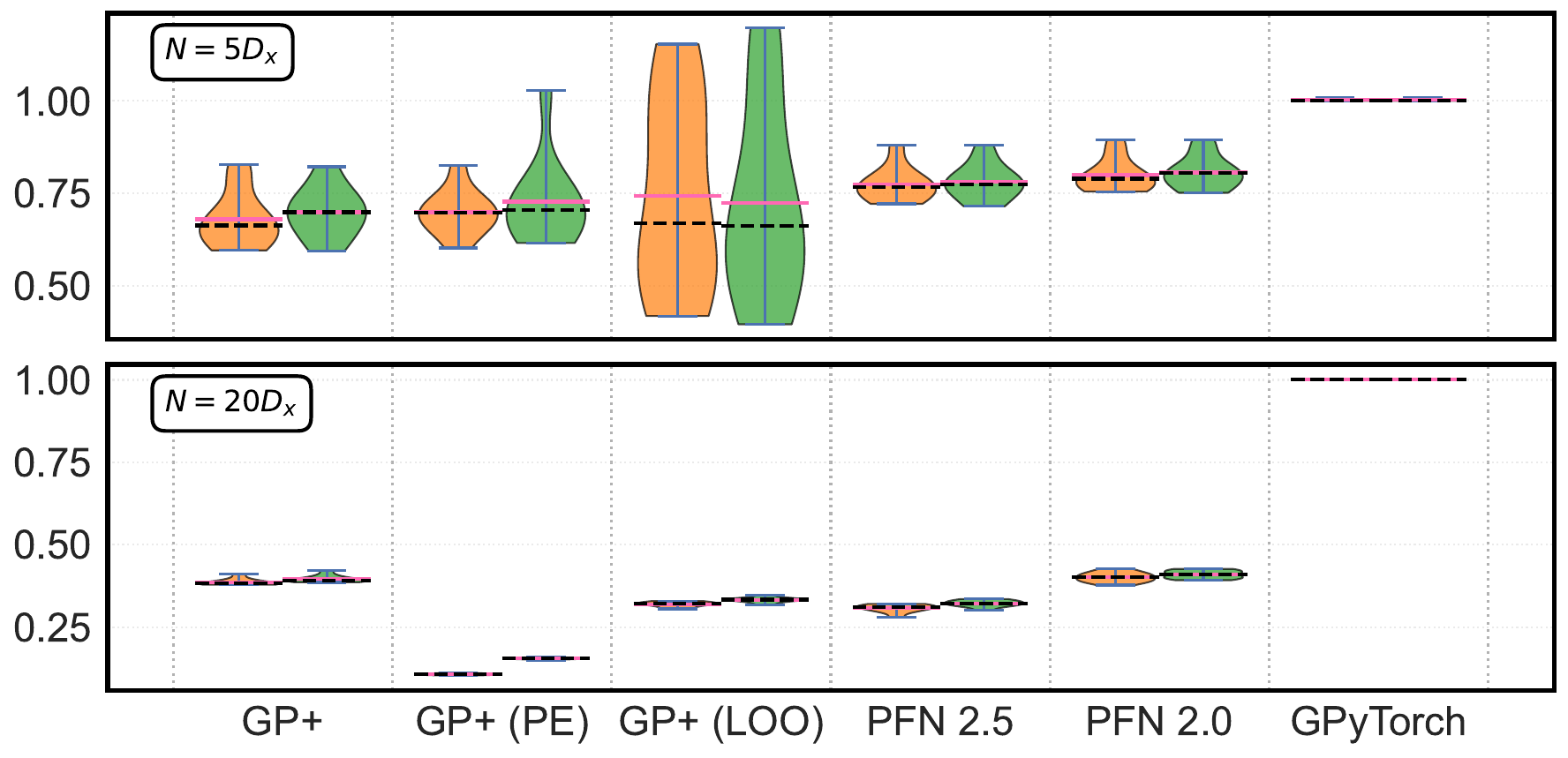}
        \subcaption{Dixon-Price: D$_x$=40}
        \label{fig loo rrrmse dixon}
    \end{minipage}
    
    \caption{Evaluations based on RRMSE in \Cref{eq rrmse}: Noise is either $\varepsilon \sim \mathcal{N}(0, (0.002\times c)^2)$ or $\varepsilon \sim \mathcal{N}(0, (0.08\times c)^2)$. GPs are used on CPU while TabPFN is used on GPU. All the GPs are trained with the \textit{default} settings of GP+ and GPyTorch packages. GP+ (PE) and GP+ (LOO) denote cases where the default Gaussian kernel and loss function are switched to, respectively, the PE kernel and LOO-based loss in \Cref{eq loocv}.}

    \label{fig results loo rrrmse}
\end{figure*}
\begin{figure*}[hptb]
    \centering
    \captionsetup{font=footnotesize}
    \begin{minipage}[b]{0.48\textwidth}
        \centering
        \includegraphics[width=\textwidth]{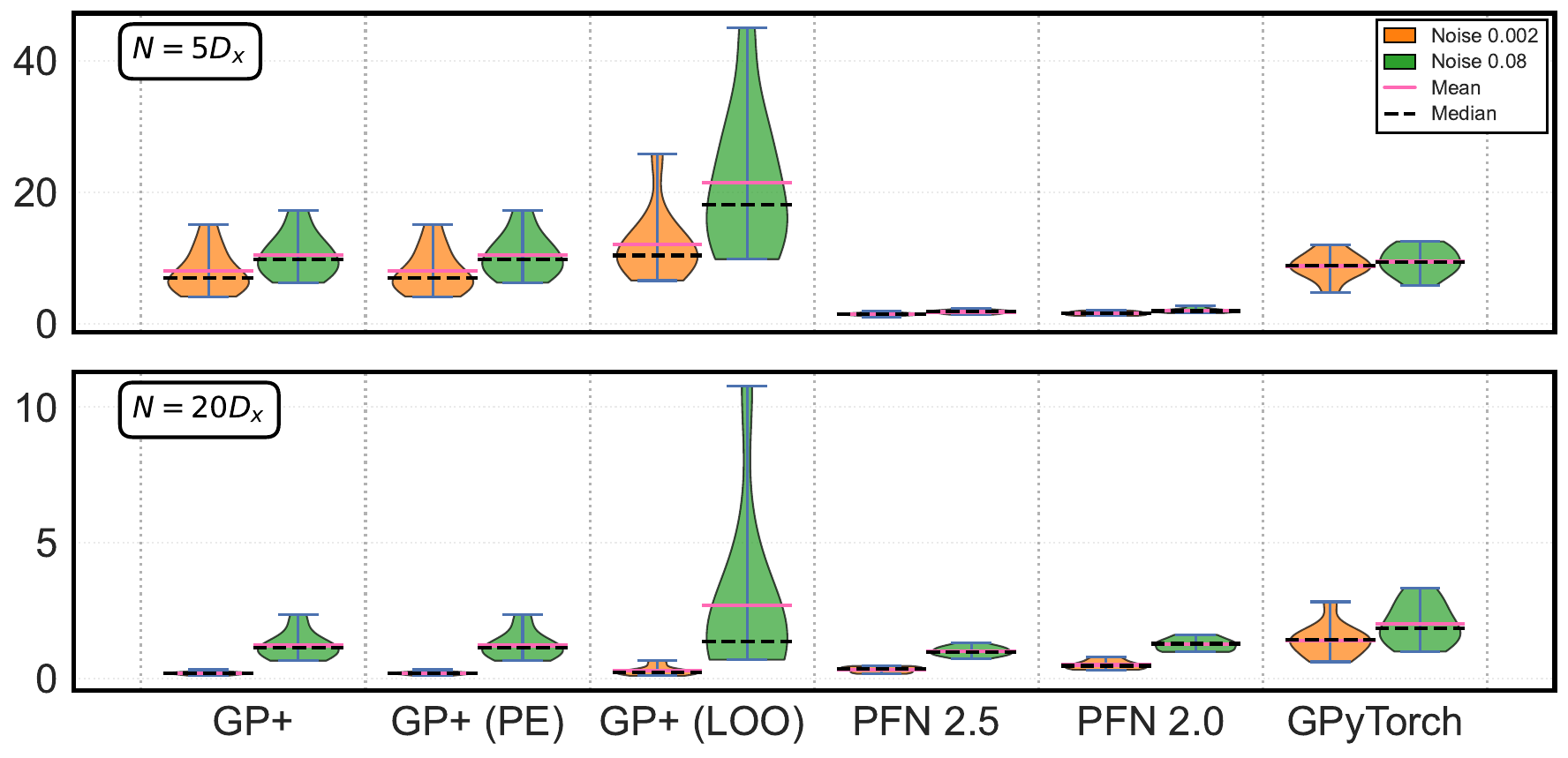}
        \subcaption{Buckling: D$_x$=4}
        \label{fig loo nis buckling}
    \end{minipage}
    \hfill
    \begin{minipage}[b]{0.48\textwidth}
        \centering
        \includegraphics[width=\textwidth]{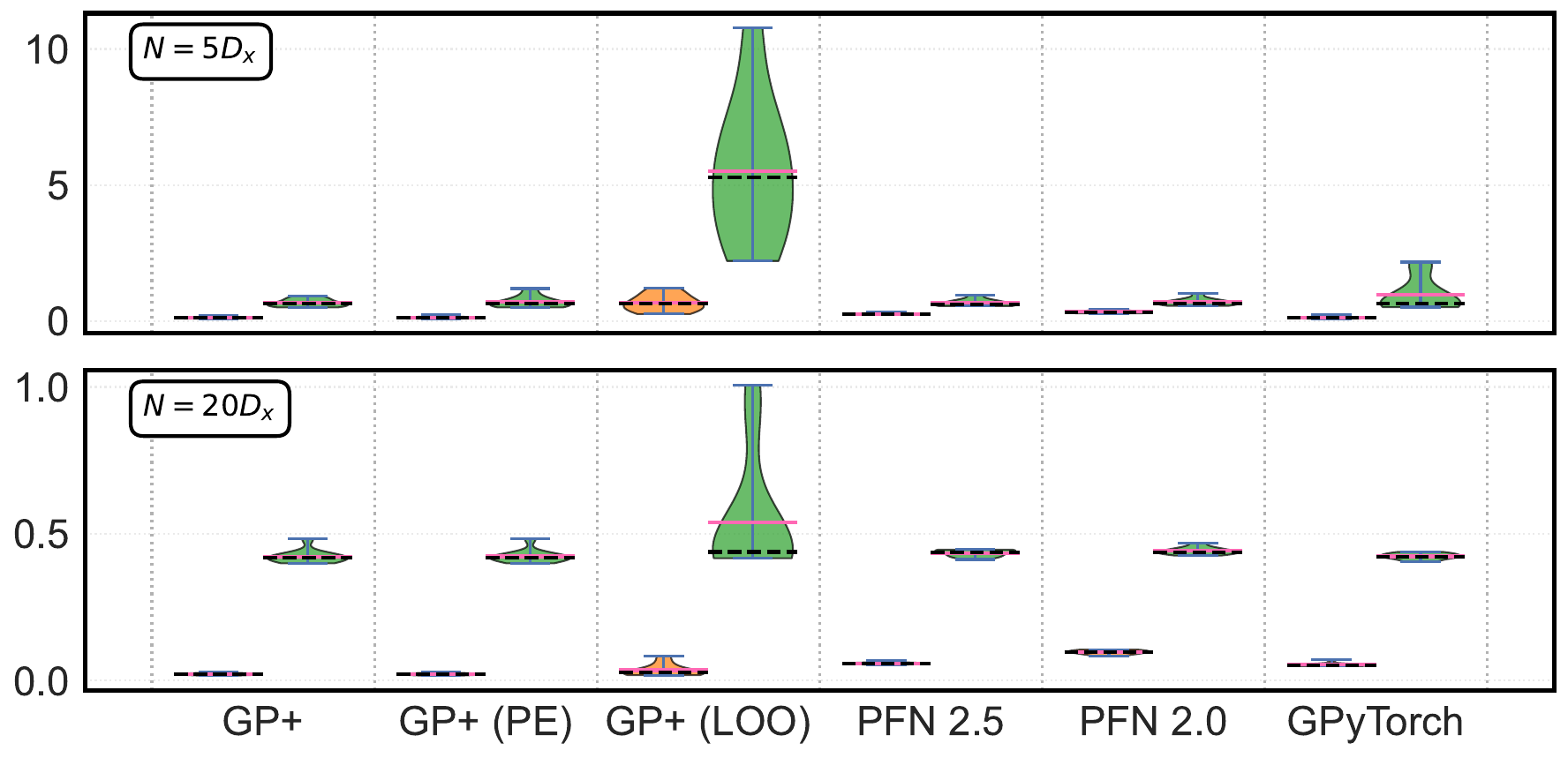}
        \subcaption{Borehole: D$_x$=8}
        \label{fig loo nis borehole}
    \end{minipage}
    
    \vspace{1em}
    
    \begin{minipage}[b]{0.48\textwidth}
        \centering
        \includegraphics[width=\textwidth]{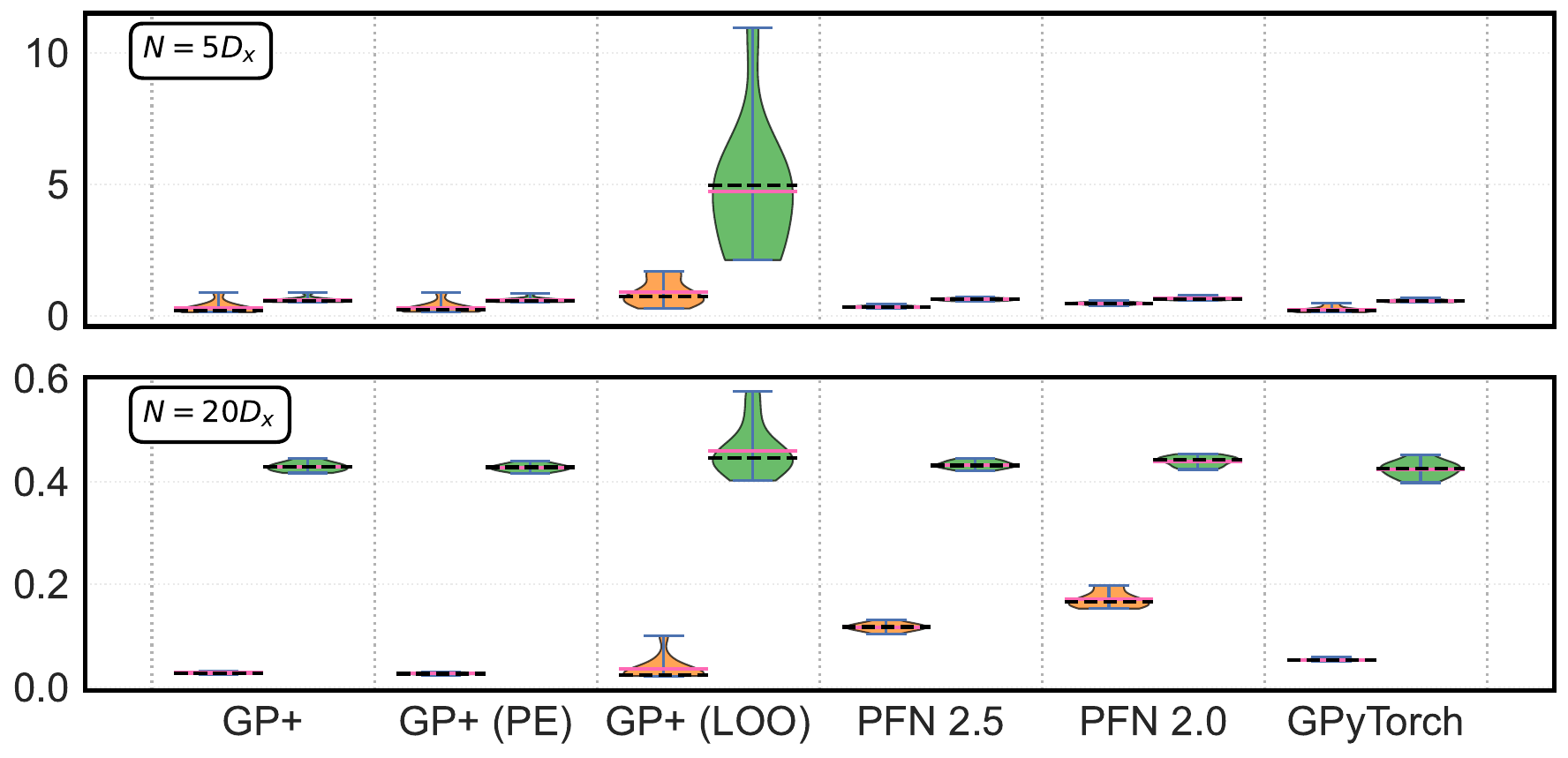}
        \subcaption{Wing Weight: D$_x$=10}
        \label{fig loo nis wing}
    \end{minipage}
    \hfill
    \begin{minipage}[b]{0.48\textwidth}
        \centering
        \includegraphics[width=\textwidth]{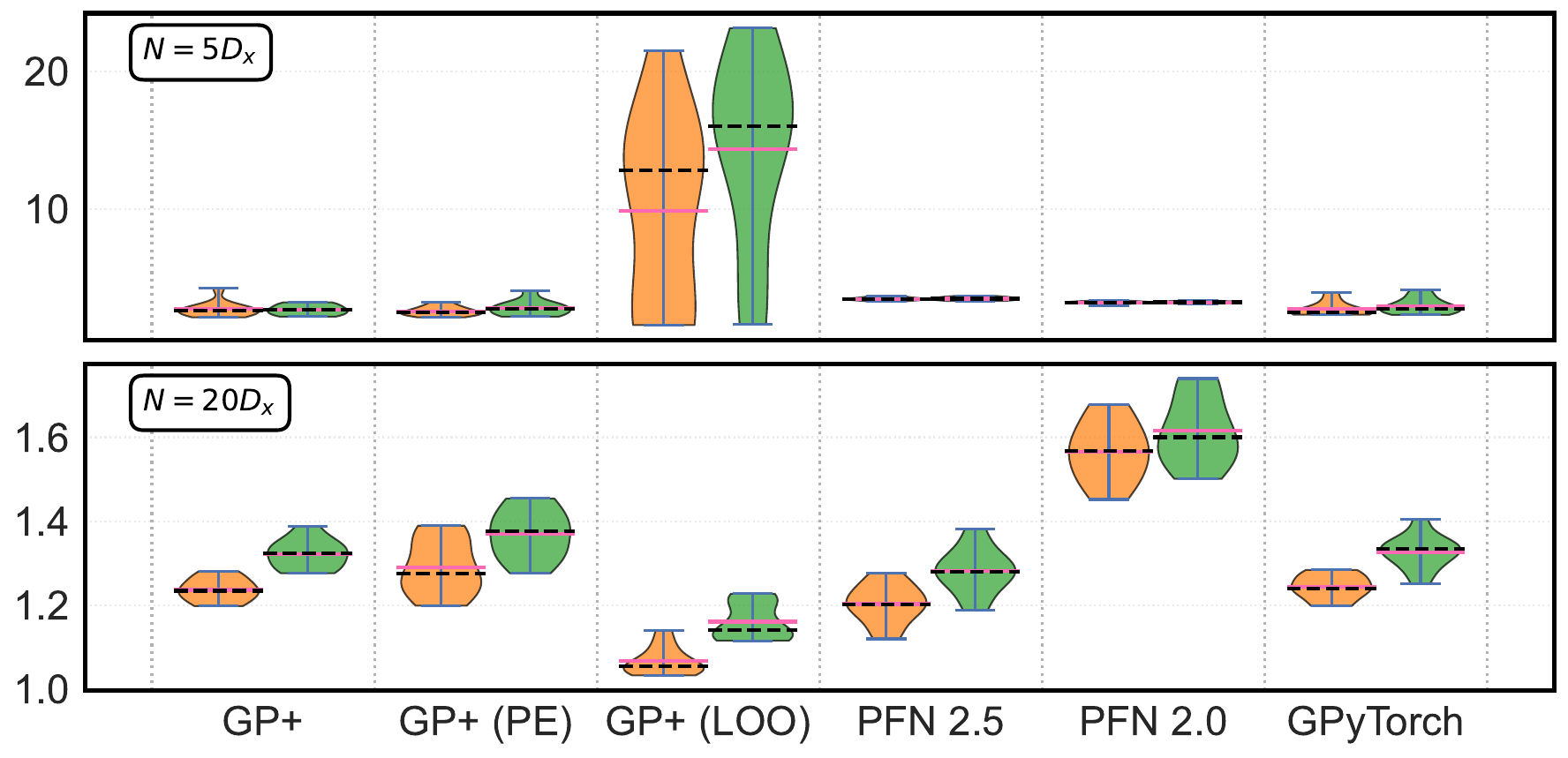}
        \subcaption{Ackley: D$_x$=20}
        \label{fig loo nis ackley20}
    \end{minipage}

    \vspace{1em}
    
    \begin{minipage}[b]{0.48\textwidth}
        \centering
        \includegraphics[width=\textwidth]{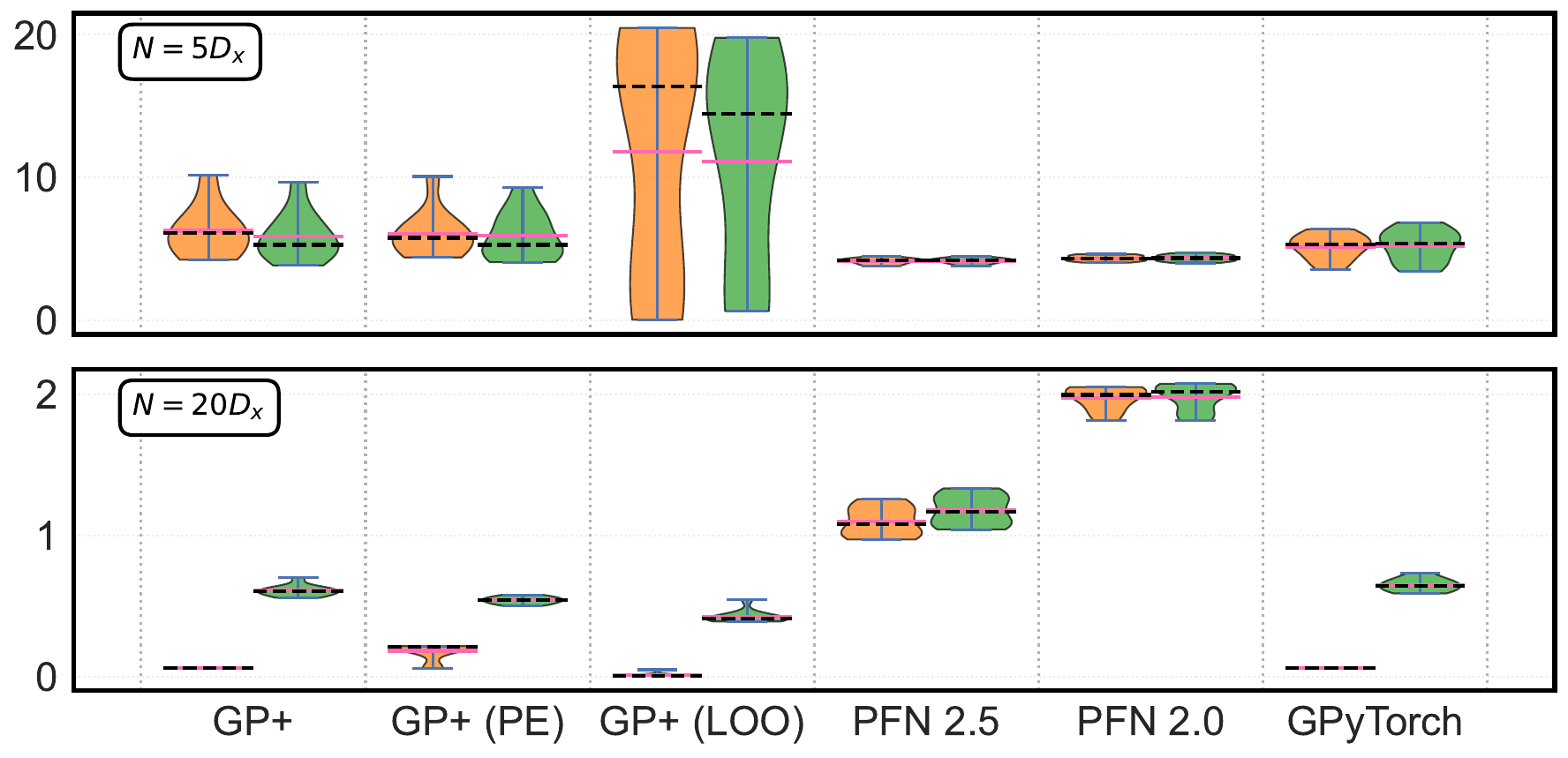}
        \subcaption{Griewank: D$_x$=20}
        \label{fig loo nis griewank}
    \end{minipage}
    \hfill
    \begin{minipage}[b]{0.48\textwidth}
        \centering
        \includegraphics[width=\textwidth]{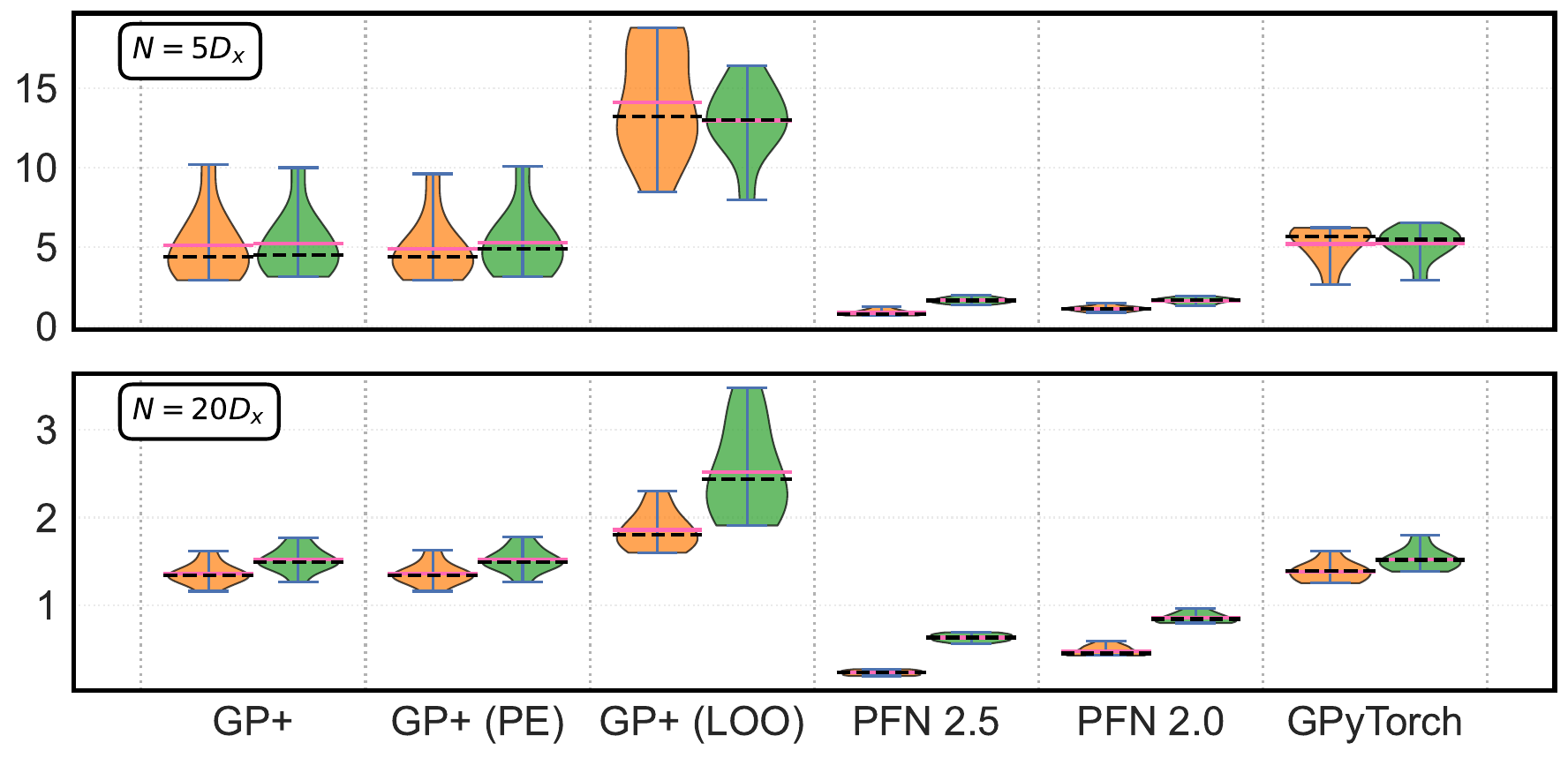}
        \subcaption{Zakharov: D$_x$=20}
        \label{fig loo nis zakharov}
    \end{minipage}

    \vspace{1em}
    
    \begin{minipage}[b]{0.48\textwidth}
        \centering
        \includegraphics[width=\textwidth]{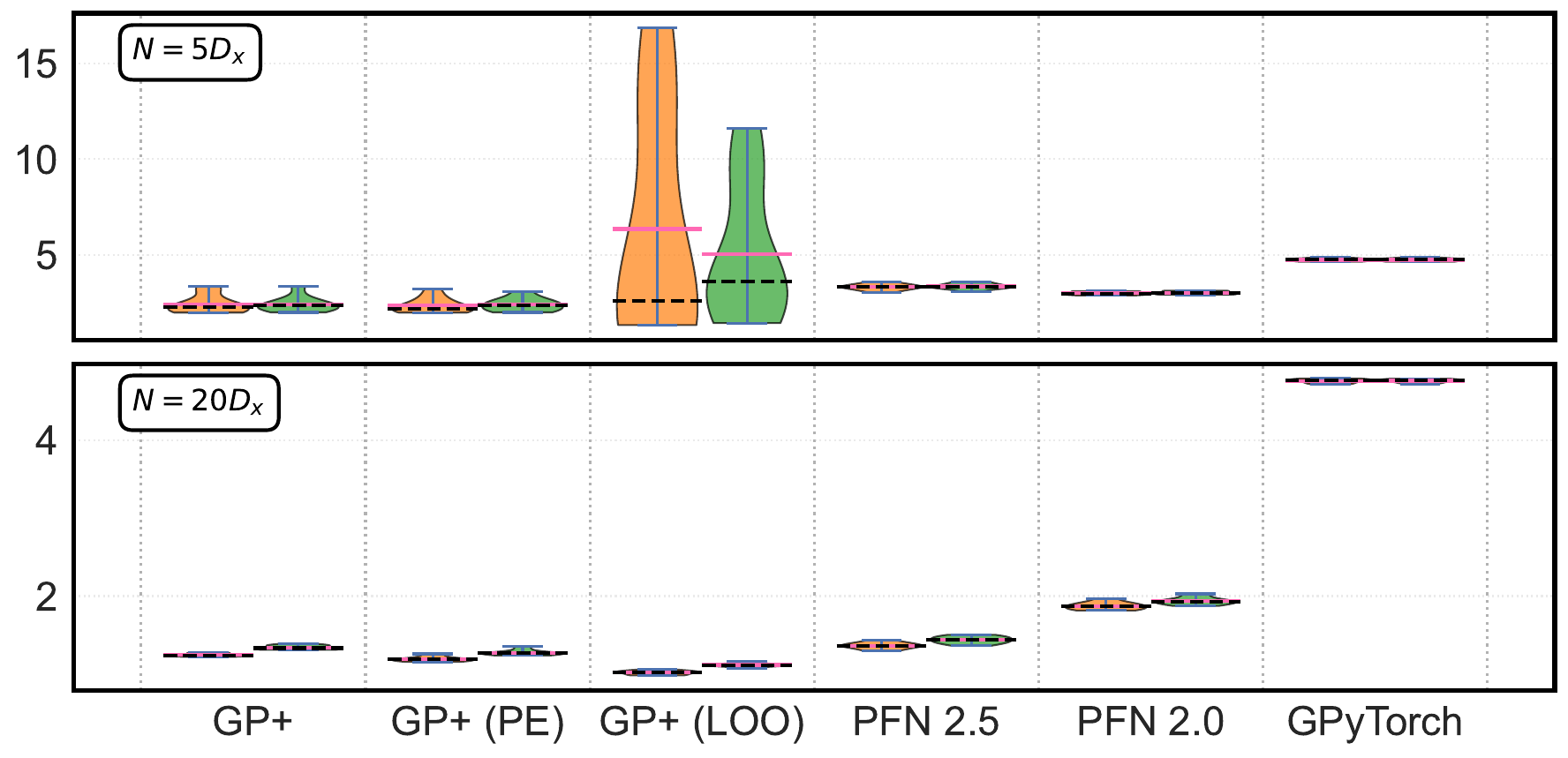}
        \subcaption{Ackley: D$_x$=40}
        \label{fig loo nis ackley40}
    \end{minipage}
    \hfill
    \begin{minipage}[b]{0.48\textwidth}
        \centering
        \includegraphics[width=\textwidth]{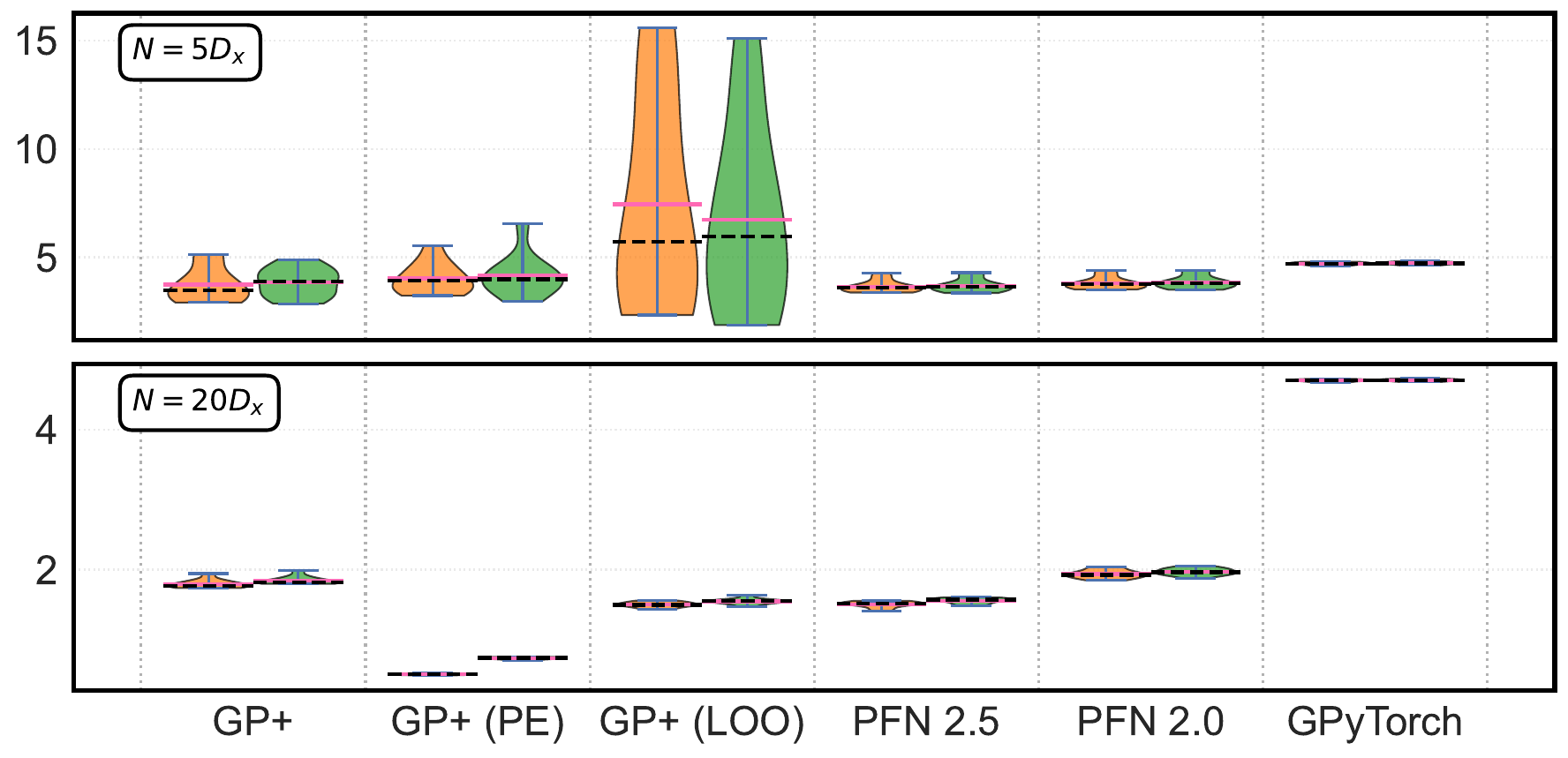}
        \subcaption{Dixon-Price: D$_x$=40}
        \label{fig loo nis dixon}
    \end{minipage}

    

    \caption{Evaluations based on NIS in \Cref{eq nis}: Noise is either $\varepsilon \sim \mathcal{N}(0, (0.002\times c)^2)$ or $\varepsilon \sim \mathcal{N}(0, (0.08\times c)^2)$. GPs are used on CPU while TabPFN is used on GPU. All the GPs are trained with the \textit{default} settings of GP+ and GPyTorch packages. GP+ (PE) and GP+ (LOO) denote cases where the default Gaussian kernel and loss function are switched to, respectively, the PE kernel and LOO-based loss in \Cref{eq loocv}.}

    \label{fig results loo nis}
\end{figure*}
\begin{figure*}[hptb]
    \centering
    \captionsetup{font=footnotesize}
    \begin{minipage}[b]{0.48\textwidth}
        \centering
        \includegraphics[width=\textwidth]{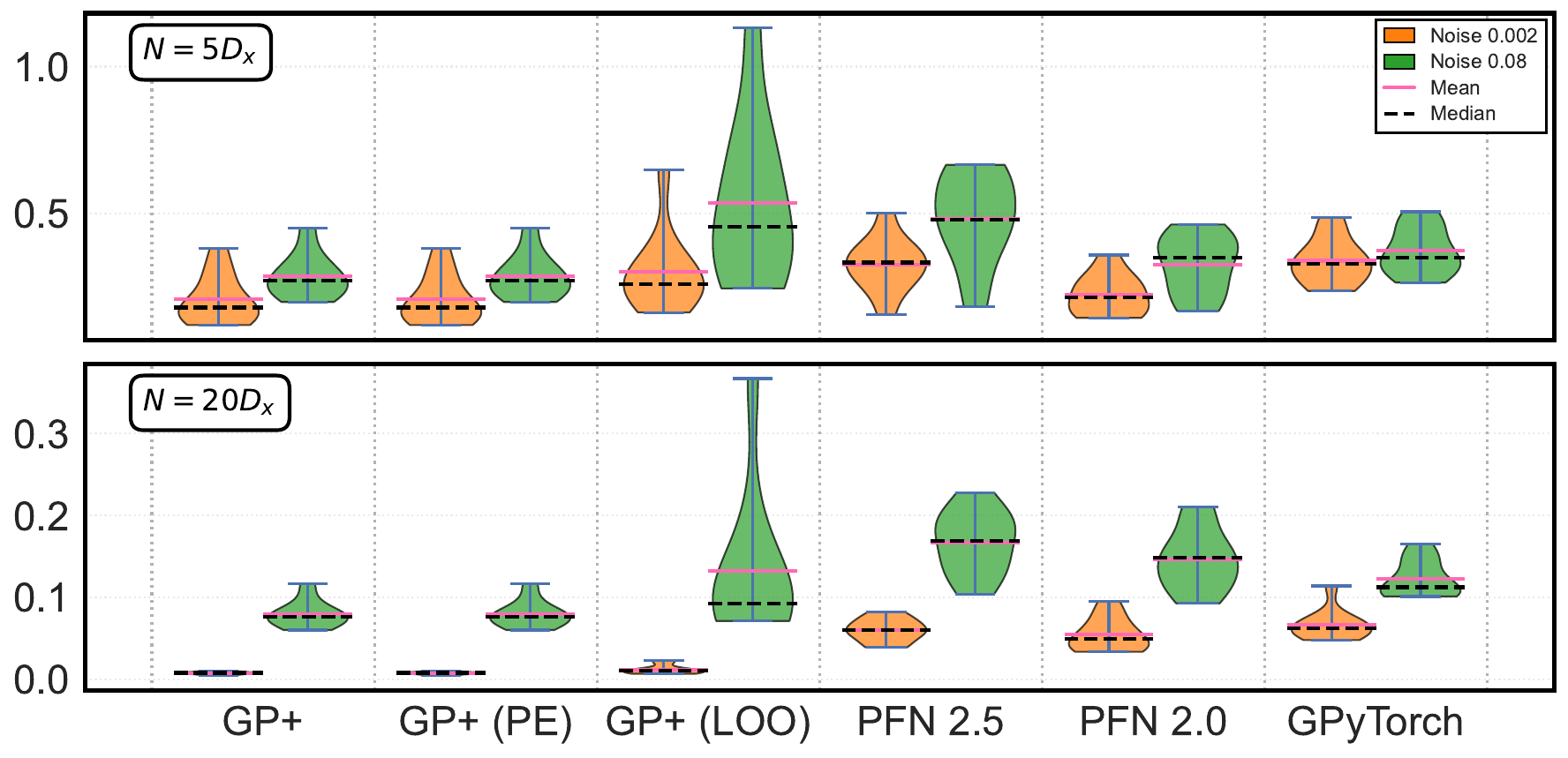}
        \subcaption{Buckling: D$_x$=4}
        \label{fig loo NCRPS buckling}
    \end{minipage}
    \hfill
    \begin{minipage}[b]{0.48\textwidth}
        \centering
        \includegraphics[width=\textwidth]{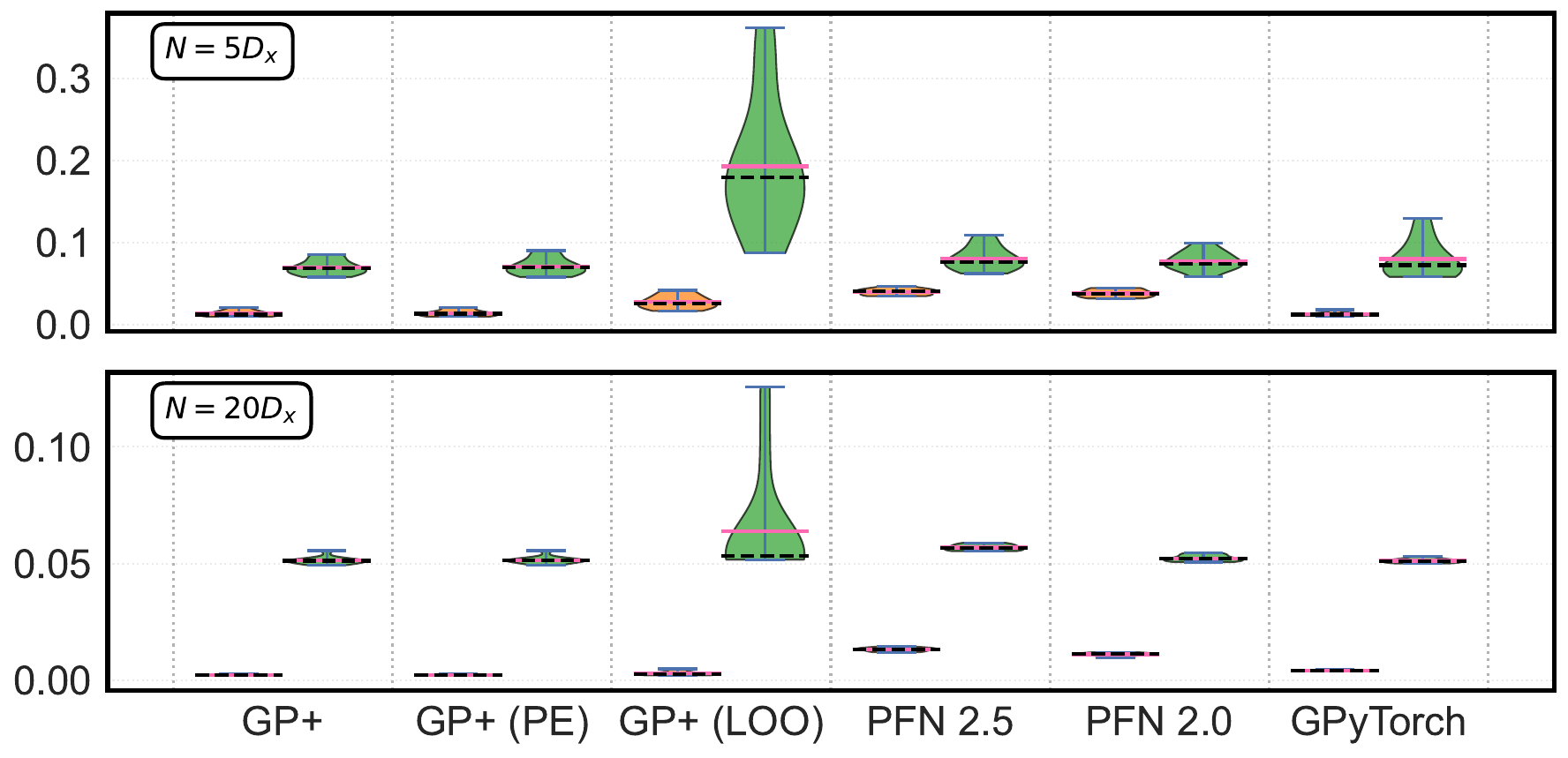}
        \subcaption{Borehole: D$_x$=8}
        \label{fig loo NCRPS borehole}
    \end{minipage}
    
    \vspace{1em}
    
    \begin{minipage}[b]{0.48\textwidth}
        \centering
        \includegraphics[width=\textwidth]{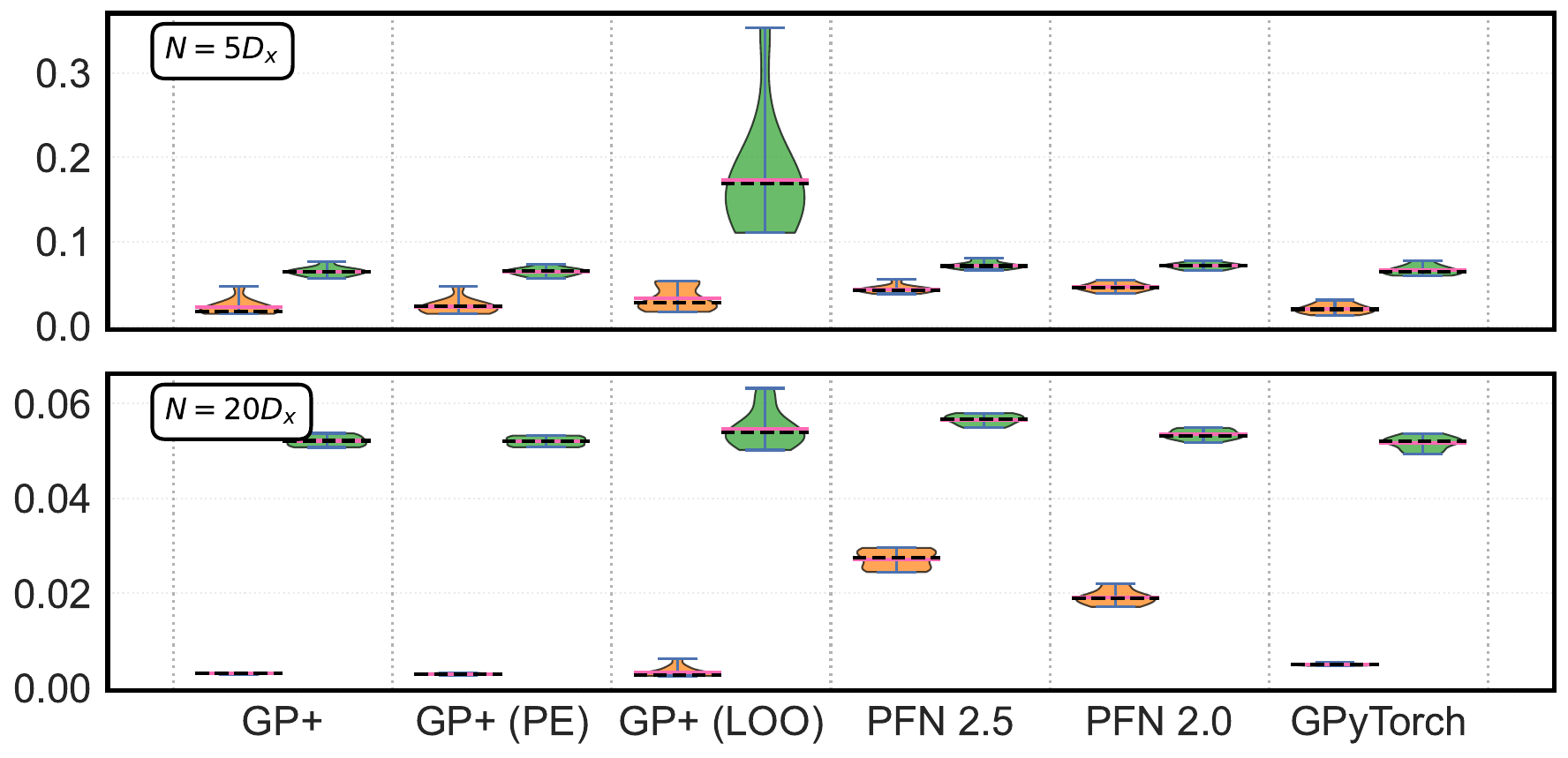}
        \subcaption{Wing Weight: D$_x$=10}
        \label{fig loo NCRPS wing}
    \end{minipage}
    \hfill
    \begin{minipage}[b]{0.48\textwidth}
        \centering
        \includegraphics[width=\textwidth]{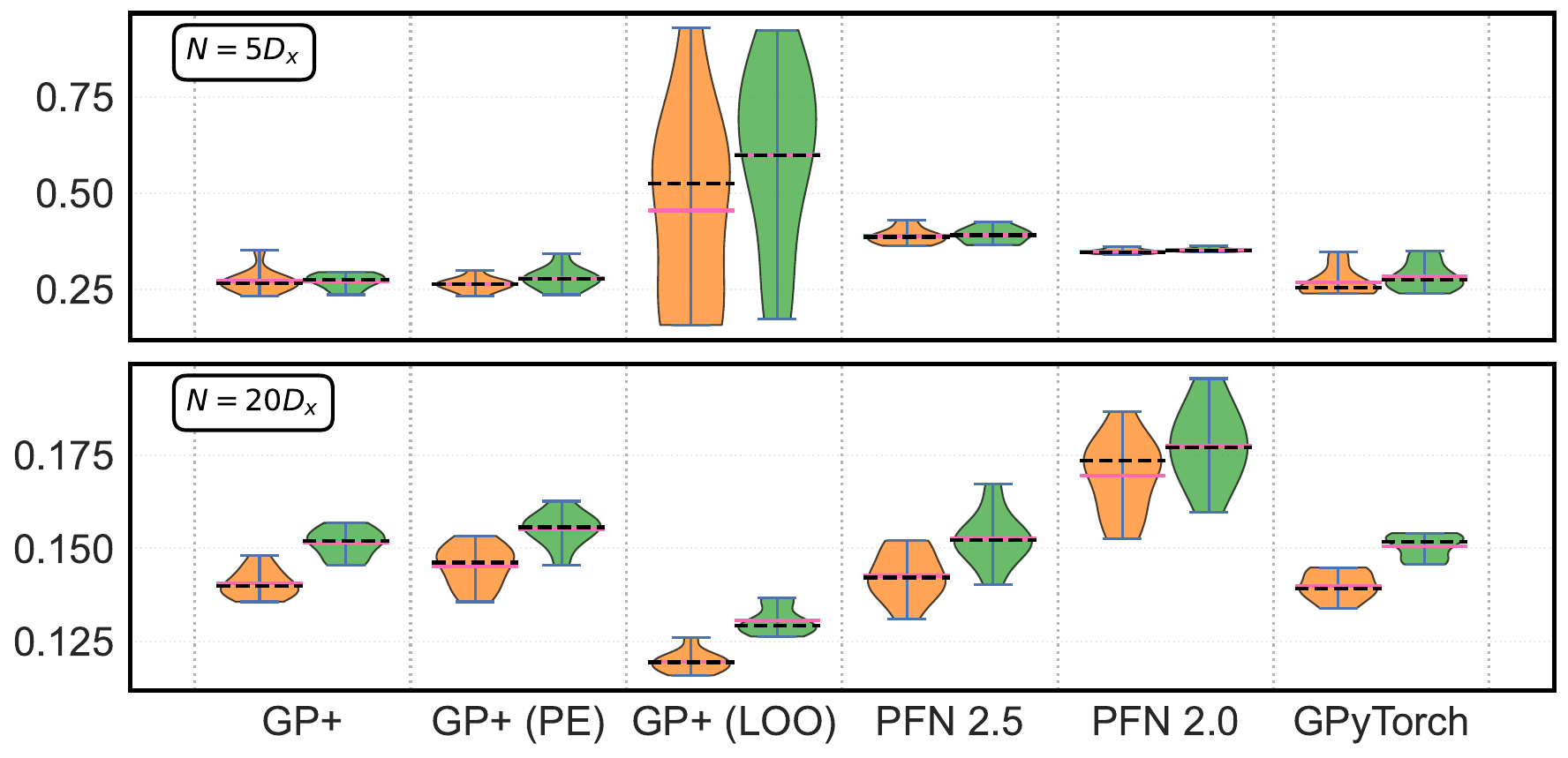}
        \subcaption{Ackley: D$_x$=20}
        \label{fig loo NCRPS ackley20}
    \end{minipage}

    \vspace{1em}
    
    \begin{minipage}[b]{0.48\textwidth}
        \centering
        \includegraphics[width=\textwidth]{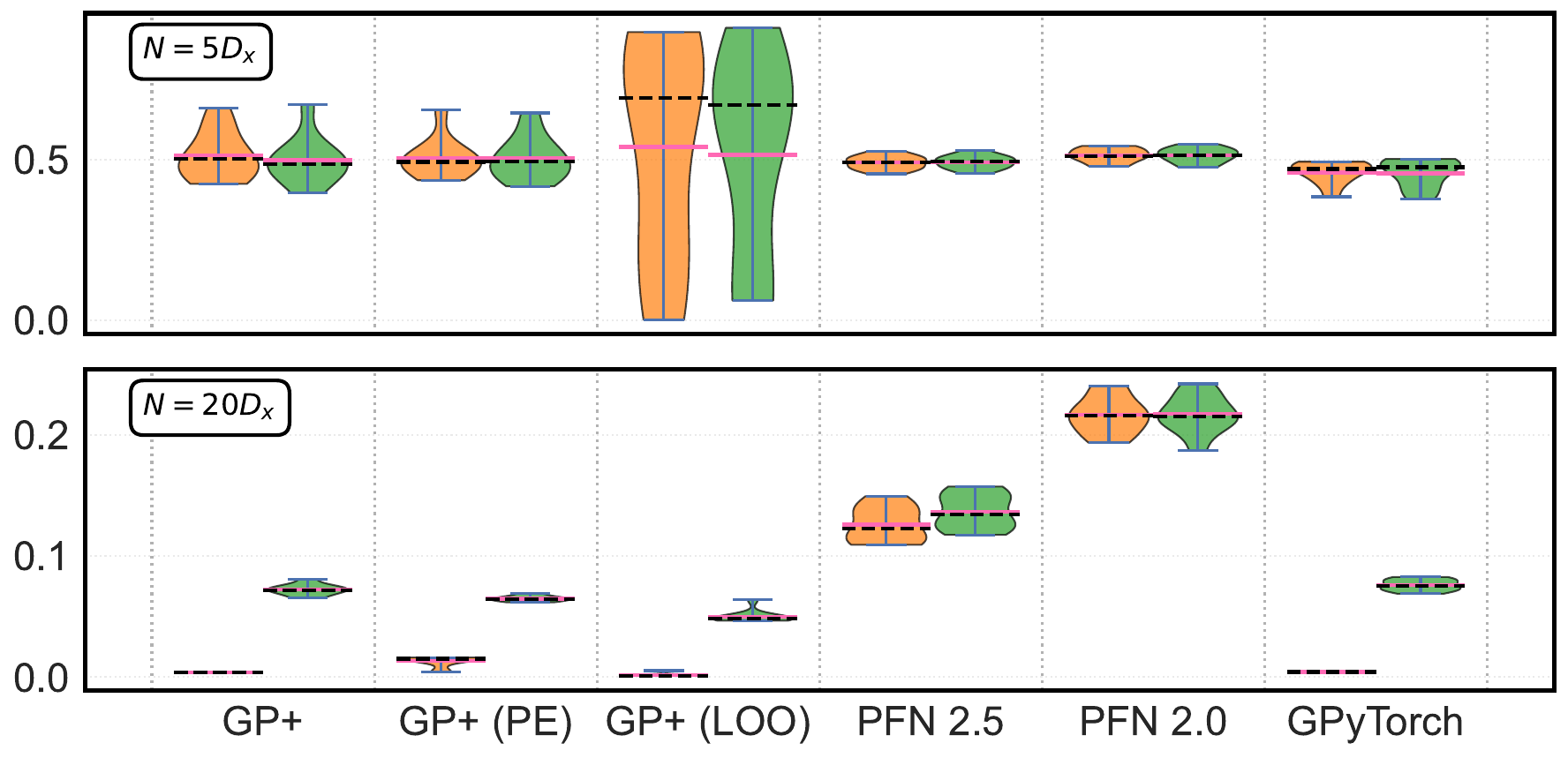}
        \subcaption{Griewank: D$_x$=20}
        \label{fig loo NCRPS griewank}
    \end{minipage}
    \hfill
    \begin{minipage}[b]{0.48\textwidth}
        \centering
        \includegraphics[width=\textwidth]{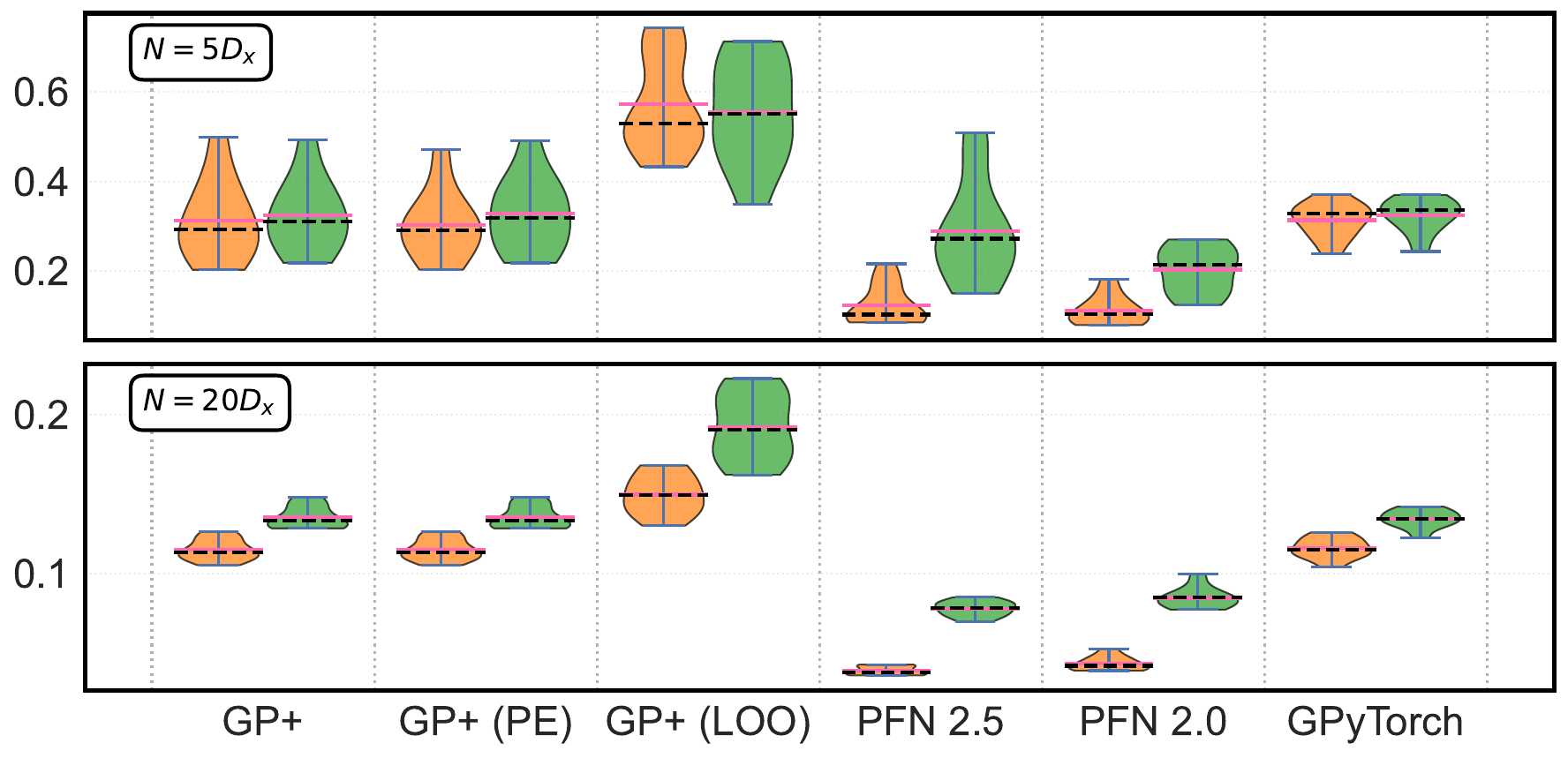}
        \subcaption{Zakharov: D$_x$=20}
        \label{fig loo NCRPS zakharov}
    \end{minipage}

    \vspace{1em}
    
    \begin{minipage}[b]{0.48\textwidth}
        \centering
        \includegraphics[width=\textwidth]{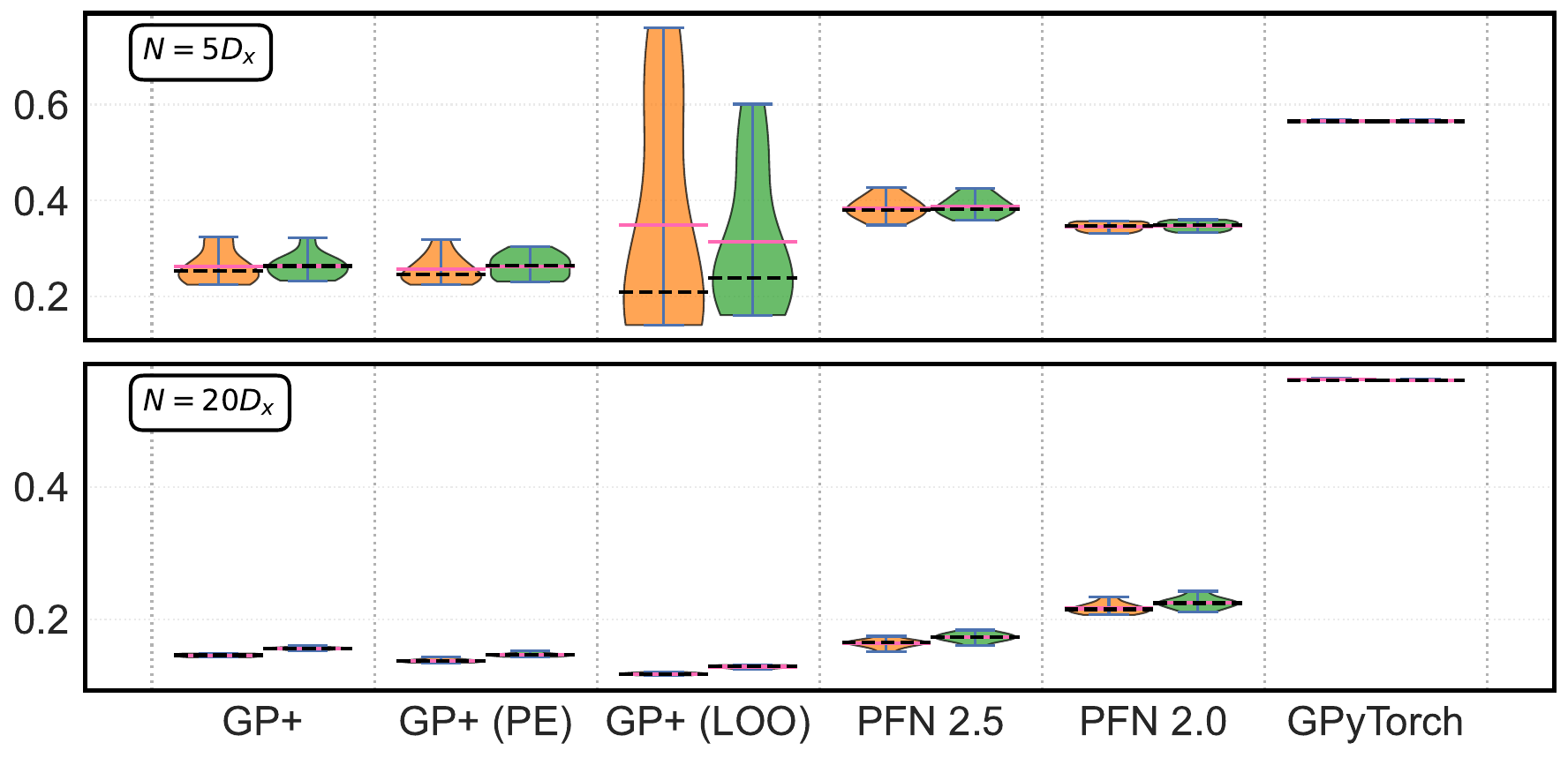}
        \subcaption{Ackley: D$_x$=40}
        \label{fig loo NCRPS ackley40}
    \end{minipage}
    \hfill
    \begin{minipage}[b]{0.48\textwidth}
        \centering
        \includegraphics[width=\textwidth]{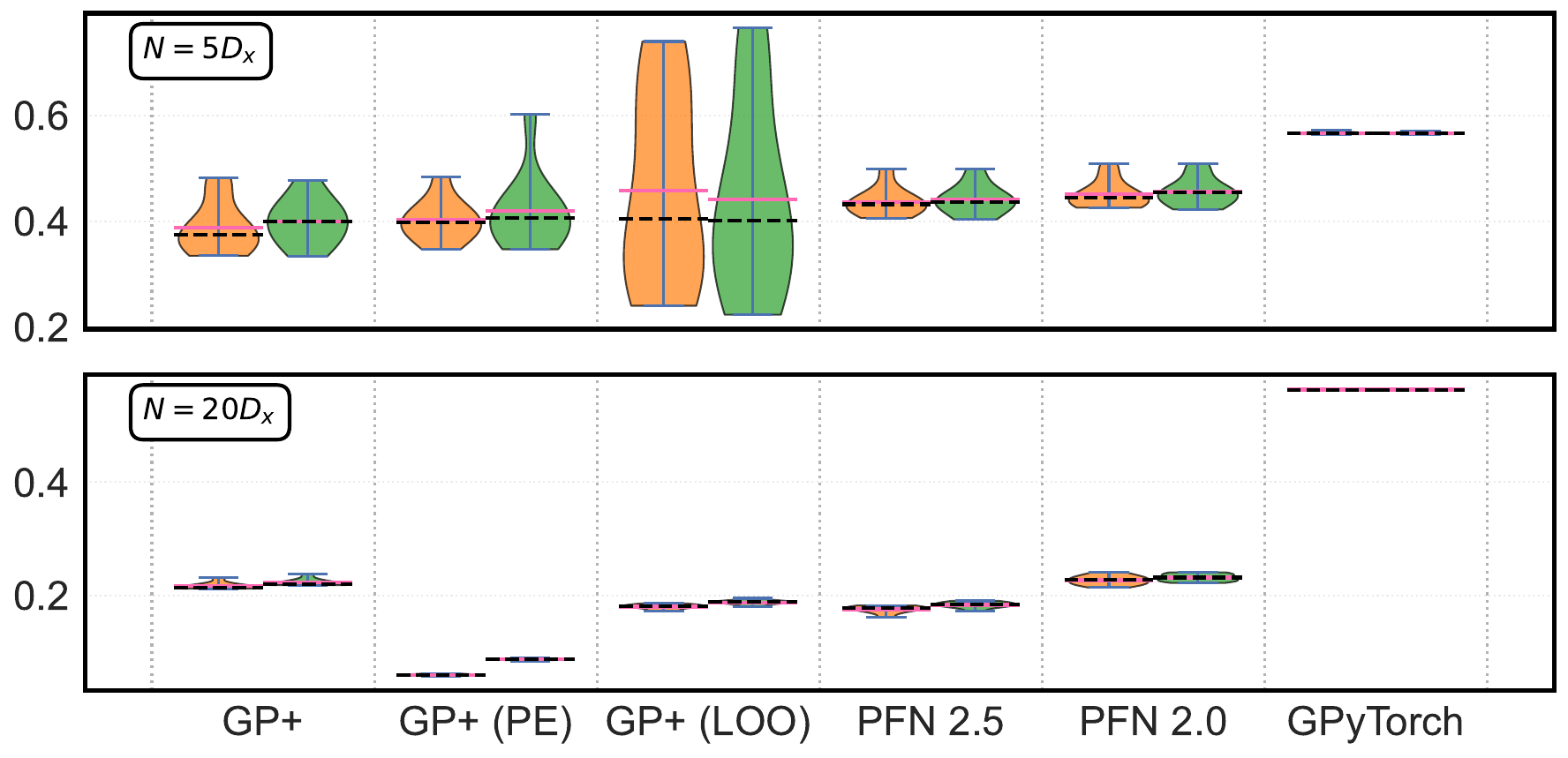}
        \subcaption{Dixon-Price: D$_x$=40}
        \label{fig loo NCRPS dixon}
    \end{minipage}
    
    \caption{Evaluations based on NCRPS in \Cref{eq ncrps}: Noise is either $\varepsilon \sim \mathcal{N}(0, (0.002\times c)^2)$ or $\varepsilon \sim \mathcal{N}(0, (0.08\times c)^2)$. GPs are used on CPU while TabPFN is used on GPU. All the GPs are trained with the \textit{default} settings of GP+ and GPyTorch packages. GP+ (PE) and GP+ (LOO) denote cases where the default Gaussian kernel and loss function are switched to, respectively, the PE kernel and LOO-based loss in \Cref{eq loocv}.}

    \label{fig results loo NCRPS}
\end{figure*}
\begin{figure*}[hptb]
    \centering
    \captionsetup{font=footnotesize}
    \begin{minipage}[b]{0.48\textwidth}
        \centering
        \includegraphics[width=\textwidth]{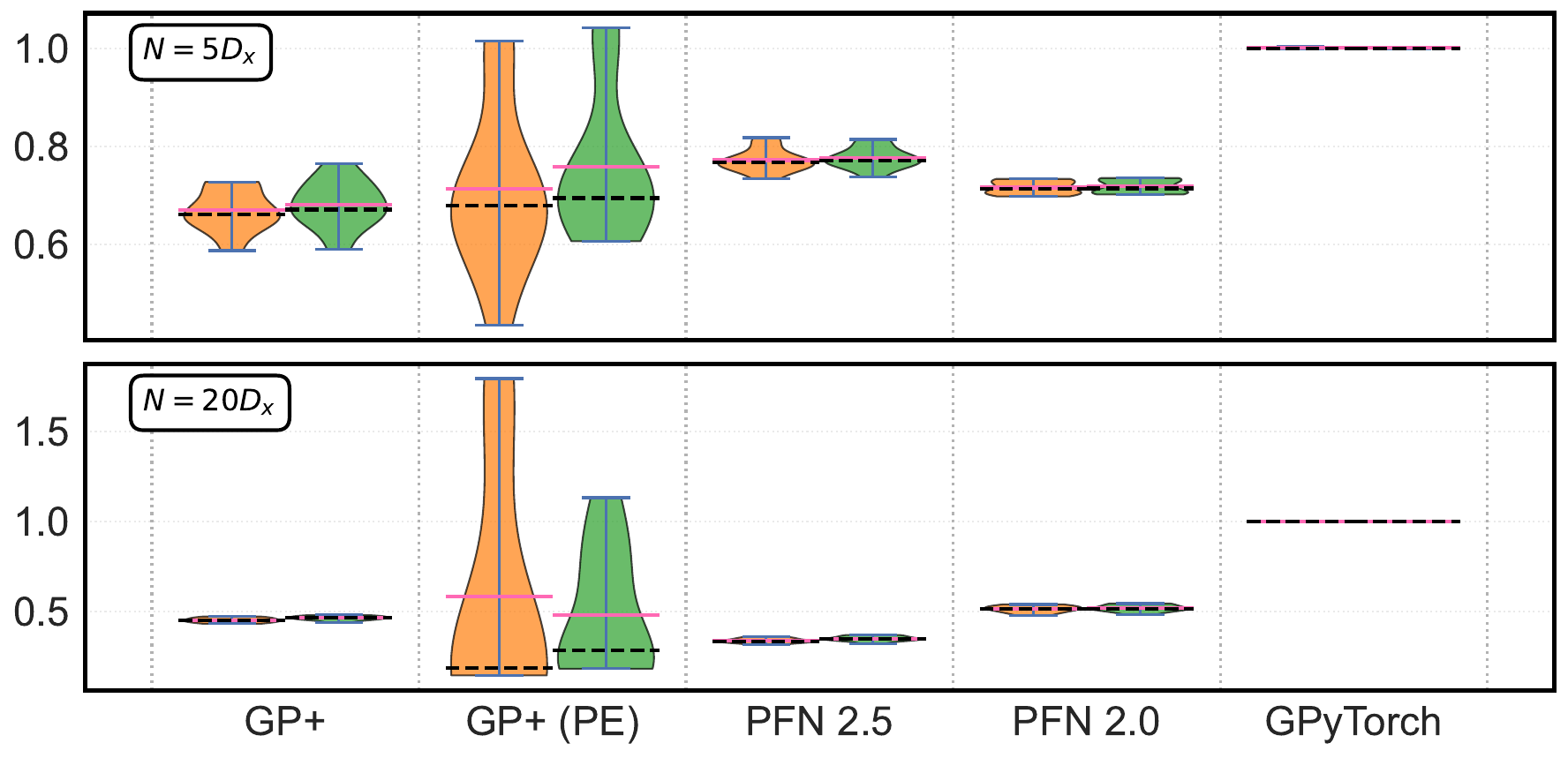}
        \subcaption{RRMSE}
        \label{fig rrmse rosenbrock}
    \end{minipage}
    \hfill
    \begin{minipage}[b]{0.48\textwidth}
        \centering
        \includegraphics[width=\textwidth]{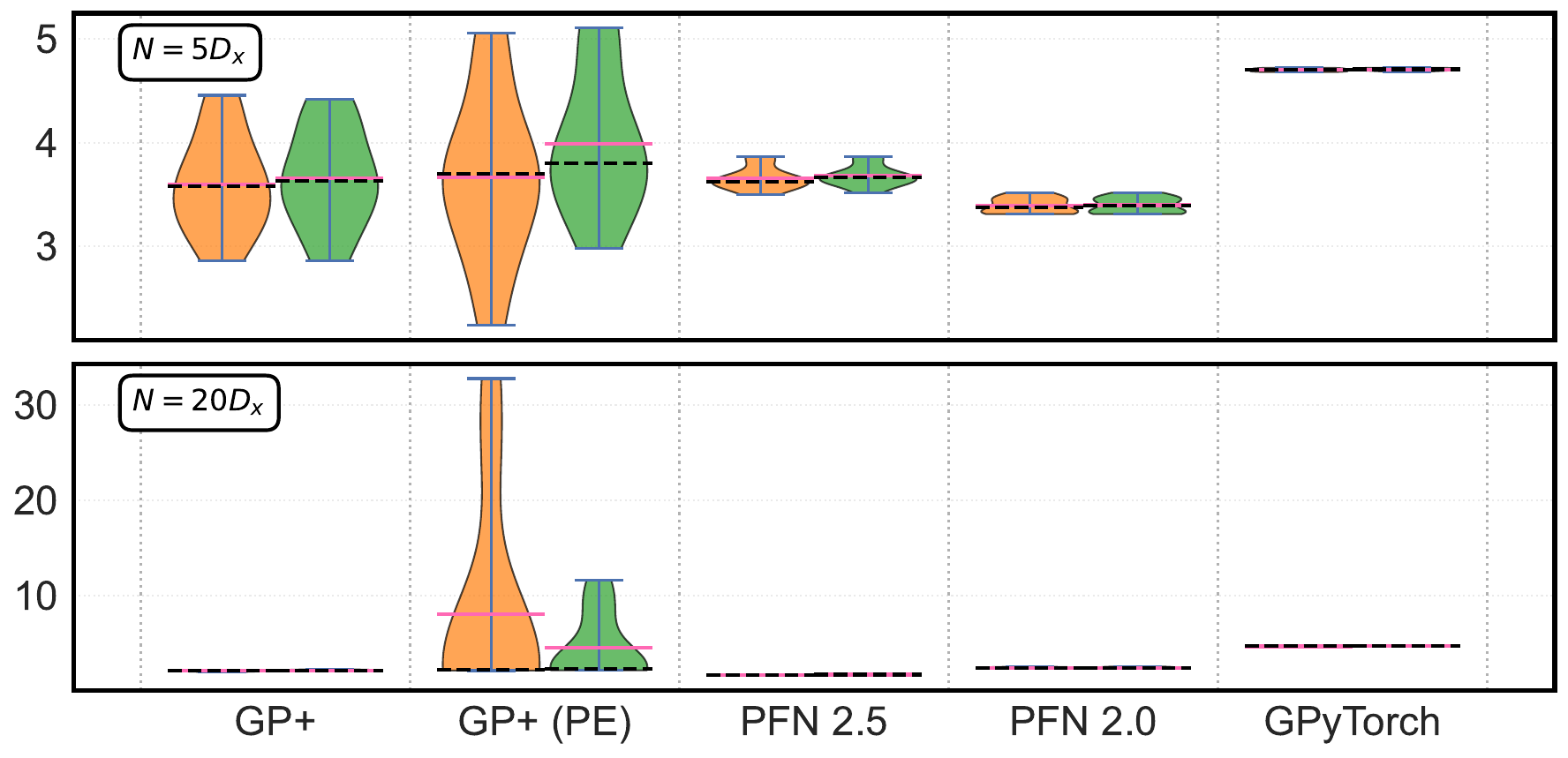}
        \subcaption{NIS}
        \label{fig nis rosenbrock}
    \end{minipage}
    \caption{Evaluations based on RRMSE and NIS (Rosenbrock $D_x=80$): Noise is either $\varepsilon \sim \mathcal{N}(0, (0.002\times c)^2)$ or $\varepsilon \sim \mathcal{N}(0, (0.08\times c)^2)$. GPs are used on CPU while TabPFN is used on GPU. All the GPs are trained with the \textit{default} settings of GP+ and GPyTorch packages. GP+ (PE) denotes the case where the default Gaussian kernel is switched to the PE kernel.}

    \label{fig results rosenbrock}
\end{figure*}

We compare GP+, PFN 2.5, PFN 2.0, and GPyTorch using RRMSE, NIS, and NCRPS. NIS and NCRPS behave similarly in our benchmarks, so we refer to NIS as our default proper scoring rule. When changing the default kernel from the Gaussian to PE or switching the loss from \cref{eq mle} to \cref{eq loocv}, the corresponding models built by GP+ are denoted by GP+ (PE) and GP+ (LOO). 

The results of our studies are visually summarized in Figures \ref{fig results loo rrrmse}, \ref{fig results loo nis}, and \ref{fig results rosenbrock} which indicate that GP+, GP+ (PE), and PFN 2.5 perform similarly and no one model can be selected as the best. However, it is important to note that GP+ (LOO), PFN 2.0, and GPyTorch performed the best in some cases, but they perform quite poorly in other cases (e.g., Gpytorch for Griewank vs Dixon-Price). These results clearly demonstrate the importance of selecting good kernels and loss functions, but how can we make that decision using the training data \textit{only}?

To answer this question, we can compare the optimized $\mathcal{L}(\cdot)$ in \cref{eq mle} under different settings and select the model with the best training loss. As shown in \cref{fig nll rankings}, the model with the best train-likelihood provides the best (or very close to best) test performance. 

Figures \ref{fig loo rrrmse buckling} and \ref{fig loo nis buckling} show that no model performs well in the low data regime of the Buckling example, but the UQ performance of TabPFN is significantly ahead of GPs (the performance of GPs can be improved by tuning the default categorical-to-latent variable mapping). 
Once more data is introduced to the training dataset, GP+ using the Gaussian and PE kernel are able to capture the behavior of the categorical inputs better than other models and their UQ performance meets or exceeds that of TabPFN. While GP+ (LOO) performed similarly to the other GP+ models, it suffers when more noise is introduced to the data. It is worth noting that PFN 2.5 performed less consistently than even PFN 2.0 with more training data.

\begin{figure*}[!b]
    \centering
    \captionsetup{font=footnotesize}
    \begin{minipage}[b]{0.48\textwidth}
        \centering
        \includegraphics[width=\textwidth]{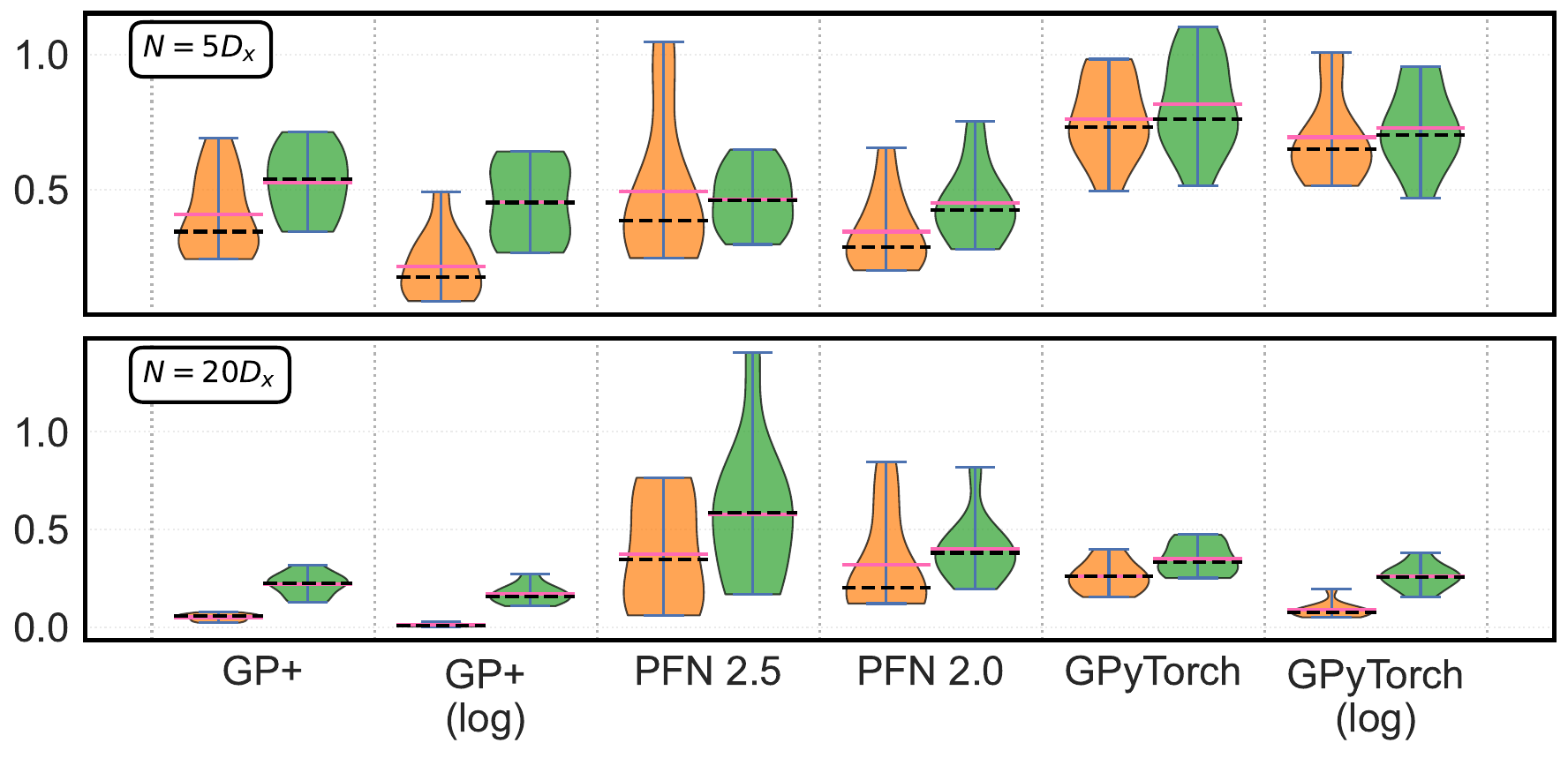}
        \subcaption{RRMSE}
        \label{fig logscale buckling rrmse}
    \end{minipage}
    \hfill
    \begin{minipage}[b]{0.48\textwidth}
        \centering
        \includegraphics[width=\textwidth]{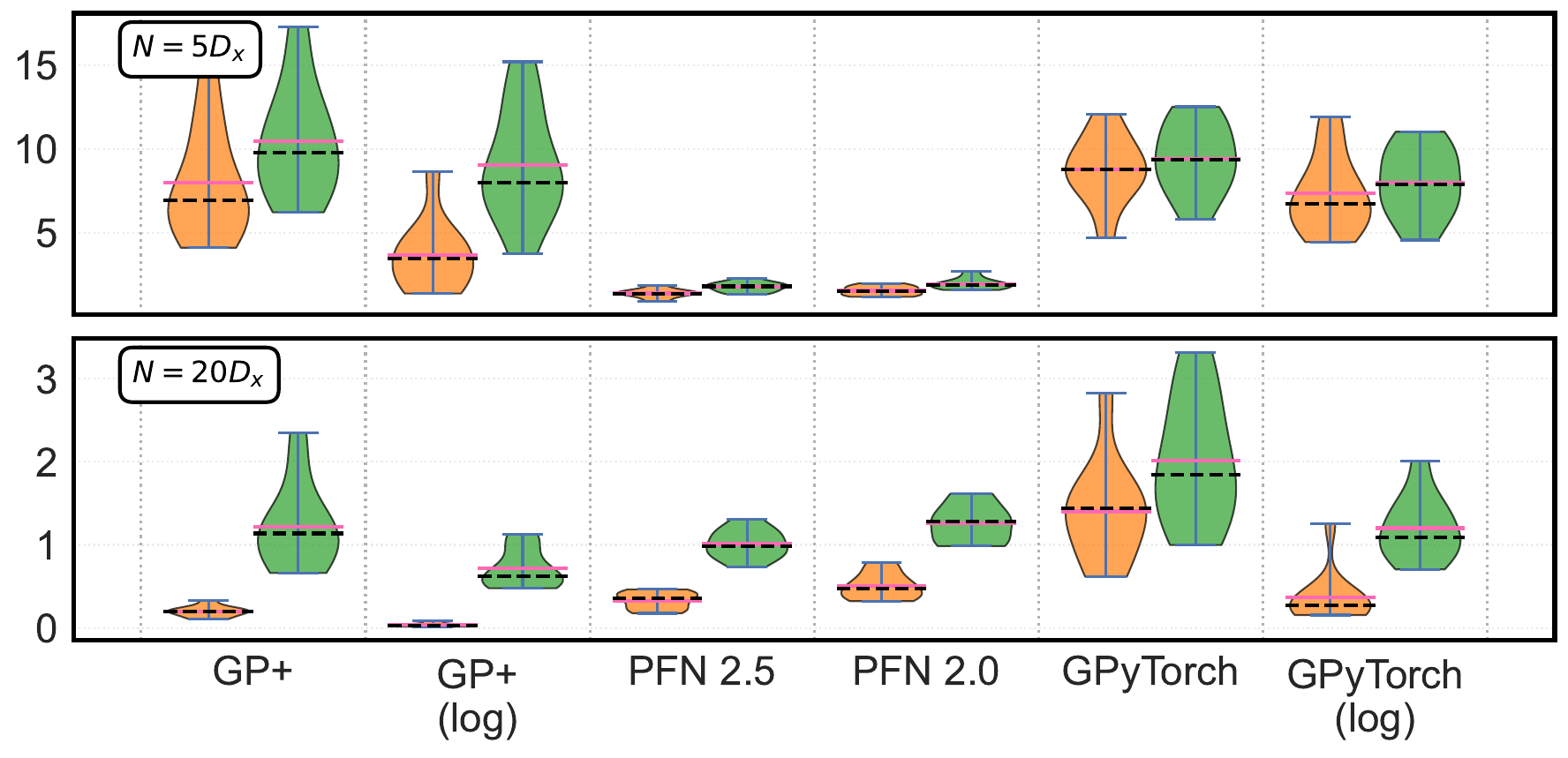}
        \subcaption{NIS}
        \label{fig logscale buckling nis}
    \end{minipage}
    \caption{Evaluations based on RRMSE and NIS (Buckling $D_x=4$) with and without log-scale target preprocessing: Noise is either $\varepsilon \sim \mathcal{N}(0, (0.002\times c)^2)$ or $\varepsilon \sim \mathcal{N}(0, (0.08\times c)^2)$. GPs are used on CPU while TabPFN is used on GPU. All the GPs are trained with the \textit{default} settings of GP+ and GPyTorch packages. GP+ (log) and GPyTorch (log) denote models trained after a log transform of the response, with metrics evaluated on the original scale.}
    \label{fig logscale buckling}
\end{figure*}

Target preprocessing can also matter for GPs. As the Buckling response follows a log-scale distribution, \Cref{fig logscale buckling} shows that preprocessing the output targets using a log-normal distribution improves GP+/GPyTorch RRMSE and NIS relative to default standardization, while PFN 2.5 and PFN 2.0 stay essentially unchanged. Further results for GP+ (PE) and GP+ (LOO) log-scale results for Buckling and Zakharov are available in the project repository \cite{GithubRepo}. This motivates future work on automated preprocessing for GPs, since TabPFN is already robust to such scaling choices.

The Borehole example clearly identifies GP+ (LOO) and Gpytorch as the worst models in the small-data large-noise case and large-data case, respectively. With Gpytorch, many repetitions have RRMSE of $1.0$ which indicates that the model is mostly learning a constant value equal to the data mean. By improving the optimization process GP+ and GP+ (PE) avoid these issues and perform on par with each other and slightly better than PFN 2.5 and PFN 2.0.

In Figures \ref{fig loo rrrmse wing} and \ref{fig loo nis wing}, we observe that GP+ (LOO) is the only GP model that performs poorly--in line with the expectation that \cref{eq loocv} can be a sensitive loss function. For this example, the other GPs outperform TabPFN versions in the case of small noise variance but perform similarly with large noise where the errors are dominated by the noise variance.


The potential strength of alternative loss functions, specifically \cref{eq loocv}, is shown in the Ackley $D_x=20$ and $D_x=40$ examples with $N=20D_x$ where $\mathcal{L}_{PL}(\cdot)$ produces the best model. However, as mentioned before, this loss is sensitive and this is evident in the small-data versions of the same examples.


In all examples except Rosenbrock, GP+ (LOO) has the poorest UQ ability in the low data regime out of all the other models. Upon examining the UQ performance of GP+ (LOO) on the test data, we see that the first term in \cref{eq interval score}, width, is smaller than the models using MLE-based optimization. The NIS is penalized more from the second and third terms, which shows overconfidence in the point predictions.

With the Griewank example we see that none of the models perform well in the low data regime, especially GP+ (LOO) which manages to learn the function well in some random trials but performs very poorly on other trials. With $N=20D_x$, all GP variants consistently outperform TabPFN versions. 

Both versions of TabPFN perform better in emulation than GPs on the Zakharov example. This trend is due to the observation scale which is very large and not properly handled by the default standardization adopted in GP+ and Gpytorch. This issue will reappear in \cref{subsec results bo}.

Higher dimensional examples such as Ackley $D_x=40$, Dixon-Price, and Rosenbrock show that GP+ (PE) can train models that are far better than the other array of methods benchmarked. One clear tradeoff seen in \cref{fig results rosenbrock} is that using MLE with the PE kernel can make the model more susceptible to reaching bad local minima, especially in higher dimensions. GP+ (LOO) is not shown in \cref{fig results rosenbrock} because its performance is far worse than the other models and it would obstruct the clarity of the figure. By increasing the number of initializations of the hyperparameters, a more consistent performance from GP+ (PE) can be achieved but this would drastically increase the cost of an already expensive method, see \cref{tab timing loo}. 

Lastly, \cref{tab timing loo} lists the computational cost of the models (GPs use CPU while TabPFN uses GPU). We note that there is an important practical trade-off: PFN methods have lower fit/setup time, while fitted GP models have much lower prediction time. The cost of training a GP increases sharply with dimensionality and sample size, especially for GP+ (PE) which can be over ten times as expensive to train. GP+ (LOO) is also more expensive than the default GP+ model using the Gaussian kernel, but significantly less expensive than the PE kernel. There is a noticeable discrepancy in the inference speed of PFN versions: PFN 2.5 has a lower inference cost that is highlighted in higher dimensional regimes like the Rosenbrock example, but the model will not always perform better than its older counterpart.

The training cost of GPs can be reduced in a number of ways which will be introduced to GP+ in the near future.

\subsection{BO Benchmarks} \label{subsec results bo}
\begin{figure*}[htb]
    \centering
    \captionsetup{font=footnotesize}
    \begin{minipage}[b]{0.32\linewidth}
        \centering
        \includegraphics[width=\linewidth]{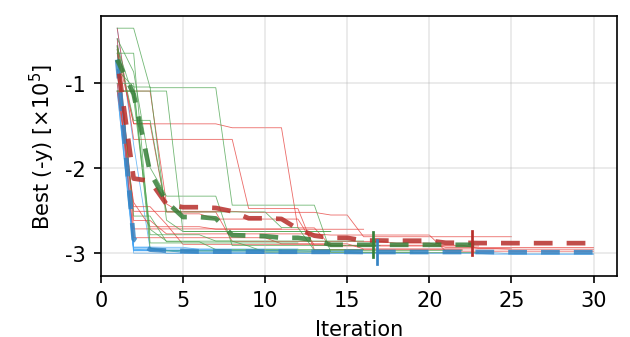}
        \subcaption{Buckling: D$_x$=4}
        \label{fig bo buckling noisehigh}
    \end{minipage}
    \hfill
    \begin{minipage}[b]{0.32\linewidth}
        \centering
        \includegraphics[width=\linewidth]{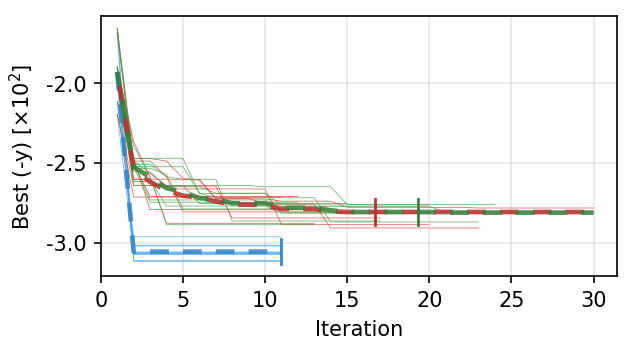}
        \subcaption{Borehole: D$_x$=8}
        \label{fig bo borehole noisehigh}
    \end{minipage}
    \hfill
    \begin{minipage}[b]{0.32\linewidth}
        \centering
        \includegraphics[width=\linewidth]{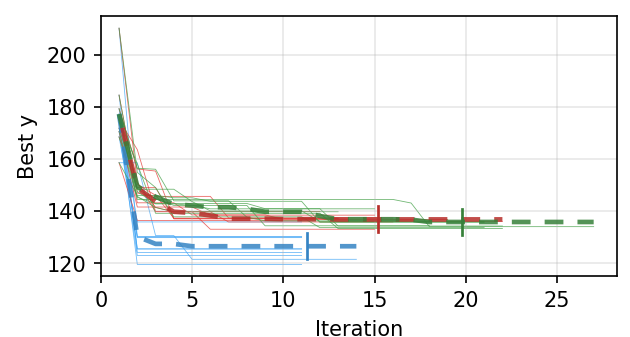}
        \subcaption{Wing Weight: D$_x$=10}
        \label{fig bo wing noisehigh}
    \end{minipage}

    \vspace{0.5em}

    \begin{minipage}[b]{0.32\linewidth}
        \centering
        \includegraphics[width=\linewidth]{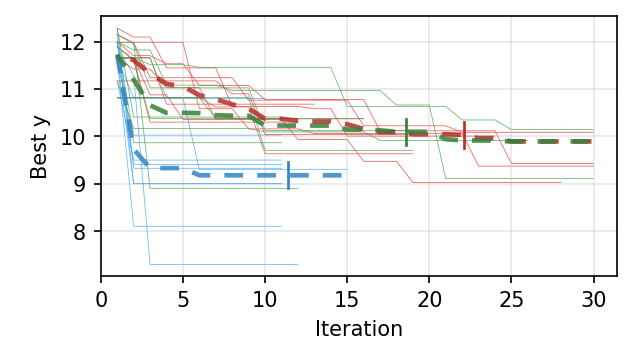}
        \subcaption{Ackley: D$_x$=20}
        \label{fig bo ackleyD20 noisehigh}
    \end{minipage}
    \hfill
    \begin{minipage}[b]{0.32\linewidth}
        \centering
        \includegraphics[width=\linewidth]{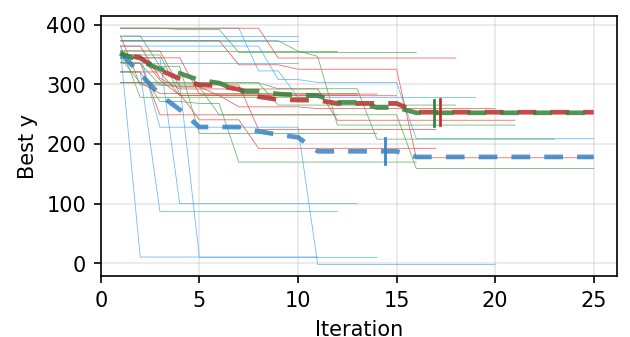}
        \subcaption{Griewank: D$_x$=20}
        \label{fig bo griewank noisehigh}
    \end{minipage}
    \hfill
    \begin{minipage}[b]{0.32\linewidth}
        \centering
        \includegraphics[width=\linewidth]{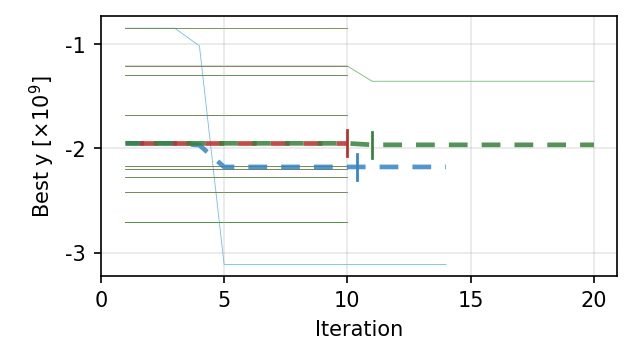}
        \subcaption{Zakharov: D$_x$=20}
        \label{fig bo zakharov noisehigh}
    \end{minipage}

    \vspace{0.5em}

    \begin{minipage}[b]{0.32\linewidth}
        \centering
        \includegraphics[width=\linewidth]{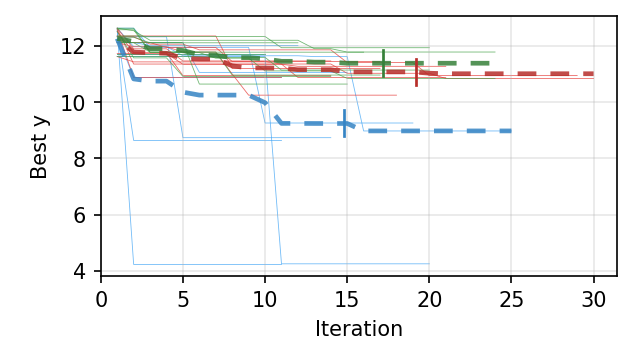}
        \subcaption{Ackley: D$_x$=40}
        \label{fig bo ackleyD40 noisehigh}
    \end{minipage}
    \hfill
    \begin{minipage}[b]{0.32\linewidth}
        \centering
        \includegraphics[width=\linewidth]{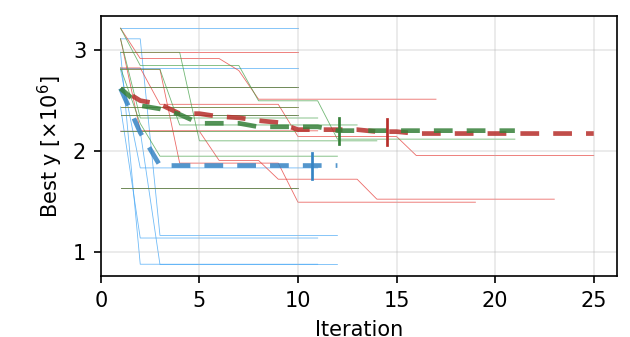}
        \subcaption{Dixon-Price: D$_x$=40}
        \label{fig bo dixonprice noisehigh}
    \end{minipage}
    \hfill
    \begin{minipage}[b]{0.32\linewidth}
        \centering
        \includegraphics[width=\linewidth, trim = 0 10pt 20pt 0, clip]{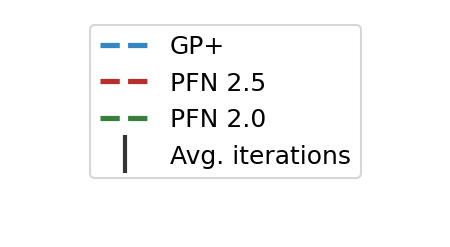}
        \subcaption*{\phantom{X}}
    \end{minipage}
    \caption{Bayesian optimization performance for GP+ and TabPFN across benchmark problems with observation noise $\varepsilon \sim \mathcal{N}(0,(0.08\times c)^2)$. Buckling and Borehole are maximization problems, hence best negative y values are reported.}
    \label{fig bo plots noisehigh}
\end{figure*}

We evaluate GP and TabPFN as surrogates within BO on the identical benchmark problems, data-generation procedures, preprocessing steps, and noise levels described in \cref{subsec results regression}. For each problem, we conduct 10 independent BO trials. The initial design for each trial consists of \(N_0 = 5 D_x\) points drawn via a scrambled Sobol sequence over the problem bounds. Each BO run is allotted a maximum of 30 sequential evaluations beyond the initial design, with early stopping triggered if the best-observed value does not improve for 10 consecutive iterations. The AF across all experiments is expected improvement (EI). We only use GP+ with its default settings in this section. 


For the GP-based BO, the surrogate is retrained from scratch via MLE after each new observation, maintaining the training protocol from \cref{subsec results regression}. To select the next query point, we maximize EI using multi-start gradient-based optimization carried out via L-BFGS with $64$ random initializations. 
For TabPFN-based BO, we utilize the same in-context regression setup from \cref{subsec results regression}. To select the next query point, we evaluate EI on 5,000 candidate points drawn from a scrambled Sobol sequence and select the candidate with the highest EI.

The results of our studies are summarized in \cref{fig bo plots noisehigh} for the large noise version of the problems (very similar trends are observed with small noise versions and hence the corresponding plot is excluded to save space). 
We observe that GP trajectories tend to approach the optimum faster in Buckling, Borehole, and Wing Weight. More broadly, GP outperforms TabPFN on all problems except Zakharov in \cref{fig bo zakharov noisehigh} where the observations are standardized for the GP, consistent with the trends in \cref{subsec results regression}. 
Overall, the BO results suggest a practical trade-off between surrogate capability and computational overhead. On the one hand, TabPFN eliminates per-iteration retraining costs but it renders gradient-based AF optimization very expensive. On the other hand, GP requires retraining at each iteration via MLE but supports analytic AFs and renders fast post-fit predictions over large candidate pools. It is worth noting that GP can also adopt a candidate-pool-based AF strategy in lieu of continuous optimization, which substantially reduces BO overhead while retaining the predictive advantages of a well-trained surrogate. In small-data regimes where GP retraining is inexpensive, this can make GP computationally competitive with or even preferable to TabPFN. Given the observed BO outcomes and the common setting where objective evaluations are the dominant cost, we consider GPs as the preferred surrogate for BO at least in the dimensionality ranges explored in this work (i.e., roughly $D_x<100$).

\subsection{Classification Benchmarks} \label{subsec results classification}
\begin{figure*}[ht]
\centering
\captionsetup{font=footnotesize}
\includegraphics[width=0.62\textwidth]{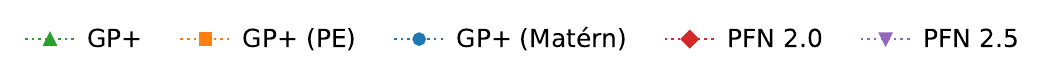}\\[0.3em]
\begin{minipage}[b]{0.24\textwidth}
    \centering
    \includegraphics[width=\textwidth]{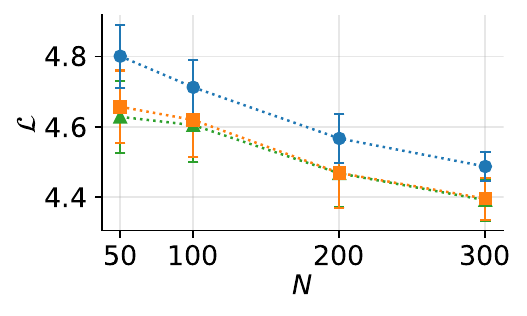}
    \subcaption{Electrical Grid: $\mathrm{D}_x{=}11$}
    \label{fig electrical grid nll}
\end{minipage}
\hfill
\begin{minipage}[b]{0.24\textwidth}
    \centering
    \includegraphics[width=\textwidth]{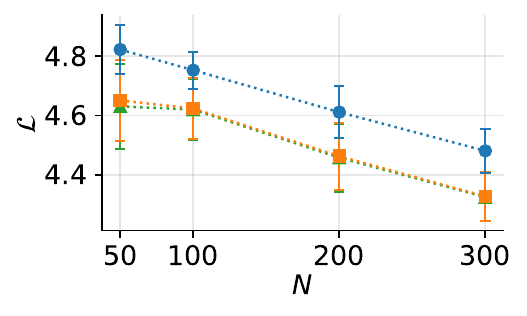}
    \subcaption{Truss 6D: $\mathrm{D}_x{=}6$}
    \label{fig truss 6d nll}
\end{minipage}
\hfill
\begin{minipage}[b]{0.24\textwidth}
    \centering
    \includegraphics[width=\textwidth]{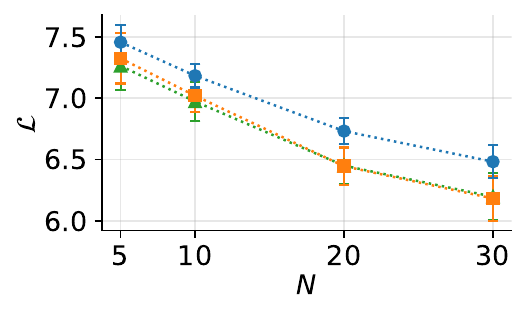}
    \subcaption{Stellar: $\mathrm{D}_x{=}8$}
    \label{fig stellar nll}
\end{minipage}
\hfill
\begin{minipage}[b]{0.24\textwidth}
    \centering
    \includegraphics[width=\textwidth]{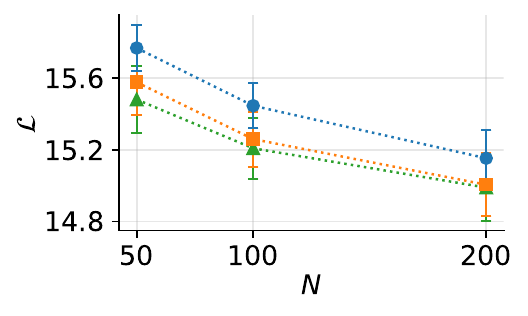}
    \subcaption{Steel Plates Faults: $\mathrm{D}_x{=}27$}
    \label{fig steel plates nll}
\end{minipage}
\begin{minipage}[b]{0.24\textwidth}
    \centering
    \includegraphics[width=\textwidth]{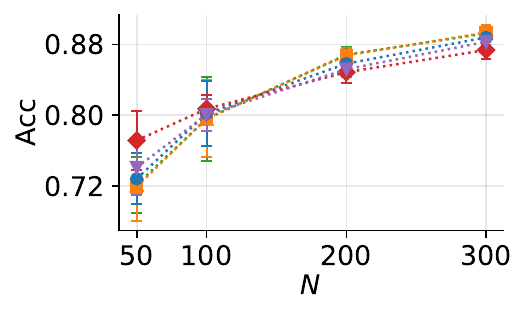}
    \subcaption{}
    \label{fig electrical grid acc}
\end{minipage}
\hfill
\begin{minipage}[b]{0.24\textwidth}
    \centering
    \includegraphics[width=\textwidth]{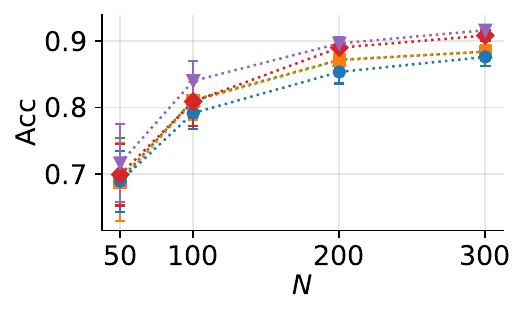}
    \subcaption{}
    \label{fig truss 6d acc}
\end{minipage}
\hfill
\begin{minipage}[b]{0.24\textwidth}
    \centering
    \includegraphics[width=\textwidth]{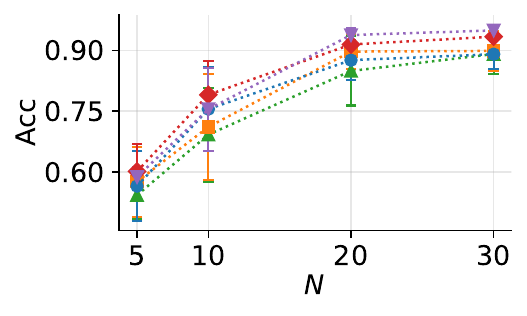}
    \subcaption{}
    \label{fig stellar acc}
\end{minipage}
\hfill
\begin{minipage}[b]{0.24\textwidth}
    \centering
    \includegraphics[width=\textwidth]{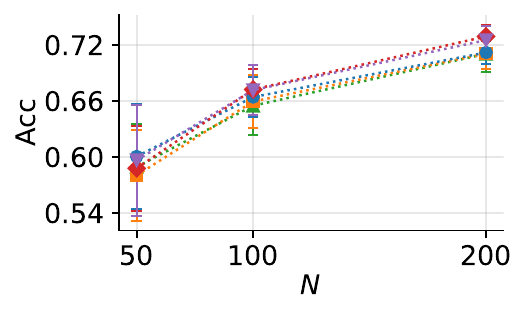}
    \subcaption{}
    \label{fig steel plates acc}
\end{minipage}
\begin{minipage}[b]{0.24\textwidth}
    \centering
    \includegraphics[width=\textwidth]{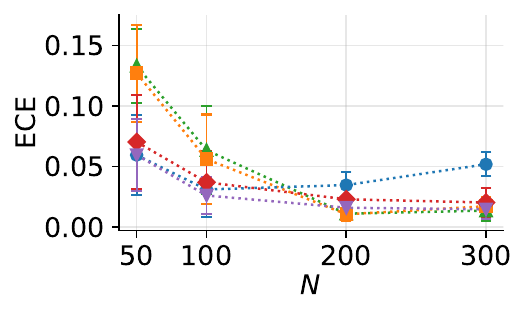}
    \subcaption{}
    \label{fig electrical grid ece}
\end{minipage}
\hfill
\begin{minipage}[b]{0.24\textwidth}
    \centering
    \includegraphics[width=\textwidth]{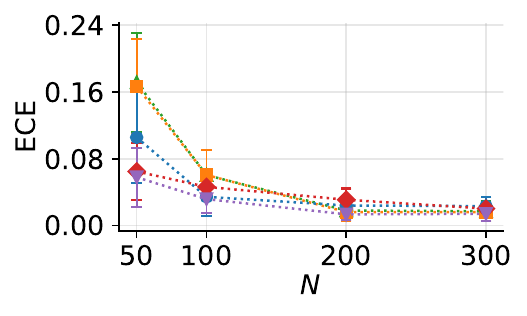}
    \subcaption{}
    \label{fig truss 6d ece}
\end{minipage}
\hfill
\begin{minipage}[b]{0.24\textwidth}
    \centering
    \includegraphics[width=\textwidth]{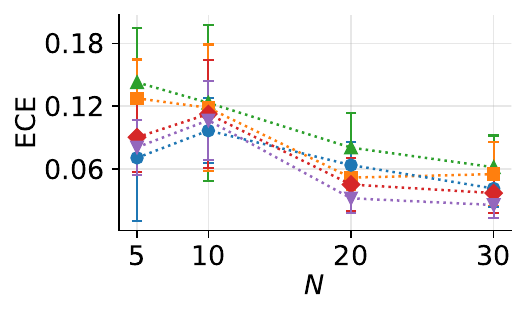}
    \subcaption{}
    \label{fig stellar ece}
\end{minipage}
\hfill
\begin{minipage}[b]{0.24\textwidth}
    \centering
    \includegraphics[width=\textwidth]{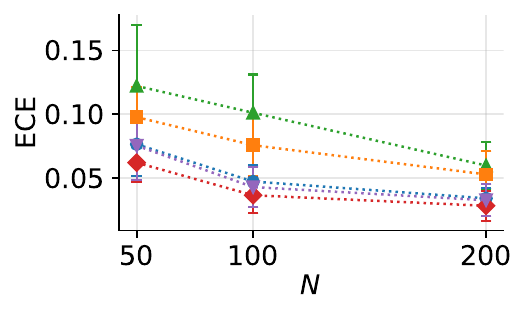}
    \subcaption{}
    \label{fig steel plates ece}
\end{minipage}
\caption{Classification performance on the Electrical Grid Stability, Truss 6D, Stellar, Steel Plates Faults benchmarks for varying training set size $N$. Columns correspond to the datasets in that order and rows to $\mathcal{L}$, Accuracy, ECE, respectively. With the exception of using Adam as the optimizer, GP+ package \textit{default} settings are used for training of all GPs. Error bars denote one standard deviation across 10 seeds.}
\label{subsec classification results}
\end{figure*} 
We use four datasets to compare TabPFN and GP+ with Gaussian, PE, and Mat\'ern ($\nu=0.5$) kernels. We aim to investigate whether small MLE-based training loss leads to small test error and how this relationship is affected across varying training dataset size, number of classes, and kernel choice.

Figures~\ref{fig electrical grid nll}--\ref{fig steel plates nll} report that increasing $N$ decreases $\mathcal{L}$ on all four datasets. The ordering among kernels is consistent. Gaussian and PE kernels follow one another closely while Mat\'ern remains above them at all $N$. The agreement between Gaussian and PE suggests both roughness assumptions suit these problems comparably. Mat\'ern ($\nu=0.5$) assumes a much rougher underlying function that these problems do not exhibit.

The ordering of models with respect to MLE-based loss does not strictly transfer to test performance. It holds only on Electrical Grid Stability. Figure~\ref{fig electrical grid acc} and Figure~\ref{fig electrical grid ece} show that Mat\'ern has the lowest accuracy and worst calibration among the GP variants for $N \geq 200$. Steel Plates Faults reverses this relationship. Figures~\ref{fig steel plates ece} and~\ref{fig steel plates acc} show the Mat\'ern kernel achieving the lowest ECE and marginally the highest accuracy among the GP variants despite having the highest training loss. Figure~\ref{fig stellar ece} similarly subverts the assumption that models with lower training loss exhibit better test performance. MLE-based training loss should therefore not be the sole criterion when deciding between kernels.

On average, TabPFN outperforms GP+ by a small margin across train sizes and number of classes. Electrical Grid Stability is the only purely continuous binary problem shown. In this problem, Figure~\ref{fig electrical grid acc} shows that while TabPFN leads at $N=50$ all GP variants overtake both versions by $N \geq 200$ in accuracy. Figure~\ref{fig electrical grid ece} indicates that by $N \geq 200$ the Gaussian and PE variants exhibit competitive or better calibration. On the remaining three problems TabPFN 2.5 maintains an advantage across the training sizes depicted (Figures~\ref{fig truss 6d acc},~\ref{fig stellar acc},
and~\ref{fig steel plates acc}). These three differ from Electrical Grid Stability in that these problems introduce categorical variables and additional classes. It is worth noting that TabPFN handles categorical variables internally. The gap observed in these three problems may be attributable to this native handling along with the multi-class setting. With that said, TabPFN 2.5's advantage does diminish with increasing $N$ across all four problems.

Overall, GP+ with MLE-based training is a viable and interpretable alternative to TabPFN given that kernel selection and optimization are performed with sufficient care.
    \section{Conclusion} \label{sec conclusion}

In this paper, we investigated the brittleness of MLE in the context of training GPs for engineering design tasks involving regression and classification. We demonstrated that while MLE is the dominant training paradigm for GPs, its effectiveness depends on several assumptions that are frequently violated in practical engineering applications, including correct model specification, sufficient data, and well-behaved optimization landscapes. Through $3,170$ independent simulations spanning many combinations of noise levels, dataset sizes, and input dimensionalities, we showed that these failures are systematic and consequential.

Our core finding is that MLE can produce poorly calibrated GPs particularly in the very small-data, high-dimensional regime where the NLL landscape becomes highly non-convex and the nugget parameter is vulnerable to bad local optima. Simple and practical remedies such as reparameterizing hyperparameters in log-scale, using multiple initializations for optimization, and bounding hyperparameters for numerical stability can substantially mitigate these failure modes without abandoning MLE entirely. When implemented carefully, GPs trained via MLE remain highly competitive and can outperform foundation models such as TabPFN in prediction accuracy, UQ quality, and inference cost.

We also evaluated the LOO log-pseudo-likelihood as an alternative to MLE. While LOO is theoretically more robust to kernel misspecification since it directly approximates generalization error, it increases training cost without consistently outperforming the default MLE settings in GP+. We believe investing in proper MLE implementation is a better use of computational budget than switching loss functions entirely. Kernel flows offer a useful batch-friendly alternative but lack native UQ capabilities and should be combined with likelihood-based objectives in uncertainty-aware applications.

One insight from the comparison with TabPFN is the potential value of meta-learning GP hyperpriors or adopting TabPFN-inspired preprocessing principles. TabPFN's robustness to per-dataset tuning comes from being trained across a broad prior over synthetic datasets, which effectively amortizes model selection. A similar philosophy could be applied to GPs by learning a prior over hyperparameters or kernel families from a large corpus of synthetic engineering tasks, and using it to warm-start or constrain MLE. This could reduce sensitivity to initialization and misspecification while preserving the interpretability and sample efficiency that make GPs attractive.

Our studies were mostly limited to analytic benchmarks with Gaussian observation noise. Future work should extend this analysis to non-Gaussian noise models, real-world engineering datasets, and downstream tasks beyond BO such as multi-fidelity modeling and calibration, where UQ errors can compound across fidelity levels. Additionally, regularized versions of MLE provide an attractive alternative for effectively learning from very small datasets. For a fair comparison, we refrained from fine-tuning the training process and kernel of the GPs when comparing them to TabPFN but it is interesting to evaluate these models under their best settings. Lastly, the classification results warrant a deeper investigation to assess how much of the errors are due to the classification-to-regression conversion rather than GPs' regression ability.
    \section*{Acknowledgments}
We appreciate the support from the Office of the Naval Research (award number N000142312485) and National Science Foundation (award number 2238038).
    \clearpage
\appendix
\onecolumn
\section{Additional Benchmark Results}
\label{app:additional-results}

\renewcommand{\thefigure}{A\arabic{figure}}
\renewcommand{\thetable}{A\arabic{table}}
\setcounter{figure}{0}
\setcounter{table}{0}

\begin{table}[!h]
    \centering
    \captionsetup{font=footnotesize}
    {
    \setlength{\tabcolsep}{2.5pt}
    \renewcommand{\arraystretch}{1.15}
    \tiny
    \caption{RRMSE and NIS statistics across benchmark problems for GP+, GP+ (PE), GP+ (LOO), PFN 2.5, PFN 2.0, and GPyTorch. Entries are median $\pm$ std over runs. Best models for a given problem (lowest median per metric) are in bold.}
    \resizebox{\dimexpr\paperwidth-2in\relax}{!}{%
    \begin{tabular}{>{\centering\arraybackslash}p{1.4cm}>{\centering\arraybackslash}p{0.7cm}>{\centering\arraybackslash}p{0.4cm}>{\centering\arraybackslash}p{0.7cm}|c|c|c|c|c|c!{\vrule width 1.0pt}c|c|c|c|c|c}
    \hline
    & & & & \multicolumn{6}{c!{\vrule width 1.0pt}}{\textbf{RRMSE}} & \multicolumn{6}{c}{\textbf{NIS}} \\
    \hline
    \textbf{Problem} & \textbf{N} & \textbf{D$_x$} & \textbf{Noise} & \textbf{GP+} & \textbf{GP+ (PE)} & \textbf{GP+ (LOO)} & \textbf{PFN 2.5} & \textbf{PFN 2.0} & \textbf{GPyTorch} & \textbf{GP+} & \textbf{GP+ (PE)} & \textbf{GP+ (LOO)} & \textbf{PFN 2.5} & \textbf{PFN 2.0} & \textbf{GPyTorch} \\
    \hline
    \textbf{Buckling} & \textbf{5D$_x$} & \textbf{4} & 0.002 & 0.344 $\pm$ 0.14 & 0.344 $\pm$ 0.14 & 0.519 $\pm$ 0.18 & 0.385 $\pm$ 0.26 & \textbf{0.286 $\pm$ 0.14} & 0.730 $\pm$ 0.15 & 6.940 $\pm$ 3.20 & 6.940 $\pm$ 3.20 & 10.35 $\pm$ 5.09 & \textbf{1.377 $\pm$ 0.24} & 1.524 $\pm$ 0.27 & 8.769 $\pm$ 1.92 \\[2pt]
     &  &  & 0.08 & 0.537 $\pm$ 0.12 & 0.537 $\pm$ 0.12 & 0.851 $\pm$ 0.57 & 0.459 $\pm$ 0.11 & \textbf{0.423 $\pm$ 0.14} & 0.759 $\pm$ 0.17 & 9.781 $\pm$ 3.18 & 9.781 $\pm$ 3.18 & 18.10 $\pm$ 10.4 & \textbf{1.806 $\pm$ 0.27} & 1.902 $\pm$ 0.32 & 9.340 $\pm$ 2.00 \\[2pt]
     & \textbf{20D$_x$} & \textbf{4} & 0.002 & 0.058 $\pm$ 0.02 & 0.058 $\pm$ 0.02 & \textbf{0.049 $\pm$ 0.02} & 0.346 $\pm$ 0.27 & 0.205 $\pm$ 0.25 & 0.261 $\pm$ 0.07 & \textbf{0.198 $\pm$ 0.06} & \textbf{0.198 $\pm$ 0.06} & 0.235 $\pm$ 0.16 & 0.354 $\pm$ 0.10 & 0.476 $\pm$ 0.15 & 1.438 $\pm$ 0.58 \\[2pt]
     &  &  & 0.08 & \textbf{0.226 $\pm$ 0.06} & \textbf{0.226 $\pm$ 0.06} & 0.249 $\pm$ 0.23 & 0.585 $\pm$ 0.35 & 0.382 $\pm$ 0.16 & 0.335 $\pm$ 0.08 & 1.139 $\pm$ 0.49 & 1.139 $\pm$ 0.49 & 1.359 $\pm$ 2.92 & \textbf{0.981 $\pm$ 0.17} & 1.285 $\pm$ 0.22 & 1.847 $\pm$ 0.72 \\[2pt]
    \hline
    \textbf{Borehole} & \textbf{5D$_x$} & \textbf{8} & 0.002 & \textbf{0.024 $\pm$ 0.01} & 0.027 $\pm$ 0.01 & 0.044 $\pm$ 0.01 & 0.049 $\pm$ 0.01 & 0.065 $\pm$ 0.01 & 0.027 $\pm$ 0.00 & \textbf{0.131 $\pm$ 0.04} & 0.139 $\pm$ 0.05 & 0.645 $\pm$ 0.30 & 0.271 $\pm$ 0.03 & 0.345 $\pm$ 0.05 & 0.138 $\pm$ 0.04 \\[2pt]
     &  &  & 0.08 & \textbf{0.120 $\pm$ 0.02} & 0.121 $\pm$ 0.02 & 0.259 $\pm$ 0.12 & 0.127 $\pm$ 0.02 & 0.132 $\pm$ 0.02 & 0.131 $\pm$ 0.04 & 0.653 $\pm$ 0.13 & 0.653 $\pm$ 0.21 & 5.299 $\pm$ 2.46 & \textbf{0.631 $\pm$ 0.12} & 0.647 $\pm$ 0.13 & 0.666 $\pm$ 0.59 \\[2pt]
     & \textbf{20D$_x$} & \textbf{8} & 0.002 & \textbf{0.005 $\pm$ 0.00} & \textbf{0.005 $\pm$ 0.00} & \textbf{0.005 $\pm$ 0.00} & 0.012 $\pm$ 0.00 & 0.024 $\pm$ 0.01 & 0.008 $\pm$ 0.00 & \textbf{0.023 $\pm$ 0.00} & \textbf{0.023 $\pm$ 0.00} & 0.028 $\pm$ 0.02 & 0.059 $\pm$ 0.00 & 0.098 $\pm$ 0.01 & 0.053 $\pm$ 0.01 \\[2pt]
     &  &  & 0.08 & 0.091 $\pm$ 0.00 & 0.091 $\pm$ 0.00 & 0.212 $\pm$ 0.03 & \textbf{0.090 $\pm$ 0.00} & \textbf{0.090 $\pm$ 0.00} & 0.091 $\pm$ 0.00 & \textbf{0.420 $\pm$ 0.02} & \textbf{0.420 $\pm$ 0.02} & 0.955 $\pm$ 0.05 & 0.437 $\pm$ 0.01 & 0.438 $\pm$ 0.01 & 0.423 $\pm$ 0.01 \\[2pt]
    \hline
    \textbf{Wing Weight} & \textbf{5D$_x$} & \textbf{10} & 0.002 & \textbf{0.034 $\pm$ 0.01} & 0.043 $\pm$ 0.01 & 0.057 $\pm$ 0.02 & 0.058 $\pm$ 0.01 & 0.085 $\pm$ 0.01 & 0.042 $\pm$ 0.01 & \textbf{0.185 $\pm$ 0.22} & 0.206 $\pm$ 0.21 & 1.107 $\pm$ 0.58 & 0.328 $\pm$ 0.04 & 0.461 $\pm$ 0.05 & 0.218 $\pm$ 0.18 \\[2pt]
     &  &  & 0.08 & \textbf{0.116 $\pm$ 0.01} & 0.117 $\pm$ 0.01 & 0.232 $\pm$ 0.04 & 0.119 $\pm$ 0.01 & 0.129 $\pm$ 0.01 & 0.121 $\pm$ 0.03 & \textbf{0.553 $\pm$ 0.10} & 0.560 $\pm$ 0.09 & 3.676 $\pm$ 1.57 & 0.618 $\pm$ 0.04 & 0.630 $\pm$ 0.06 & 0.583 $\pm$ 0.68 \\[2pt]
     & \textbf{20D$_x$} & \textbf{10} & 0.002 & 0.006 $\pm$ 0.00 & 0.006 $\pm$ 0.00 & \textbf{0.005 $\pm$ 0.00} & 0.023 $\pm$ 0.00 & 0.035 $\pm$ 0.02 & 0.009 $\pm$ 0.00 & 0.027 $\pm$ 0.00 & 0.026 $\pm$ 0.00 & \textbf{0.023 $\pm$ 0.00} & 0.117 $\pm$ 0.01 & 0.166 $\pm$ 0.02 & 0.052 $\pm$ 0.00 \\[2pt]
     &  &  & 0.08 & \textbf{0.092 $\pm$ 0.00} & \textbf{0.092 $\pm$ 0.00} & 0.095 $\pm$ 0.01 & 0.094 $\pm$ 0.00 & 0.094 $\pm$ 0.00 & \textbf{0.092 $\pm$ 0.00} & 0.429 $\pm$ 0.01 & 0.428 $\pm$ 0.01 & 0.446 $\pm$ 0.05 & 0.431 $\pm$ 0.01 & 0.442 $\pm$ 0.01 & \textbf{0.427 $\pm$ 0.02} \\[2pt]
    \hline
    \textbf{Ackley} & \textbf{5D$_x$} & \textbf{20} & 0.002 & 0.469 $\pm$ 0.05 & 0.466 $\pm$ 0.03 & 0.782 $\pm$ 0.35 & 0.684 $\pm$ 0.04 & 0.622 $\pm$ 0.01 & \textbf{0.448 $\pm$ 0.06} & 2.604 $\pm$ 0.57 & 2.498 $\pm$ 0.32 & 12.82 $\pm$ 6.73 & 3.466 $\pm$ 0.12 & 3.199 $\pm$ 0.10 & \textbf{2.468 $\pm$ 0.51} \\[2pt]
     &  &  & 0.08 & 0.482 $\pm$ 0.03 & 0.485 $\pm$ 0.04 & 1.001 $\pm$ 0.16 & 0.693 $\pm$ 0.04 & 0.629 $\pm$ 0.01 & \textbf{0.477 $\pm$ 0.06} & \textbf{2.655 $\pm$ 0.35} & 2.716 $\pm$ 0.54 & 16.65 $\pm$ 0.93 & 3.473 $\pm$ 0.12 & 3.226 $\pm$ 0.08 & 2.760 $\pm$ 0.55 \\[2pt]
     & \textbf{20D$_x$} & \textbf{20} & 0.002 & 0.248 $\pm$ 0.01 & 0.258 $\pm$ 0.01 & \textbf{0.213 $\pm$ 0.00} & 0.251 $\pm$ 0.01 & 0.305 $\pm$ 0.02 & 0.247 $\pm$ 0.01 & 1.235 $\pm$ 0.03 & 1.276 $\pm$ 0.06 & \textbf{1.054 $\pm$ 0.03} & 1.201 $\pm$ 0.04 & 1.568 $\pm$ 0.07 & 1.240 $\pm$ 0.03 \\[2pt]
     &  &  & 0.08 & 0.269 $\pm$ 0.01 & 0.276 $\pm$ 0.01 & \textbf{0.235 $\pm$ 0.01} & 0.270 $\pm$ 0.01 & 0.313 $\pm$ 0.02 & 0.269 $\pm$ 0.01 & 1.323 $\pm$ 0.03 & 1.376 $\pm$ 0.06 & \textbf{1.175 $\pm$ 0.04} & 1.280 $\pm$ 0.05 & 1.600 $\pm$ 0.08 & 1.334 $\pm$ 0.04 \\[2pt]
    \hline
    \textbf{Griewank} & \textbf{5D$_x$} & \textbf{20} & 0.002 & 0.846 $\pm$ 0.11 & 0.839 $\pm$ 0.08 & 1.040 $\pm$ 0.51 & 0.872 $\pm$ 0.04 & 0.905 $\pm$ 0.04 & \textbf{0.805 $\pm$ 0.05} & 6.114 $\pm$ 1.73 & 5.769 $\pm$ 1.57 & 16.37 $\pm$ 8.12 & \textbf{4.203 $\pm$ 0.19} & 4.321 $\pm$ 0.19 & 5.296 $\pm$ 0.85 \\[2pt]
     &  &  & 0.08 & 0.826 $\pm$ 0.10 & 0.837 $\pm$ 0.10 & 1.013 $\pm$ 0.46 & 0.875 $\pm$ 0.04 & 0.911 $\pm$ 0.04 & \textbf{0.805 $\pm$ 0.06} & 5.277 $\pm$ 1.61 & 5.272 $\pm$ 1.60 & 14.45 $\pm$ 7.22 & \textbf{4.220 $\pm$ 0.19} & 4.354 $\pm$ 0.21 & 5.379 $\pm$ 1.08 \\[2pt]
     & \textbf{20D$_x$} & \textbf{20} & 0.002 & 0.004 $\pm$ 0.00 & 0.018 $\pm$ 0.01 & \textbf{0.002 $\pm$ 0.00} & 0.212 $\pm$ 0.03 & 0.374 $\pm$ 0.03 & 0.003 $\pm$ 0.00 & 0.063 $\pm$ 0.00 & 0.216 $\pm$ 0.06 & \textbf{0.010 $\pm$ 0.01} & 1.082 $\pm$ 0.10 & 1.994 $\pm$ 0.08 & 0.067 $\pm$ 0.00 \\[2pt]
     &  &  & 0.08 & 0.128 $\pm$ 0.01 & 0.114 $\pm$ 0.00 & \textbf{0.085 $\pm$ 0.01} & 0.233 $\pm$ 0.03 & 0.373 $\pm$ 0.03 & 0.135 $\pm$ 0.01 & 0.608 $\pm$ 0.04 & 0.543 $\pm$ 0.02 & \textbf{0.412 $\pm$ 0.04} & 1.167 $\pm$ 0.10 & 2.017 $\pm$ 0.09 & 0.644 $\pm$ 0.04 \\[2pt]
    \hline
    \textbf{Zakharov} & \textbf{5D$_x$} & \textbf{20} & 0.002 & 0.596 $\pm$ 0.14 & 0.601 $\pm$ 0.12 & 0.991 $\pm$ 0.21 & 0.288 $\pm$ 0.17 & \textbf{0.221 $\pm$ 0.08} & 0.572 $\pm$ 0.10 & 4.365 $\pm$ 2.09 & 4.354 $\pm$ 1.91 & 15.58 $\pm$ 2.61 & \textbf{0.769 $\pm$ 0.18} & 1.089 $\pm$ 0.16 & 4.292 $\pm$ 1.43 \\[2pt]
     &  &  & 0.08 & 0.620 $\pm$ 0.13 & 0.619 $\pm$ 0.14 & 0.937 $\pm$ 0.20 & 0.426 $\pm$ 0.13 & \textbf{0.272 $\pm$ 0.08} & 0.598 $\pm$ 0.07 & 4.489 $\pm$ 1.95 & 4.864 $\pm$ 1.95 & 12.98 $\pm$ 2.22 & \textbf{1.618 $\pm$ 0.17} & 1.653 $\pm$ 0.19 & 4.365 $\pm$ 1.11 \\[2pt]
     & \textbf{20D$_x$} & \textbf{20} & 0.002 & 0.240 $\pm$ 0.02 & 0.240 $\pm$ 0.01 & 0.284 $\pm$ 0.02 & \textbf{0.122 $\pm$ 0.07} & 0.149 $\pm$ 0.05 & 0.239 $\pm$ 0.01 & 1.344 $\pm$ 0.12 & 1.344 $\pm$ 0.12 & 1.802 $\pm$ 0.21 & \textbf{0.237 $\pm$ 0.03} & 0.453 $\pm$ 0.05 & 1.342 $\pm$ 0.12 \\[2pt]
     &  &  & 0.08 & 0.274 $\pm$ 0.02 & 0.274 $\pm$ 0.02 & 0.347 $\pm$ 0.03 & 0.260 $\pm$ 0.12 & \textbf{0.188 $\pm$ 0.07} & 0.273 $\pm$ 0.02 & 1.489 $\pm$ 0.13 & 1.489 $\pm$ 0.13 & 2.438 $\pm$ 0.50 & \textbf{0.634 $\pm$ 0.04} & 0.845 $\pm$ 0.05 & 1.492 $\pm$ 0.14 \\[2pt]
    \hline
    \textbf{Ackley} & \textbf{5D$_x$} & \textbf{40} & 0.002 & 0.450 $\pm$ 0.06 & 0.433 $\pm$ 0.05 & \textbf{0.354 $\pm$ 0.33} & 0.668 $\pm$ 0.04 & 0.618 $\pm$ 0.02 & 1.002 $\pm$ 0.00 & 2.273 $\pm$ 0.46 & \textbf{2.210 $\pm$ 0.37} & 2.604 $\pm$ 5.78 & 3.347 $\pm$ 0.15 & 2.990 $\pm$ 0.08 & 4.756 $\pm$ 0.07 \\[2pt]
     &  &  & 0.08 & 0.465 $\pm$ 0.04 & 0.467 $\pm$ 0.04 & \textbf{0.400 $\pm$ 0.21} & 0.674 $\pm$ 0.04 & 0.622 $\pm$ 0.02 & 1.002 $\pm$ 0.00 & \textbf{2.358 $\pm$ 0.37} & 2.387 $\pm$ 0.33 & 3.628 $\pm$ 3.42 & 3.352 $\pm$ 0.14 & 3.005 $\pm$ 0.08 & 4.755 $\pm$ 0.07 \\[2pt]
     & \textbf{20D$_x$} & \textbf{40} & 0.002 & 0.258 $\pm$ 0.00 & 0.244 $\pm$ 0.00 & \textbf{0.209 $\pm$ 0.00} & 0.291 $\pm$ 0.01 & 0.383 $\pm$ 0.01 & 1.000 $\pm$ 0.00 & 1.239 $\pm$ 0.02 & 1.186 $\pm$ 0.03 & \textbf{1.019 $\pm$ 0.02} & 1.358 $\pm$ 0.04 & 1.867 $\pm$ 0.05 & 4.771 $\pm$ 0.03 \\[2pt]
     &  &  & 0.08 & 0.276 $\pm$ 0.00 & 0.260 $\pm$ 0.00 & \textbf{0.228 $\pm$ 0.00} & 0.305 $\pm$ 0.01 & 0.401 $\pm$ 0.02 & 1.000 $\pm$ 0.00 & 1.333 $\pm$ 0.03 & 1.268 $\pm$ 0.03 & \textbf{1.113 $\pm$ 0.02} & 1.435 $\pm$ 0.04 & 1.928 $\pm$ 0.05 & 4.770 $\pm$ 0.02 \\[2pt]
    \hline
    \textbf{Dixon-Price} & \textbf{5D$_x$} & \textbf{40} & 0.002 & \textbf{0.663 $\pm$ 0.07} & 0.698 $\pm$ 0.06 & 0.669 $\pm$ 0.25 & 0.767 $\pm$ 0.04 & 0.789 $\pm$ 0.04 & 1.002 $\pm$ 0.00 & \textbf{3.475 $\pm$ 0.77} & 3.909 $\pm$ 0.69 & 5.710 $\pm$ 4.78 & 3.585 $\pm$ 0.27 & 3.756 $\pm$ 0.27 & 4.706 $\pm$ 0.06 \\[2pt]
     &  &  & 0.08 & 0.699 $\pm$ 0.06 & 0.704 $\pm$ 0.11 & \textbf{0.661 $\pm$ 0.26} & 0.775 $\pm$ 0.05 & 0.806 $\pm$ 0.04 & 1.002 $\pm$ 0.00 & 3.889 $\pm$ 0.63 & 3.978 $\pm$ 0.92 & 5.946 $\pm$ 4.22 & \textbf{3.624 $\pm$ 0.29} & 3.793 $\pm$ 0.29 & 4.716 $\pm$ 0.06 \\[2pt]
     & \textbf{20D$_x$} & \textbf{40} & 0.002 & 0.382 $\pm$ 0.01 & \textbf{0.106 $\pm$ 0.00} & 0.320 $\pm$ 0.01 & 0.310 $\pm$ 0.01 & 0.402 $\pm$ 0.01 & 1.000 $\pm$ 0.00 & 1.769 $\pm$ 0.06 & \textbf{0.501 $\pm$ 0.01} & 1.495 $\pm$ 0.04 & 1.512 $\pm$ 0.04 & 1.925 $\pm$ 0.06 & 4.711 $\pm$ 0.02 \\[2pt]
     &  &  & 0.08 & 0.392 $\pm$ 0.01 & \textbf{0.154 $\pm$ 0.00} & 0.333 $\pm$ 0.01 & 0.322 $\pm$ 0.01 & 0.410 $\pm$ 0.01 & 1.000 $\pm$ 0.00 & 1.817 $\pm$ 0.06 & \textbf{0.736 $\pm$ 0.02} & 1.547 $\pm$ 0.04 & 1.570 $\pm$ 0.04 & 1.961 $\pm$ 0.05 & 4.714 $\pm$ 0.02 \\[2pt]
    \hline
    \textbf{Rosenbrock} & \textbf{5D$_x$} & \textbf{80} & 0.002 & 0.661 $\pm$ 0.04 & 0.679 $\pm$ 0.17 & \textbf{0.521 $\pm$ 0.20} & 0.767 $\pm$ 0.02 & 0.714 $\pm$ 0.01 & 1.000 $\pm$ 0.00 & 3.578 $\pm$ 0.46 & 3.701 $\pm$ 0.77 & 3.483 $\pm$ 3.62 & 3.621 $\pm$ 0.11 & \textbf{3.374 $\pm$ 0.07} & 4.701 $\pm$ 0.01 \\[2pt]
     &  &  & 0.08 & \textbf{0.671 $\pm$ 0.05} & 0.695 $\pm$ 0.14 & 0.717 $\pm$ 0.28 & 0.772 $\pm$ 0.02 & 0.715 $\pm$ 0.01 & 1.001 $\pm$ 0.00 & 3.634 $\pm$ 0.46 & 3.798 $\pm$ 0.66 & 5.980 $\pm$ 4.74 & 3.661 $\pm$ 0.10 & \textbf{3.388 $\pm$ 0.07} & 4.707 $\pm$ 0.01 \\[2pt]
     & \textbf{20D$_x$} & \textbf{80} & 0.002 & 0.450 $\pm$ 0.01 & \textbf{0.189 $\pm$ 0.64} & 2.460 $\pm$ 2.17 & 0.336 $\pm$ 0.01 & 0.516 $\pm$ 0.02 & 1.000 $\pm$ 0.00 & 2.138 $\pm$ 0.07 & 2.261 $\pm$ 10.6 & 53.06 $\pm$ 68.5 & \textbf{1.682 $\pm$ 0.02} & 2.455 $\pm$ 0.06 & 4.698 $\pm$ 0.01 \\[2pt]
     &  &  & 0.08 & 0.465 $\pm$ 0.01 & \textbf{0.285 $\pm$ 0.34} & 6.411 $\pm$ 17.9 & 0.347 $\pm$ 0.01 & 0.517 $\pm$ 0.02 & 1.000 $\pm$ 0.00 & 2.184 $\pm$ 0.06 & 2.329 $\pm$ 3.37 & 181.5 $\pm$ 681.1 & \textbf{1.724 $\pm$ 0.03} & 2.448 $\pm$ 0.07 & 4.698 $\pm$ 0.01 \\[2pt]
    \bottomrule
    \end{tabular}
    }
    }
\end{table}



\begin{figure*}[!h]
    \centering
    \captionsetup{font=footnotesize}
    \begin{minipage}[b]{0.24\textwidth}
        \centering
        \includegraphics[width=\textwidth]{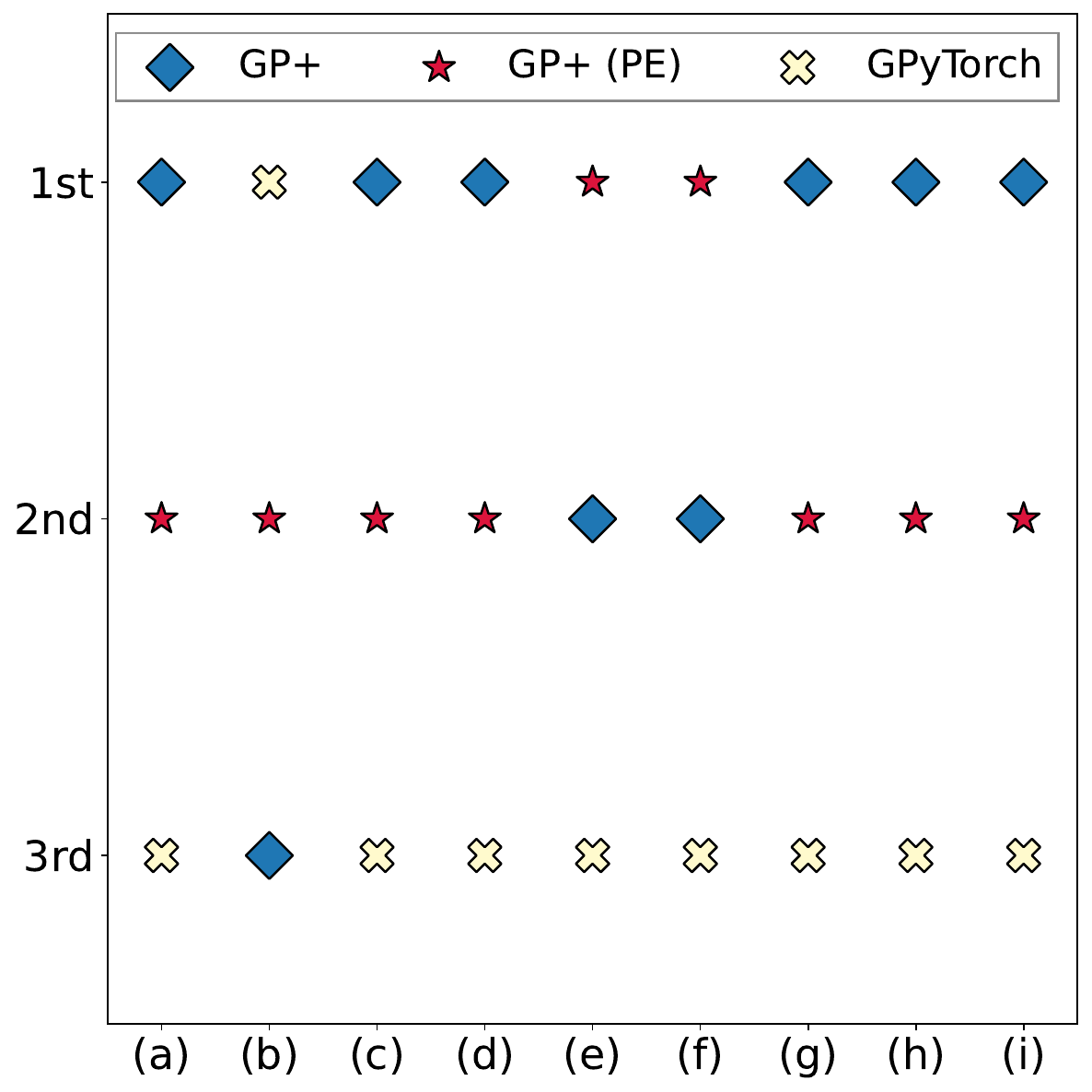}
        \subcaption{$N=5D_x$, $\varepsilon\sim\mathcal{N}(0, (0.002\times c)^2)$}
        \label{fig nll rank 5Dx low}
    \end{minipage}
    \hfill
    \begin{minipage}[b]{0.24\textwidth}
        \centering
        \includegraphics[width=\textwidth]{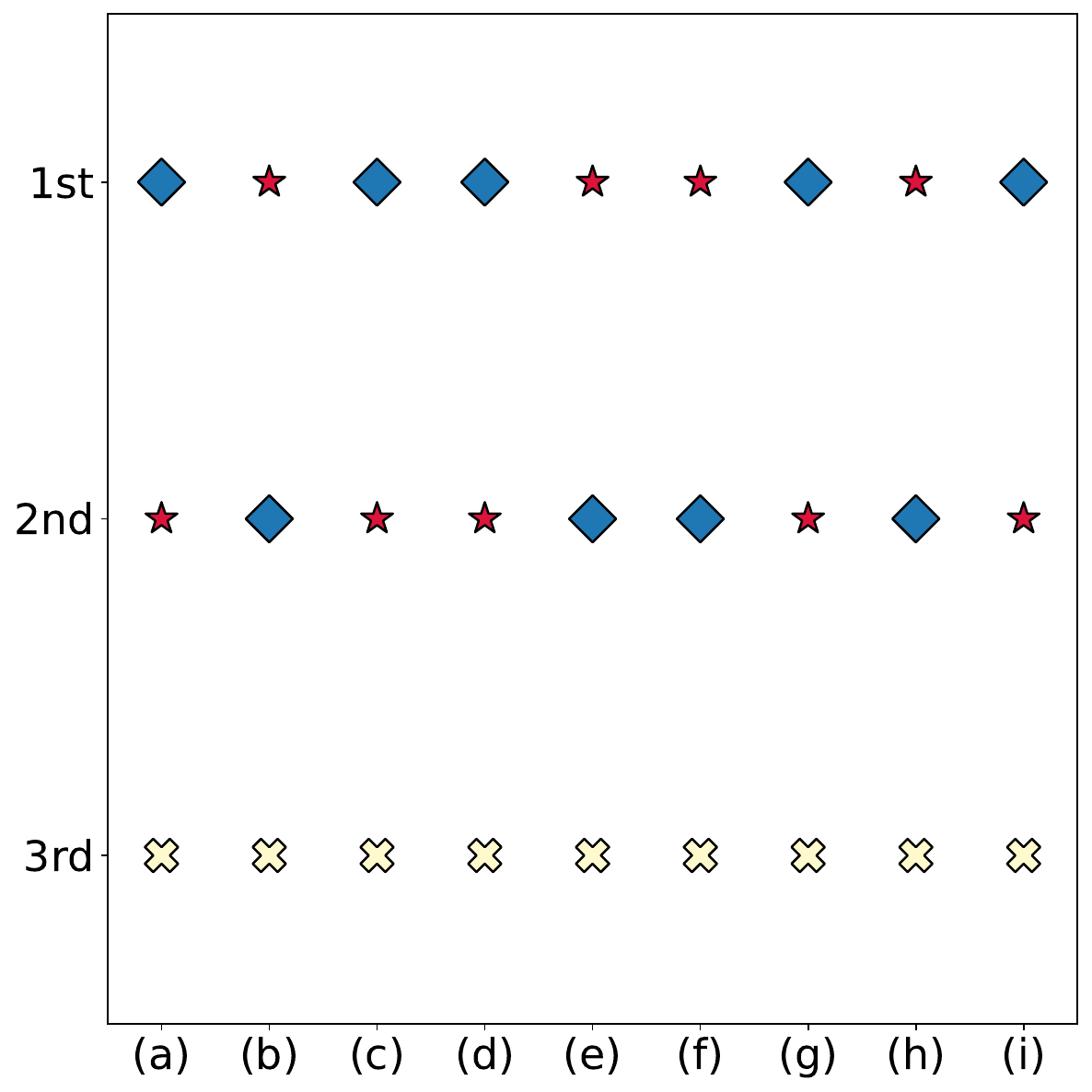}
        \subcaption{$N=5D_x$, $\varepsilon \sim \mathcal{N}(0, (0.08\times c)^2)$}
        \label{fig nll rank 5Dx high}
    \end{minipage}
    \hfill
    \begin{minipage}[b]{0.24\textwidth}
        \centering
        \includegraphics[width=\textwidth]{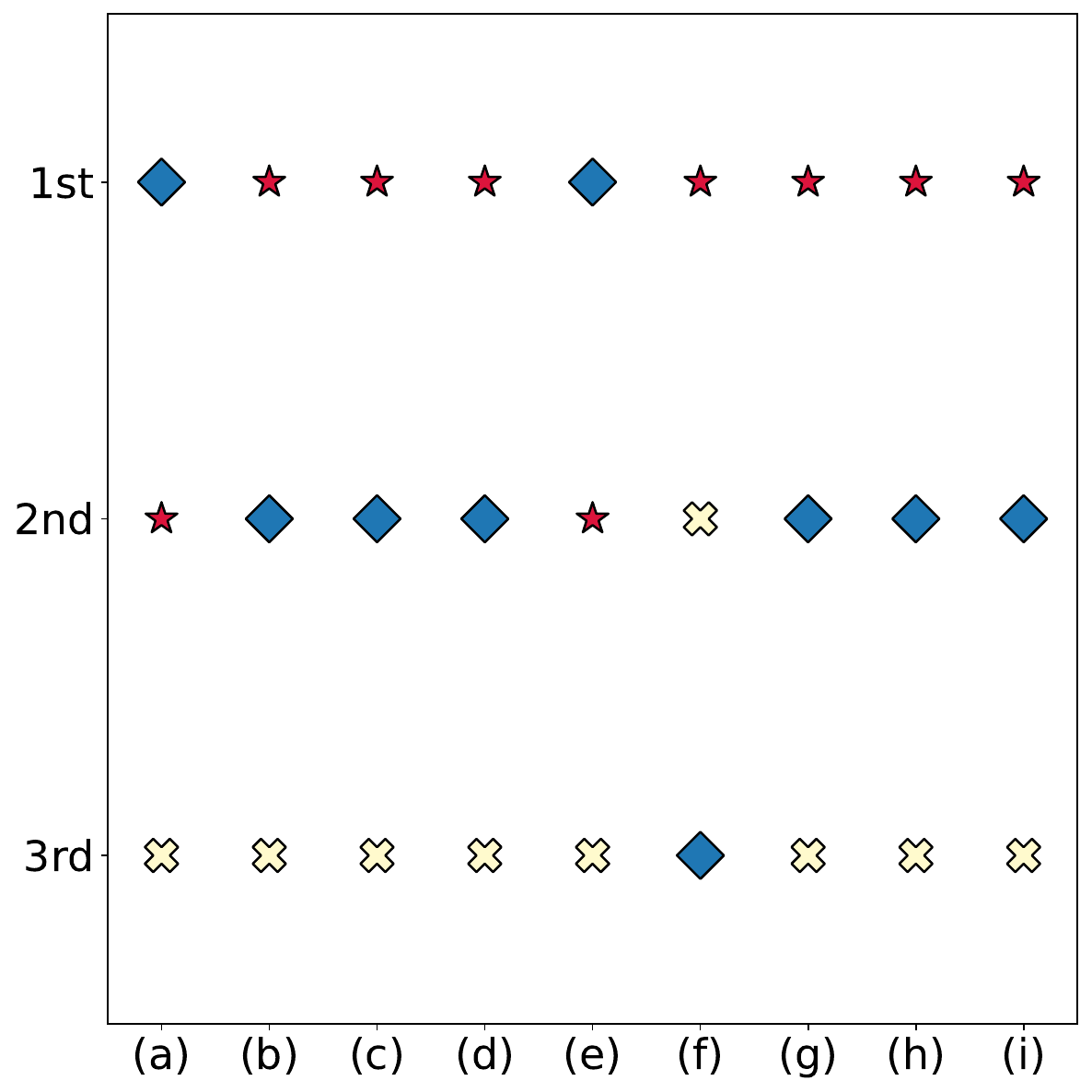}
        \subcaption{$N=20D_x$, $\varepsilon \sim \mathcal{N}(0, (0.002\times c)^2)$}
        \label{fig nll rank 20Dx low}
    \end{minipage}
    \hfill
    \begin{minipage}[b]{0.24\textwidth}
        \centering
        \includegraphics[width=\textwidth]{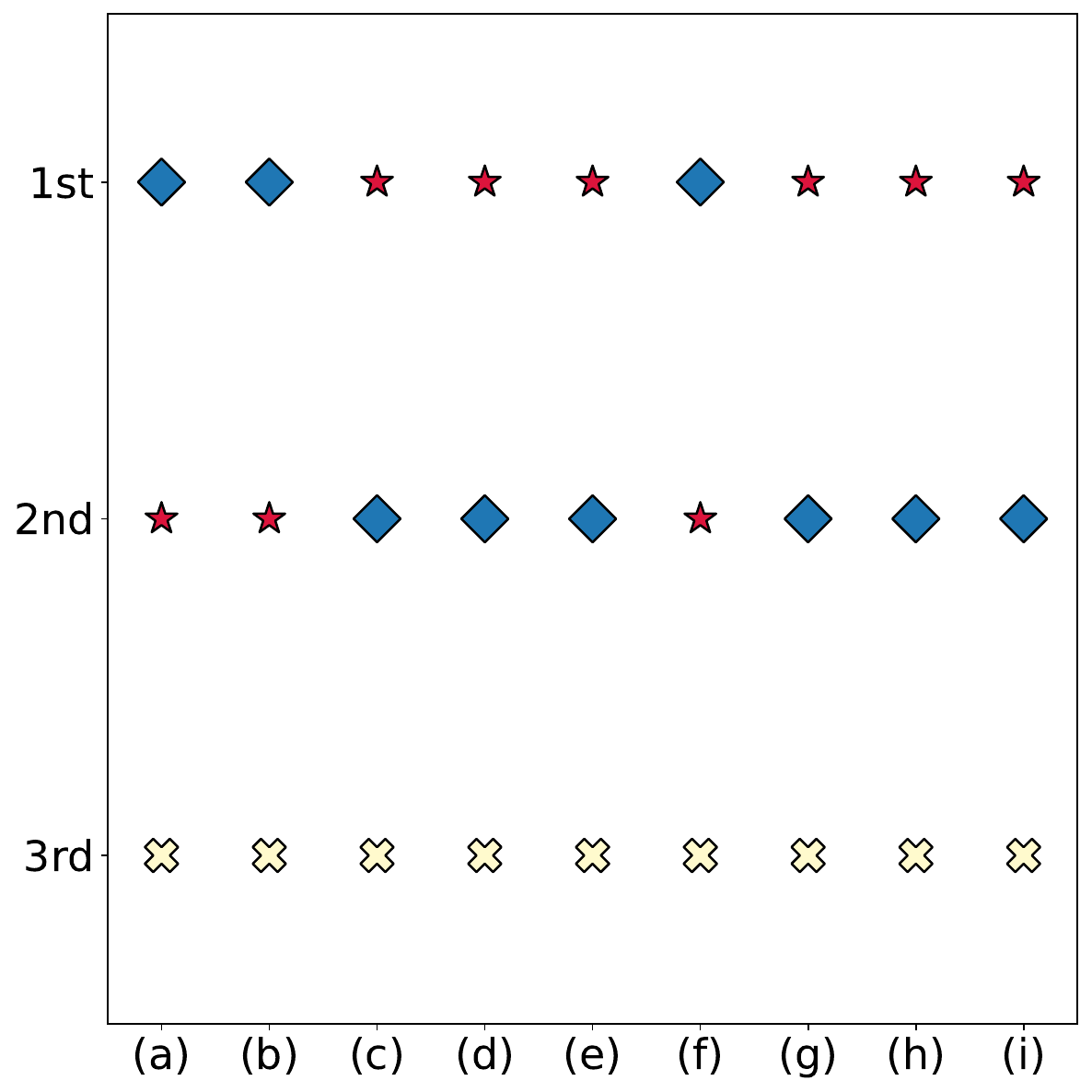}
        \subcaption{$N=20D_x$, $\varepsilon \sim \mathcal{N}(0, (0.08\times c)^2)$}
        \label{fig nll rank 20Dx high}
    \end{minipage}
    \caption{Ranking of Models by NLL: These plots rank the training NLL of GP+, GP+ (PE), and GPyTorch on problems (a)-(i), where (a) is Buckling, (b) is Borehole, (c) is Wing Weight, (d) is Ackley $D_x=20$, (e) is Griewank $D_x=20$, (f) is Zakharov $D_x=20$, (g) is Ackley $D_x=40$, (h) is Dixon-Price $D_x=40$, and (i) is Rosenbrock $D_x=80$.}

    \label{fig nll rankings}
\end{figure*}

\begin{table*}[!t]
    \centering
    
    \setlength{\tabcolsep}{2.5pt}
    \renewcommand{\arraystretch}{1.15}
    \tiny
    \caption{Training and inference time (median $\pm$ std) for GP+, GP+ (PE), GP+ (LOO), PFN 2.5, PFN 2.0, and GPyTorch across benchmark problems.}
    \resizebox{\dimexpr\paperwidth-2in\relax}{!}{%
    \begin{tabular}{>{\centering\arraybackslash}p{1.4cm}>{\centering\arraybackslash}p{0.7cm}>{\centering\arraybackslash}p{0.4cm}>{\centering\arraybackslash}p{0.7cm}|c|c|c|c|c|c|c|c|c|c|c|c}
    \hline
    \multicolumn{4}{c|}{} & \multicolumn{2}{c|}{\textbf{GP+}} & \multicolumn{2}{c|}{\textbf{GP+ (PE)}} & \multicolumn{2}{c|}{\textbf{GP+ (LOO)}} & \multicolumn{2}{c|}{\textbf{PFN 2.5}} & \multicolumn{2}{c|}{\textbf{PFN 2.0}} & \multicolumn{2}{c}{\textbf{GPyTorch}} \\
    \hline
    \textbf{Problem} & \textbf{N} & \textbf{D$_x$} & \textbf{Noise} & \textbf{Train (s)} & \textbf{Pred (s)} & \textbf{Train (s)} & \textbf{Pred (s)} & \textbf{Train (s)} & \textbf{Pred (s)} & \textbf{Fit (s)} & \textbf{Inf (s)} & \textbf{Fit (s)} & \textbf{Inf (s)} & \textbf{Train (s)} & \textbf{Pred (s)} \\
    \hline
    \textbf{Buckling} & \textbf{5D$_x$} & \textbf{4} & 0.002 & 5.23 $\pm$ 1.16 & 0.011 $\pm$ 0.002 & 4.55 $\pm$ 1.01 & 0.011 $\pm$ 0.037 & 7.14 $\pm$ 0.97 & 0.010 $\pm$ 0.003 & 0.17 $\pm$ 0.07 & 1.05 $\pm$ 0.06 & 0.17 $\pm$ 0.02 & 1.23 $\pm$ 0.04 & 1.54 $\pm$ 0.30 & 0.006 $\pm$ 0.004 \\[2pt]
     &  &  & 0.08 & 5.54 $\pm$ 1.59 & 0.012 $\pm$ 0.002 & 4.79 $\pm$ 1.35 & 0.011 $\pm$ 0.002 & 6.88 $\pm$ 0.93 & 0.010 $\pm$ 0.002 & 0.16 $\pm$ 0.15 & 1.05 $\pm$ 0.04 & 0.15 $\pm$ 0.03 & 1.21 $\pm$ 0.06 & 1.63 $\pm$ 0.49 & 0.008 $\pm$ 0.002 \\[2pt]
    \textbf{Buckling} & \textbf{20D$_x$} & \textbf{4} & 0.002 & 6.81 $\pm$ 1.37 & 0.016 $\pm$ 0.001 & 5.87 $\pm$ 1.25 & 0.017 $\pm$ 0.003 & 6.15 $\pm$ 1.36 & 0.015 $\pm$ 0.002 & 0.18 $\pm$ 0.04 & 1.04 $\pm$ 0.06 & 0.18 $\pm$ 0.04 & 1.24 $\pm$ 0.04 & 2.62 $\pm$ 0.62 & 0.010 $\pm$ 0.004 \\[2pt]
     &  &  & 0.08 & 2.92 $\pm$ 1.98 & 0.017 $\pm$ 0.003 & 2.73 $\pm$ 1.73 & 0.014 $\pm$ 0.002 & 6.03 $\pm$ 2.15 & 0.015 $\pm$ 0.003 & 0.17 $\pm$ 0.02 & 1.06 $\pm$ 0.04 & 0.15 $\pm$ 0.03 & 1.20 $\pm$ 0.04 & 2.08 $\pm$ 0.41 & 0.008 $\pm$ 0.001 \\[2pt]
    \hline
    \textbf{Borehole} & \textbf{5D$_x$} & \textbf{8} & 0.002 & 0.50 $\pm$ 0.10 & 0.006 $\pm$ 0.002 & 0.68 $\pm$ 0.08 & 0.013 $\pm$ 0.002 & 0.92 $\pm$ 0.18 & 0.009 $\pm$ 0.002 & 0.17 $\pm$ 0.02 & 1.21 $\pm$ 0.06 & 0.19 $\pm$ 0.04 & 1.43 $\pm$ 0.05 & 2.47 $\pm$ 0.21 & 0.006 $\pm$ 0.001 \\[2pt]
     &  &  & 0.08 & 0.54 $\pm$ 0.08 & 0.005 $\pm$ 0.002 & 0.65 $\pm$ 0.10 & 0.010 $\pm$ 0.004 & 0.76 $\pm$ 0.14 & 0.010 $\pm$ 0.002 & 0.18 $\pm$ 0.01 & 1.19 $\pm$ 0.04 & 0.19 $\pm$ 0.02 & 1.43 $\pm$ 0.03 & 1.94 $\pm$ 0.44 & 0.006 $\pm$ 0.002 \\[2pt]
    \textbf{Borehole} & \textbf{20D$_x$} & \textbf{8} & 0.002 & 0.74 $\pm$ 0.11 & 0.012 $\pm$ 0.004 & 2.93 $\pm$ 0.43 & 0.039 $\pm$ 0.003 & 1.60 $\pm$ 0.45 & 0.014 $\pm$ 0.003 & 0.17 $\pm$ 0.02 & 1.22 $\pm$ 0.04 & 0.19 $\pm$ 0.02 & 1.44 $\pm$ 0.05 & 5.36 $\pm$ 0.77 & 0.011 $\pm$ 0.002 \\[2pt]
     &  &  & 0.08 & 0.84 $\pm$ 0.15 & 0.012 $\pm$ 0.003 & 3.21 $\pm$ 0.50 & 0.038 $\pm$ 0.003 & 1.27 $\pm$ 0.07 & 0.015 $\pm$ 0.001 & 0.20 $\pm$ 0.05 & 1.17 $\pm$ 0.12 & 0.18 $\pm$ 0.01 & 1.47 $\pm$ 0.05 & 5.12 $\pm$ 0.71 & 0.012 $\pm$ 0.004 \\[2pt]
    \hline
    \textbf{Wing Weight} & \textbf{5D$_x$} & \textbf{10} & 0.002 & 0.69 $\pm$ 0.11 & 0.011 $\pm$ 0.002 & 0.88 $\pm$ 0.11 & 0.017 $\pm$ 0.002 & 0.75 $\pm$ 0.12 & 0.010 $\pm$ 0.002 & 0.18 $\pm$ 0.04 & 1.22 $\pm$ 0.04 & 0.18 $\pm$ 0.02 & 1.51 $\pm$ 0.04 & 3.41 $\pm$ 0.45 & 0.008 $\pm$ 0.002 \\[2pt]
     &  &  & 0.08 & 0.73 $\pm$ 0.10 & 0.008 $\pm$ 0.002 & 0.95 $\pm$ 0.17 & 0.018 $\pm$ 0.003 & 1.49 $\pm$ 0.42 & 0.015 $\pm$ 0.005 & 0.17 $\pm$ 0.04 & 1.21 $\pm$ 0.04 & 0.19 $\pm$ 0.02 & 1.48 $\pm$ 0.04 & 2.13 $\pm$ 0.63 & 0.007 $\pm$ 0.002 \\[2pt]
    \textbf{Wing Weight} & \textbf{20D$_x$} & \textbf{10} & 0.002 & 1.24 $\pm$ 0.23 & 0.014 $\pm$ 0.002 & 6.36 $\pm$ 0.68 & 0.054 $\pm$ 0.005 & 6.83 $\pm$ 1.33 & 0.032 $\pm$ 0.006 & 0.19 $\pm$ 0.03 & 1.22 $\pm$ 0.05 & 0.19 $\pm$ 0.05 & 1.56 $\pm$ 0.04 & 8.17 $\pm$ 1.25 & 0.016 $\pm$ 0.004 \\[2pt]
     &  &  & 0.08 & 1.42 $\pm$ 0.21 & 0.016 $\pm$ 0.002 & 7.06 $\pm$ 0.83 & 0.052 $\pm$ 0.003 & 2.79 $\pm$ 0.42 & 0.017 $\pm$ 0.002 & 0.17 $\pm$ 0.02 & 1.22 $\pm$ 0.04 & 0.21 $\pm$ 0.02 & 1.56 $\pm$ 0.03 & 6.70 $\pm$ 1.38 & 0.016 $\pm$ 0.004 \\[2pt]
    \hline
    \textbf{Ackley} & \textbf{5D$_x$} & \textbf{20} & 0.002 & 1.47 $\pm$ 0.14 & 0.012 $\pm$ 0.002 & 6.29 $\pm$ 0.72 & 0.050 $\pm$ 0.006 & 2.04 $\pm$ 0.34 & 0.011 $\pm$ 0.003 & 0.20 $\pm$ 0.02 & 1.58 $\pm$ 0.07 & 0.21 $\pm$ 0.03 & 1.87 $\pm$ 0.05 & 1.81 $\pm$ 0.31 & 0.011 $\pm$ 0.004 \\[2pt]
     &  &  & 0.08 & 1.53 $\pm$ 0.13 & 0.011 $\pm$ 0.001 & 5.48 $\pm$ 1.02 & 0.049 $\pm$ 0.004 & 1.62 $\pm$ 0.05 & 0.013 $\pm$ 0.001 & 0.21 $\pm$ 0.04 & 1.61 $\pm$ 0.06 & 0.24 $\pm$ 0.02 & 1.98 $\pm$ 0.07 & 1.84 $\pm$ 0.35 & 0.010 $\pm$ 0.002 \\[2pt]
    \textbf{Ackley} & \textbf{20D$_x$} & \textbf{20} & 0.002 & 9.54 $\pm$ 1.23 & 0.028 $\pm$ 0.003 & 66.38 $\pm$ 13.23 & 0.165 $\pm$ 0.016 & 21.32 $\pm$ 4.42 & 0.025 $\pm$ 0.002 & 0.25 $\pm$ 0.15 & 1.71 $\pm$ 0.08 & 0.29 $\pm$ 0.02 & 2.09 $\pm$ 0.05 & 11.00 $\pm$ 1.80 & 0.039 $\pm$ 0.006 \\[2pt]
     &  &  & 0.08 & 8.40 $\pm$ 1.12 & 0.026 $\pm$ 0.002 & 71.83 $\pm$ 10.97 & 0.166 $\pm$ 0.014 & 17.88 $\pm$ 4.50 & 0.025 $\pm$ 0.000 & 0.25 $\pm$ 0.03 & 1.74 $\pm$ 0.03 & 0.27 $\pm$ 0.02 & 2.11 $\pm$ 0.05 & 10.19 $\pm$ 0.68 & 0.031 $\pm$ 0.003 \\[2pt]
    \hline
    \textbf{Griewank} & \textbf{5D$_x$} & \textbf{20} & 0.002 & 1.28 $\pm$ 0.22 & 0.009 $\pm$ 0.003 & 5.55 $\pm$ 0.85 & 0.049 $\pm$ 0.004 & 1.69 $\pm$ 0.65 & 0.013 $\pm$ 0.002 & 0.21 $\pm$ 0.04 & 1.62 $\pm$ 0.07 & 0.24 $\pm$ 0.02 & 1.95 $\pm$ 0.04 & 2.43 $\pm$ 0.33 & 0.011 $\pm$ 0.002 \\[2pt]
     &  &  & 0.08 & 1.31 $\pm$ 0.21 & 0.011 $\pm$ 0.002 & 5.51 $\pm$ 0.67 & 0.057 $\pm$ 0.006 & 1.76 $\pm$ 0.54 & 0.010 $\pm$ 0.003 & 0.22 $\pm$ 0.02 & 1.59 $\pm$ 0.06 & 0.23 $\pm$ 0.03 & 1.95 $\pm$ 0.05 & 2.45 $\pm$ 0.27 & 0.010 $\pm$ 0.002 \\[2pt]
    \textbf{Griewank} & \textbf{20D$_x$} & \textbf{20} & 0.002 & 9.49 $\pm$ 0.91 & 0.026 $\pm$ 0.002 & 77.06 $\pm$ 7.30 & 0.168 $\pm$ 0.020 & 72.24 $\pm$ 18.49 & 0.026 $\pm$ 0.003 & 0.25 $\pm$ 0.04 & 1.76 $\pm$ 0.05 & 0.27 $\pm$ 0.02 & 2.15 $\pm$ 0.04 & 30.09 $\pm$ 3.98 & 0.034 $\pm$ 0.005 \\[2pt]
     &  &  & 0.08 & 9.00 $\pm$ 0.48 & 0.025 $\pm$ 0.004 & 87.16 $\pm$ 5.14 & 0.188 $\pm$ 0.020 & 43.14 $\pm$ 9.32 & 0.025 $\pm$ 0.003 & 0.23 $\pm$ 0.04 & 1.73 $\pm$ 0.05 & 0.26 $\pm$ 0.03 & 2.14 $\pm$ 0.04 & 10.92 $\pm$ 0.98 & 0.034 $\pm$ 0.001 \\[2pt]
    \hline
    \textbf{Zakharov} & \textbf{5D$_x$} & \textbf{20} & 0.002 & 1.31 $\pm$ 0.11 & 0.010 $\pm$ 0.002 & 5.42 $\pm$ 0.69 & 0.052 $\pm$ 0.004 & 1.54 $\pm$ 0.25 & 0.009 $\pm$ 0.003 & 0.22 $\pm$ 0.02 & 1.67 $\pm$ 0.06 & 0.22 $\pm$ 0.01 & 1.95 $\pm$ 0.03 & 2.85 $\pm$ 0.42 & 0.010 $\pm$ 0.002 \\[2pt]
     &  &  & 0.08 & 1.30 $\pm$ 0.13 & 0.008 $\pm$ 0.003 & 5.13 $\pm$ 0.62 & 0.050 $\pm$ 0.005 & 1.64 $\pm$ 0.31 & 0.011 $\pm$ 0.002 & 0.22 $\pm$ 0.15 & 1.67 $\pm$ 0.06 & 0.24 $\pm$ 0.03 & 1.98 $\pm$ 0.03 & 3.03 $\pm$ 0.50 & 0.009 $\pm$ 0.003 \\[2pt]
    \textbf{Zakharov} & \textbf{20D$_x$} & \textbf{20} & 0.002 & 8.94 $\pm$ 0.98 & 0.026 $\pm$ 0.003 & 102.70 $\pm$ 9.14 & 0.168 $\pm$ 0.058 & 17.25 $\pm$ 3.20 & 0.025 $\pm$ 0.003 & 0.24 $\pm$ 0.02 & 1.83 $\pm$ 0.05 & 0.28 $\pm$ 0.03 & 2.14 $\pm$ 0.03 & 15.29 $\pm$ 3.68 & 0.032 $\pm$ 0.005 \\[2pt]
     &  &  & 0.08 & 8.49 $\pm$ 1.43 & 0.023 $\pm$ 0.003 & 92.54 $\pm$ 7.29 & 0.169 $\pm$ 0.017 & 17.20 $\pm$ 2.21 & 0.028 $\pm$ 0.002 & 0.23 $\pm$ 0.02 & 1.84 $\pm$ 0.05 & 0.27 $\pm$ 0.02 & 2.12 $\pm$ 0.04 & 16.29 $\pm$ 2.43 & 0.034 $\pm$ 0.004 \\[2pt]
    \hline
    \textbf{Ackley} & \textbf{5D$_x$} & \textbf{40} & 0.002 & 3.47 $\pm$ 0.57 & 0.015 $\pm$ 0.002 & 69.68 $\pm$ 13.72 & 0.155 $\pm$ 0.011 & 4.77 $\pm$ 0.92 & 0.016 $\pm$ 0.003 & 0.25 $\pm$ 0.15 & 2.47 $\pm$ 0.06 & 0.33 $\pm$ 0.04 & 2.92 $\pm$ 0.03 & 1.45 $\pm$ 0.03 & 0.018 $\pm$ 0.001 \\[2pt]
     &  &  & 0.08 & 4.30 $\pm$ 0.79 & 0.016 $\pm$ 0.004 & 68.61 $\pm$ 11.94 & 0.151 $\pm$ 0.016 & 5.30 $\pm$ 1.05 & 0.017 $\pm$ 0.002 & 0.25 $\pm$ 0.05 & 2.44 $\pm$ 0.07 & 0.33 $\pm$ 0.01 & 2.90 $\pm$ 0.03 & 1.47 $\pm$ 0.04 & 0.018 $\pm$ 0.001 \\[2pt]
    \textbf{Ackley} & \textbf{20D$_x$} & \textbf{40} & 0.002 & 71.51 $\pm$ 15.81 & 0.064 $\pm$ 0.003 & 1155.20 $\pm$ 249.27 & 0.689 $\pm$ 0.033 & 156.18 $\pm$ 20.55 & 0.061 $\pm$ 0.004 & 0.33 $\pm$ 0.02 & 2.99 $\pm$ 0.05 & 0.44 $\pm$ 0.05 & 3.65 $\pm$ 0.06 & 27.18 $\pm$ 0.28 & 0.096 $\pm$ 0.005 \\[2pt]
     &  &  & 0.08 & 82.32 $\pm$ 17.84 & 0.063 $\pm$ 0.004 & 1123.58 $\pm$ 175.75 & 0.731 $\pm$ 0.032 & 199.22 $\pm$ 39.32 & 0.065 $\pm$ 0.005 & 0.33 $\pm$ 0.02 & 3.04 $\pm$ 0.05 & 0.43 $\pm$ 0.010 & 3.63 $\pm$ 0.05 & 26.99 $\pm$ 0.27 & 0.091 $\pm$ 0.005 \\[2pt]
    \hline
    \textbf{Dixon-Price} & \textbf{5D$_x$} & \textbf{40} & 0.002 & 4.13 $\pm$ 0.94 & 0.016 $\pm$ 0.002 & 66.41 $\pm$ 14.90 & 0.168 $\pm$ 0.020 & 5.21 $\pm$ 0.79 & 0.015 $\pm$ 0.003 & 0.25 $\pm$ 0.03 & 2.47 $\pm$ 0.06 & 0.32 $\pm$ 0.02 & 2.91 $\pm$ 0.03 & 1.46 $\pm$ 0.03 & 0.018 $\pm$ 0.003 \\[2pt]
     &  &  & 0.08 & 3.87 $\pm$ 0.90 & 0.013 $\pm$ 0.004 & 68.17 $\pm$ 15.31 & 0.166 $\pm$ 0.017 & 5.14 $\pm$ 0.75 & 0.013 $\pm$ 0.005 & 0.27 $\pm$ 0.06 & 2.48 $\pm$ 0.08 & 0.31 $\pm$ 0.02 & 2.89 $\pm$ 0.03 & 1.46 $\pm$ 0.04 & 0.018 $\pm$ 0.003 \\[2pt]
    \textbf{Dixon-Price} & \textbf{20D$_x$} & \textbf{40} & 0.002 & 101.75 $\pm$ 16.21 & 0.063 $\pm$ 0.003 & 868.13 $\pm$ 193.62 & 0.734 $\pm$ 0.050 & 266.32 $\pm$ 41.11 & 0.062 $\pm$ 0.004 & 0.34 $\pm$ 0.35 & 2.97 $\pm$ 0.05 & 0.42 $\pm$ 0.03 & 3.63 $\pm$ 0.03 & 27.06 $\pm$ 0.34 & 0.092 $\pm$ 0.004 \\[2pt]
     &  &  & 0.08 & 103.87 $\pm$ 18.33 & 0.065 $\pm$ 0.003 & 1011.60 $\pm$ 136.63 & 0.650 $\pm$ 0.028 & 265.87 $\pm$ 43.41 & 0.064 $\pm$ 0.004 & 0.34 $\pm$ 0.03 & 2.99 $\pm$ 0.05 & 0.41 $\pm$ 0.02 & 3.63 $\pm$ 0.05 & 26.91 $\pm$ 0.18 & 0.090 $\pm$ 0.005 \\[2pt]
    \hline
    \textbf{Rosenbrock} & \textbf{5D$_x$} & \textbf{80} & 0.002 & 9.37 $\pm$ 1.80 & 0.030 $\pm$ 0.003 & 345.16 $\pm$ 175.77 & 0.551 $\pm$ 0.010 & 15.26 $\pm$ 3.01 & 0.027 $\pm$ 0.003 & 0.36 $\pm$ 0.05 & 4.95 $\pm$ 0.36 & 0.56 $\pm$ 0.05 & 24.60 $\pm$ 0.16 & 6.07 $\pm$ 0.08 & 0.039 $\pm$ 0.003 \\[2pt]
     &  &  & 0.08 & 11.11 $\pm$ 5.77 & 0.029 $\pm$ 0.003 & 387.19 $\pm$ 146.21 & 0.551 $\pm$ 0.006 & 17.45 $\pm$ 4.17 & 0.029 $\pm$ 0.004 & 0.33 $\pm$ 0.02 & 4.71 $\pm$ 0.09 & 0.54 $\pm$ 0.02 & 26.65 $\pm$ 0.02 & 6.18 $\pm$ 0.11 & 0.034 $\pm$ 0.004 \\[2pt]
    \textbf{Rosenbrock} & \textbf{20D$_x$} & \textbf{80} & 0.002 & 123.88 $\pm$ 223.89 & 1.595 $\pm$ 0.048 & 2136.47 $\pm$ 4947.49 & 4.602 $\pm$ 0.173 & 554.87 $\pm$ 148.11 & 1.514 $\pm$ 0.043 & 0.55 $\pm$ 0.33 & 17.88 $\pm$ 1.10 & 0.96 $\pm$ 0.02 & 47.17 $\pm$ 0.25 & 53.22 $\pm$ 13.55 & 0.294 $\pm$ 0.836 \\[2pt]
     &  &  & 0.08 & 118.80 $\pm$ 193.67 & 1.621 $\pm$ 0.067 & 2536.21 $\pm$ 4600.47 & 4.543 $\pm$ 0.089 & 594.46 $\pm$ 158.83 & 1.502 $\pm$ 0.046 & 0.53 $\pm$ 0.03 & 18.13 $\pm$ 0.49 & 0.96 $\pm$ 0.03 & 47.15 $\pm$ 0.03 & 56.99 $\pm$ 9.85 & 0.282 $\pm$ 0.871 \\[2pt]
    \bottomrule
    \end{tabular}
    }
    \label{tab timing loo}
\end{table*}

    \pagebreak
    \bibliography{R_Ref}

\end{document}